\documentclass[preprint, authoryear]{elsarticle}
\makeatletter
\let\c@author\relax
\makeatother

\usepackage[utf8]{inputenc} 
\usepackage[T1]{fontenc}    
\usepackage{hyperref}       
\usepackage{url}            
\usepackage{booktabs}       
\usepackage{microtype}      
\usepackage{graphicx}		
\usepackage{subcaption}		
\usepackage{newunicodechar} 
\usepackage{wrapfig}
\usepackage{float}
\newcommand{\beginsupplementaryfigures}{
	\setcounter{figure}{0}
	\renewcommand{\thefigure}{S\arabic{figure}}
}
\newunicodechar{ﬁ}{fi}
\newunicodechar{ﬀ}{ff}

\graphicspath{{newfigsjpeg/}}
\author[abdn]{Ben Lonnqvist\corref{cor1}}
\ead{ben.lonnqvist.16@aberdeen.ac.uk}
\author[essex]{Alasdair D. F. Clarke}
\ead{a.clarke@essex.ac.uk}
\author[psychabdn]{Ramakrishna Chakravarthi}
\ead{rama@abdn.ac.uk}

\cortext[cor1]{Corresponding author}

\address[abdn]{Business School, University of Aberdeen}
\address[essex]{Department of Psychology, University of Essex}
\address[psychabdn]{School of Psychology, University of Aberdeen}

\title{Crowding in humans is unlike that in convolutional neural networks}

\begin{document}

	\begin{abstract}
		Object recognition is a primary function of the human visual system. It has recently been claimed that the highly successful ability to recognise objects in a set of emergent computer vision systems---Deep Convolutional Neural Networks (DCNNs)---can form a useful guide to recognition in humans. To test this assertion, we systematically evaluated visual crowding, a dramatic breakdown of recognition in clutter, in DCNNs and compared their performance to extant research in humans. We examined crowding in three architectures of DCNNs with the same methodology as that used among humans. We manipulated multiple stimulus factors including inter-letter spacing, letter colour, size, and flanker location to assess the extent and shape of crowding in DCNNs. We found that crowding followed a predictable pattern across architectures that was different from that in humans. Some characteristic hallmarks of human crowding, such as invariance to size, the effect of target-flanker similarity, and confusions between target and flanker identities, were completely missing, minimised or even reversed. These data show that DCNNs, while proficient in object recognition, likely achieve this competence through a set of mechanisms that are distinct from those in humans. They are not necessarily equivalent models of human or primate object recognition and caution must be exercised when inferring mechanisms derived from their operation.
	\end{abstract}
	
	
	\maketitle
	\section{Introduction}
	\label{introduction}
	Recognising objects is a central function of the human visual system and the mechanisms underlying this ability have been extensively studied \citep{dicarlo_how_2012, ullman_object_2007}. One approach to studying human object recognition is to examine situations where it fails in order to determine the constraints for successful recognition. Visual crowding is one such failure of object recognition in human vision \citep{bouma_interaction_1970, levi_crowding--essential_2008, manassi_multi-level_2018} where objects that are otherwise recognisable in the visual periphery are rendered unrecognisable when surrounded by similar clutter. Studies on visual crowding have given rise to multi-stage models of object recognition \citep{pelli_crowding_2004}.
	
	In computer vision, deep convolutional neural networks (DCNNs) have proven to be extremely successful, reaching high accuracy rates in many object recognition and classification tasks \citep{simonyan_very_2014, szegedy_going_2014, he_deep_2015, huang_densely_2016}. DCCNs are loosely inspired by the human visual system and have been argued to be compelling models of primate object recognition  \citep{cadieu_deep_2014, khaligh-razavi_deep_2014, guclu_deep_2015, yamins_using_2016, bonner_coding_2017}. However, interpreting both the decision process and the relationship between inputs and layers’ outputs is difficult, and many approaches to interpreting and understanding DCNNs have been taken \citep{zeiler_visualizing_2013, zhang_interpretable_2017}. The goal of our paper is not to interpret the low-level details of the DCNN decision process, but rather to investigate if DCNNs suffer from human-like crowding patterns, and if so, whether examining these breakdowns in DCNNs can shed light on the mechanisms of object recognition. If DCNNs are to serve as fruitful models of human neural computations, it is crucial to determine the similarities and differences between human and computer vision models. That is, if DCNNs recognise objects using mechanisms analogous to that in humans, then they too should be subject to the flanker-induced interference observed in humans. It is important to understand the behaviour of crowding in DCNNs not only to help us better understand the human visual system, but also to be able to design more efficient computer vision systems.
	
	The phenomenon of crowding in humans displays certain distinctive features. Here, we highlight the most salient and relevant aspects, which form by no means an exhaustive list of its properties. The most striking observation in crowding is that closer flankers interfere with the identification of a target more than distant flankers; that is, the spacing between targets and their flankers strongly modulates identification performance \citep{bouma_interaction_1970, toet_two-dimensional_1992, pelli_crowding_2004}. Further, for a fixed spacing between a target and its flankers, crowding (interference) is stronger at larger target eccentricities (distance from fixation; \citet{toet_two-dimensional_1992, pelli_crowding_2004}). Crucially, the flankers interfere with the target over a limited region of space that scales with eccentricity. Under standard circumstances, flankers further than half the target’s eccentricity do not crowd the target. This relationship has been called the Bouma Law \citep{pelli_uncrowded_2008}. The relationship seems to hold true for a wide range of objects, from simple features such as oriented gratings and colour to complex real-world objects \citep{berg_generality_2007, wallace_object_2011}. Additionally, the size of the objects does not seem to affect crowding: small objects crowd each other as much as large objects do \citep{pelli_crowding_2004}. Hence, it was proposed that the distance between the centres of the objects is more relevant than the distance between edges\footnote{Although, there are several caveats to this `law' \citep{herzog_crowding_2015, livne_configuration_2007}.}. Another interesting characteristic of crowding, alluded to above, is that crowding occurs between similar objects but not dissimilar ones \citep{kooi_effect_1994, kennedy_chromatic_2010}. For example, a black letter is strongly crowded by other black letters, but less so by white letters or filled black circles. Finally, visual crowding displays various asymmetries. The most prominent of these asymmetries is the radial-tangential asymmetry: flankers that are in the radial direction (along the axis connecting the fovea and the target) lead to more interference than flankers that are in the tangential direction \citep{toet_two-dimensional_1992, petrov_asymmetries_2011}.
	
	Whereas visual crowding has been rigorously tested in humans over the past five decades \citep{bouma_interaction_1970, pelli_crowding_2004}, little is known about crowding in DCNNs. We know of only one previous study, in which \citet{volokitin_deep_2017} argued for the existence of crowding in DCNNs. However, their experiments do not conclusively establish crowding in DCNNs or test their similarity to humans, as their results might be explained by their method to achieve acuity loss, whereby the centres of stimuli are repeatedly sampled with increasingly higher resolution. That is, the models may have exhibited an unnatural preference to process the most central object, which reduced its ability to identify a flanked target. The models used in their research are small-scale and not capable of human-like performance, and might as such not reliably exhibit complex behaviour, such as crowding. Additionally, the methodology used in their research is different from most human crowding research. As such, to establish a conclusive and comparable picture of crowding in DCNNs, more research is needed.
	
	In this paper we take various successful architectures of DCNNs, including ones that have been previously claimed to be comparable to the human visual system \citep{cichy_comparison_2016, guclu_deep_2015, kheradpisheh_deep_2016}, and investigate the the presence and characteristics of visual crowding using methodology inspired by human crowding research. We will assess the effect of the following on target identification:
	\begin{itemize}
		\item The distance between the target and the flankers
		\item The position of the target and the flankers
		\item The size and contrast polarity of the target and the flankers 
		\item Different targets and flanker identities
	\end{itemize}
	
	The last two test the effect of similarity. To preview our results, we find that the strength of crowding, defined as flanker-induced reduction in target identification, in DCNNs varies according to the kind of network. However, the results shows a peculiar pattern that appears to be independent of the topology of the network. This pattern is in many ways dissimilar from that in humans. Finally, we discuss how these findings affect our understanding of object recognition in humans and DCNNs, and raise concerns that those employing DCNNs in object recognition tasks should keep in mind.
	
	\section{Methods}
	
	\subsection{Models}
	We investigated three sets of DCNNs of increasing complexity (and chronology). First, we examined a network that has been widely claimed to possess characteristics similar to that of the human visual system \citep{cichy_dynamics_2017, guclu_deep_2015, kheradpisheh_deep_2016}. That is, the various layers of this network are thought to capture the basic computational processes implemented by the layers of the primate visual system (from V1 to Infero-Temporal Cortex or IT). This network is a variant of the successful AlexNet \citep{Krizhevsky:2012:ICD:2999134.2999257} with 5 convolutional layers and 3 fully connected layers, followed by an activation layer; we will call this network SimpleNet throughout. We also investigated the VGG-16 network \citep{simonyan_very_2014}, which is a more successful 16-layer DCNN that uses small (3x3) filters and achieves a deeper network compared to other similar networks of its time. The family of VGG-networks achieved state-of-the-art or near state-of-the-art performance in 2014 image classification and localisation challenges. Finally, we also tested DenseNet-121 \citep{huang_densely_2016}, a 121-layer DCNN that takes advantage of two recent advancements in deep learning: batch normalisation \citep{ioffe_batch_2015} and skip connections. The DenseNet-family of networks achieved state-of-the-art performance in many competitive image classification benchmarks while being parameter-efficient. While it is much more successful than the previous two networks, and has a much deeper architecture, it is important to note that the DenseNet-121 has fewer trainable parameters than the VGG-16. We tested these different architectures, and particularly the DenseNet-121, for two reasons. We wanted to test whether networks that are highly successful in recognising objects are in general susceptible to clutter, or if certain networks recognise objects in such a way that they are robust to flanker presence. Second, we wanted to test if networks considered to be similar to the primate visual system also show characteristics of humans, which include crowding. It has been claimed that even deep networks such as DenseNet and ResNet, of which DenseNet is a variant, \citep{he_resnet_2015} are comparable to the primate visual system. In fact, recent investigations demonstrate that such networks are superior to older networks such as AlexNet and VGG-16 in terms of correspondence to the primate system \citep{SchrimpfKubilius2018BrainScore}. Hence, it is appropriate to test a range of networks to determine if they suffer from crowding.
	
	\begin{figure}
		\centering
		\includegraphics[width=0.8\textwidth]{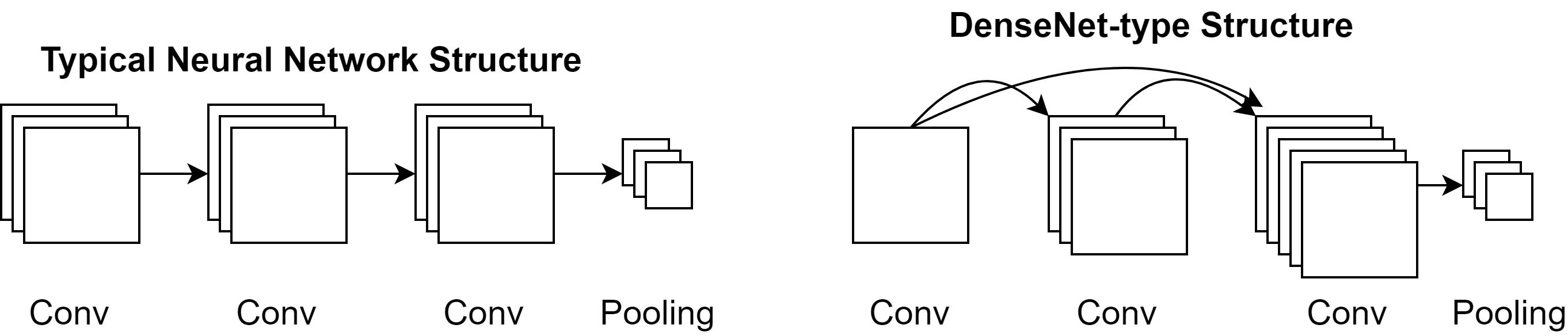}
		\caption{The skip connections of the DenseNet family of architectures. `Conv’ refers to a 2D Convolutional layer.}
	\end{figure}
	
	The DenseNet was of particular interest to us, as it includes skip connections, which are also believed to be present in the human visual cortex  \citep{essen_hierarchical_1983}. Here, a layer’s feature maps are connected to the filters of all layers that follow it within a given `dense block' (described below). For example, for $n$ layers, layer 1’s feature maps are connected to all layers’ inputs up to the $n$th layer. This process is repeated for all $n$ layers. DenseNet-121 implements this architecture within `dense blocks', where a set of layers is densely connected (skip connected) to each other, and at the final layer of the block, the feature maps are pooled using max pooling.
	
	In our research, we changed the rectified linear unit (ReLU) activations of the DenseNet and VGG to Leaky ReLU activations to avoid ‘dying neurons’ (neurons which do not allow a gradient to flow through them and end up in a perpetually inactive state) \citep{xu_empirical_2015}.
	
	We focused our primary attention not on small (either in number of parameters or depth of layers) models, such as those tested by \citep{volokitin_deep_2017}, as we wanted to investigate the behaviour of complex networks that have proved to be successful at identifying and categorising real-world images, to understand the patterns of crowding that could emerge from such networks. Additionally, while some have experimented with eccentricity-dependent models \citep{mnih_recurrent_2014}, we limited the scope of our research to better-established DCNN classes. 
	
	\subsection{Stimuli and Experimental Setup}
	\label{stimulisection}
	Two types of stimuli were used in the experiments. The first type was images of places from the Places2 dataset \citep{zhang_interpretable_2017}, which we will refer to as backgrounds. Two classes of backgrounds were used: ruins and neighbourhoods. We used these classes because they are relatively similar in shapes, requiring the networks to construct more general types of filters that might mimic general scene recognition filters, attempting to avoid egregious overfitting of our next type of stimuli. 
	
\begin{figure}
		\centering
		\textit{(a)}
		\begin{subfigure}[1]{0.2\textwidth}
			\includegraphics[width=\textwidth, keepaspectratio]{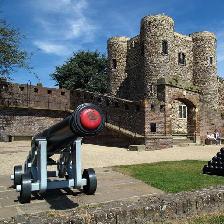}
		\end{subfigure}
		{\huge $\Rightarrow$}
		\begin{subfigure}[2]{0.2\textwidth}
			\includegraphics[width=\textwidth, keepaspectratio]{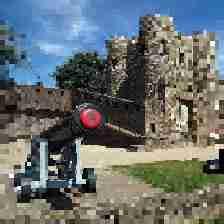}
		\end{subfigure}
		\qquad
		\qquad
		\textit{(b)}
		\begin{subfigure}[3]{0.2\textwidth}
			\includegraphics[width=\textwidth, keepaspectratio]{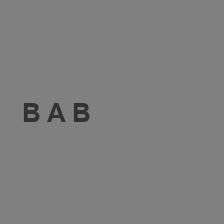}
		\end{subfigure}
		\caption{Example stimuli. \textit{(a)} shows acuity reduction in images. Acuity is reduced logarithmically between values of acuity $=1$ and acuity $=0.2$ with linear distance from the centre of the image in 20 steps. \textit{(b)} shows a full acuity letter stimulus with the target letter A, and pair flankers B.}
		\label{examplestimuli}
	\end{figure}

	The second type of stimuli were uniform grey backgrounds with letters fixed in position, which we will call targets. These stimuli are akin to the stimuli used in psychophysical experiments on crowding \citep{bouma_interaction_1970, pelli_crowding_2004}. There were 8 different target letters: \{A, B, C, E, G, M, Y, Q\}, and each of them was considered a distinct class, making a total of 10 classes of training stimuli (8 letter image classes and 2 background image classes). We chose this set of letters because they are visually dissimilar from each other, which minimises the error rate, particularly when the acuity reduction procedure was applied to images (see Figure \ref{examplestimuli} \textit{(a)}), which could have caused confusions between letters. The letters could be of either contrast polarity, near-white and near-black on a grey background, and one of two sizes, 20 and 26 points. All stimuli were 224x224 pixels.
	Each network was trained on these 10 classes of images. When trained on letters, a single letter was presented 56 pixels to the left of the centre of the image, that is, midway between the centre and the left border of the stimulus along the horizontal meridian. Similarly, during testing, a target letter was presented at the location it was trained at. It was flanked by one letter or a pair of letters. When two letters were presented, one letter was placed on each side of the target and were identical to each other. The flankers were selected from a set that included all target letters and two additional letters: \{S, H\}. The pair of flankers were placed diametrically opposite each other on either side of the target. Each pair of flankers was tested at 10 angular locations around the target, each location separated by 18 degrees of rotation, thus covering the entire region around the target. The centre-to-centre distance between a target and each flanker ranged from 25 to 45 pixels in 2-pixel increments. All combinations of target and flanker letters, contrast polarities and sizes were tested. In total, we tested 70,400 combinations of flankers and targets in each experiment. In experiments where we tested the effect of single flankers, the number of tested combinations doubled (20 angular locations instead of 10). 
	
	To study crowding in DCCNs, we wished to model human peripheral vision. This is because crowding in humans occurs most noticeably away from the fovea in peripheral vision, where visual acuity and resolution is much lower than in the centre of the visual field. We wanted to provide the DCNNs the same sort of input as the human visual system would receive. Peripheral input is impoverished relative to central input. To model peripheral vision, we used well established relationships in humans regarding acuity and eccentricity \citep{anstis_letter:_1974} and reduced acuity logarithmically with distance from the centre of the image in 20 steps, with 1 being full acuity in the centre of the image, and 0.2 being the lowest acuity at the edges of the image. We first took 20 copies of the image and assigned each a value on a logarithmic scale, ranging from 0.2 to 1. We then down-sampled each image by their assigned value, and up-sampled them to their original size using the nearest neighbour algorithm. Finally, we cropped and overlaid the images on top of each other to form a 20-step gradient of acuity reduction (see Figure \ref{examplestimuli} \textit{(a)} for an example). We did this to strictly lose information, as crowding in humans is not simply blur \citep{song_double_2014}. 
	
	We would like to emphasise that training was done on two kinds of backgrounds and 8 target unflanked letters presented in isolation; flankers were introduced only in the testing stage. Model base performance was evaluated on the set of target letters, and a separate set of validation backgrounds.
	
	\begin{figure}
		\centering
		\includegraphics[width=0.8\textwidth]{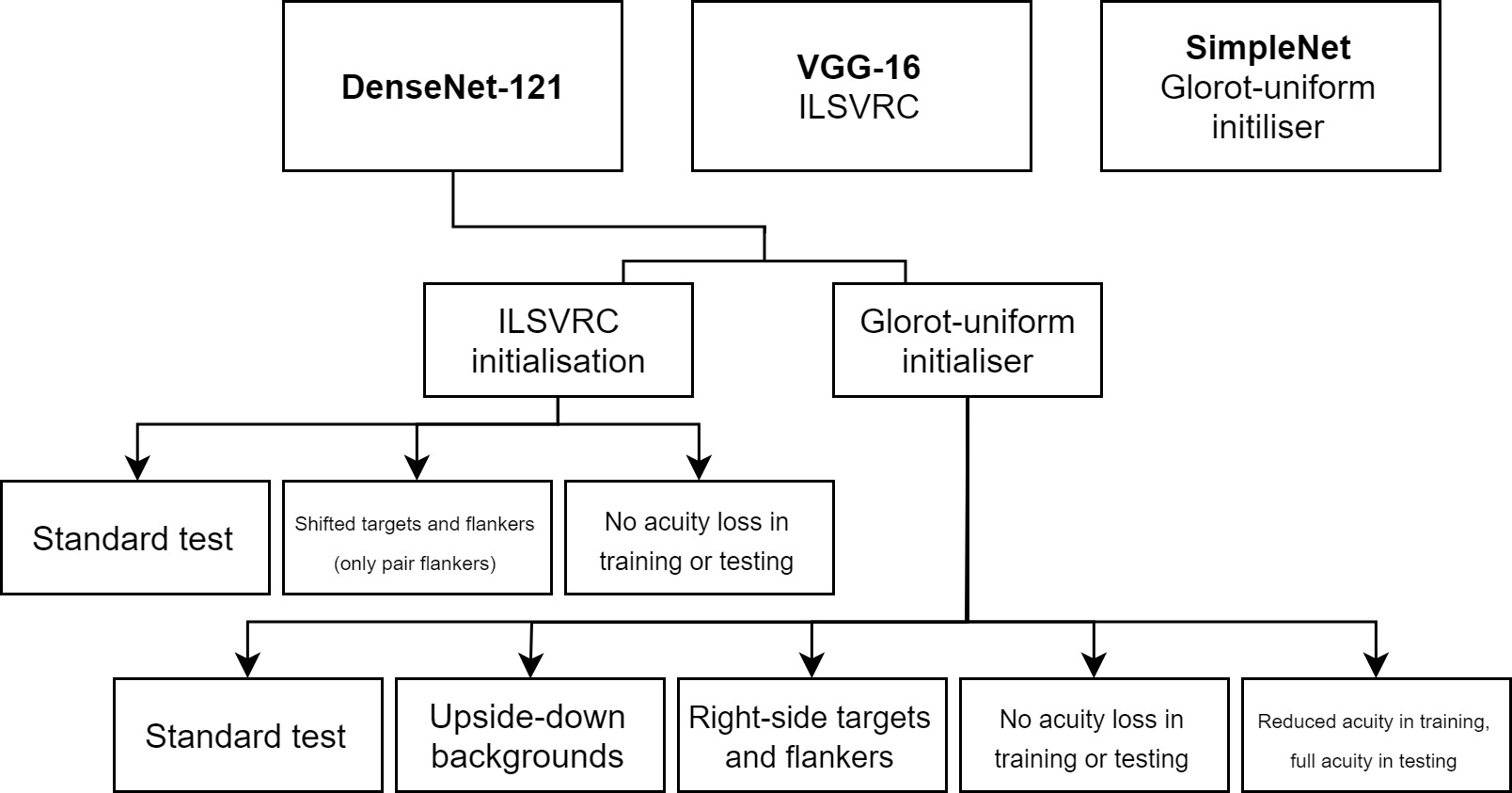}
		\caption{The range of experiments conducted in this study.}
	\end{figure}

	\subsection{Training}
	
	All models for all experiments were trained for 24 hours\footnote{This corresponds to roughly 40 epochs on the DenseNet, 80 epochs on the VGG and 200 epochs on the SimpleNet. Although this is an arbitrary time limit, given our configuration these models were run for a sufficient number of epochs to enable good recognition performance, similar to what has been implemented in earlier studies (e.g., \citet{simonyan_very_2014}.} on an NVIDIA Tesla K40c GPU using the Keras library \citep{chollet2015keras}. The ADAM-optimiser \citep{kingma_adam:_2014} was used with a learning rate of 0.01. Both random initialisation\footnote{As a random initialiser, we use the Glorot-uniform initialiser \citep{glorot_understanding_2010}.} and ImageNet Large Scale Visual Recognition Competition (ILSVRC) initialisation of weights\footnote{ILSVRC initialisation of weights refers to initial weights of the neural network as being set to the weights optimised for the ImageNet Large Scale Visual Recognition Challenge classification task (See Keras documentation; \citet{chollet2015keras}).} were tested on the DenseNet, random initialisation was tested on the SimpleNet and ILSVRC initialisation on the VGG-16. Random initialisation allowed us to test the network’s characteristics and performance in the absence of influence from outside sources on the system and controlled for the possibility that any results may have been caused by ILSVRC initialisation of weights. Initialising the network with ILSVRC weights allowed us to mimic the types of environments humans are subjected to on a regular basis in addition to testing an already trained network that has been shown to be successful in image categorisation and identification. It is important to note, however, that the ILSVRC weights had been trained without acuity loss, while our training and testing was primarily conducted on stimuli that had been reduced in acuity. When initialising the network with ILSVRC weights, the following procedure for training was taken to allow stable training and avoid `gradient nuking’\footnote{When using weights optimised for a specific task (e.g. ILSVRC), using them for a different task may cause large gradient updates in the final layers of the network which can cause large changes in the weights of the layers above them.} in the upper layers of the network:
	\begin{enumerate}
		\item Freeze all layers above the last one, initialise learning rate $=0.01$.
		\item When validation loss does not decrease for 2 epochs, open the next layer for training and reduce learning rate by ${10}^{-2}$.
		\item When validation loss does not decrease for 2 epochs, open all layers for training and reduce learning rate by ${10}^{-2}$.
		\item Training is completed after a total of 24 hours.
	\end{enumerate}
	
	\section{Results}
	In our experiments, we did not train the network to recognise targets in the presence of flankers, or letters in the locations where flankers were later placed. Our goal was to present the targets to the models in a specific part of the image, such that it learns to recognise it. We then tested its performance in the presence of flankers. Humans crowding has been attributed to either confusion a fully identified flanker for a target or to combining or pooling the features of both the target and flankers \citep{strasburger_source_2013, hanus_quantifying_2013}. We allowed our models the opportunity to implement either of these processes\footnote{Full data-frames of results are available at \href{https://github.com/benlonnqvist/CNNCrowding}{github.com/benlonnqvist/CNNCrowding}.}. 
	
\begin{figure}
		\centering
		\begin{subfigure}{0.9\textwidth}
			\includegraphics[width=\textwidth]{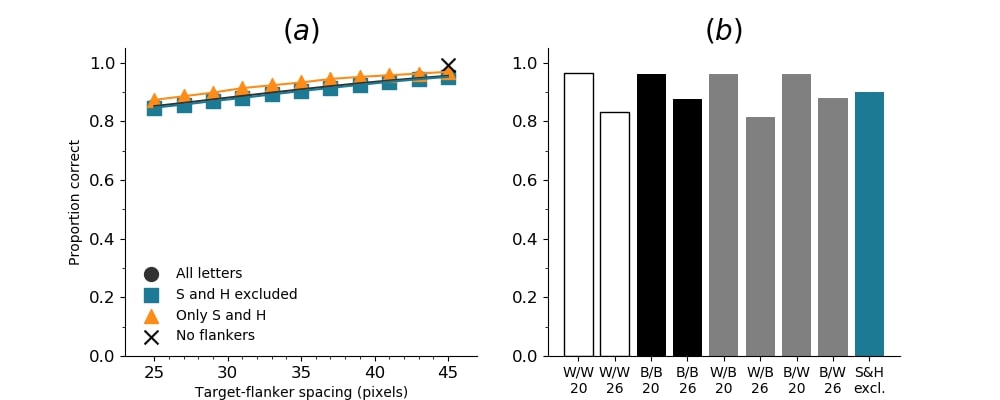}
		\end{subfigure}
		\begin{subfigure}{0.2562\textwidth}
			\includegraphics[width=\textwidth]{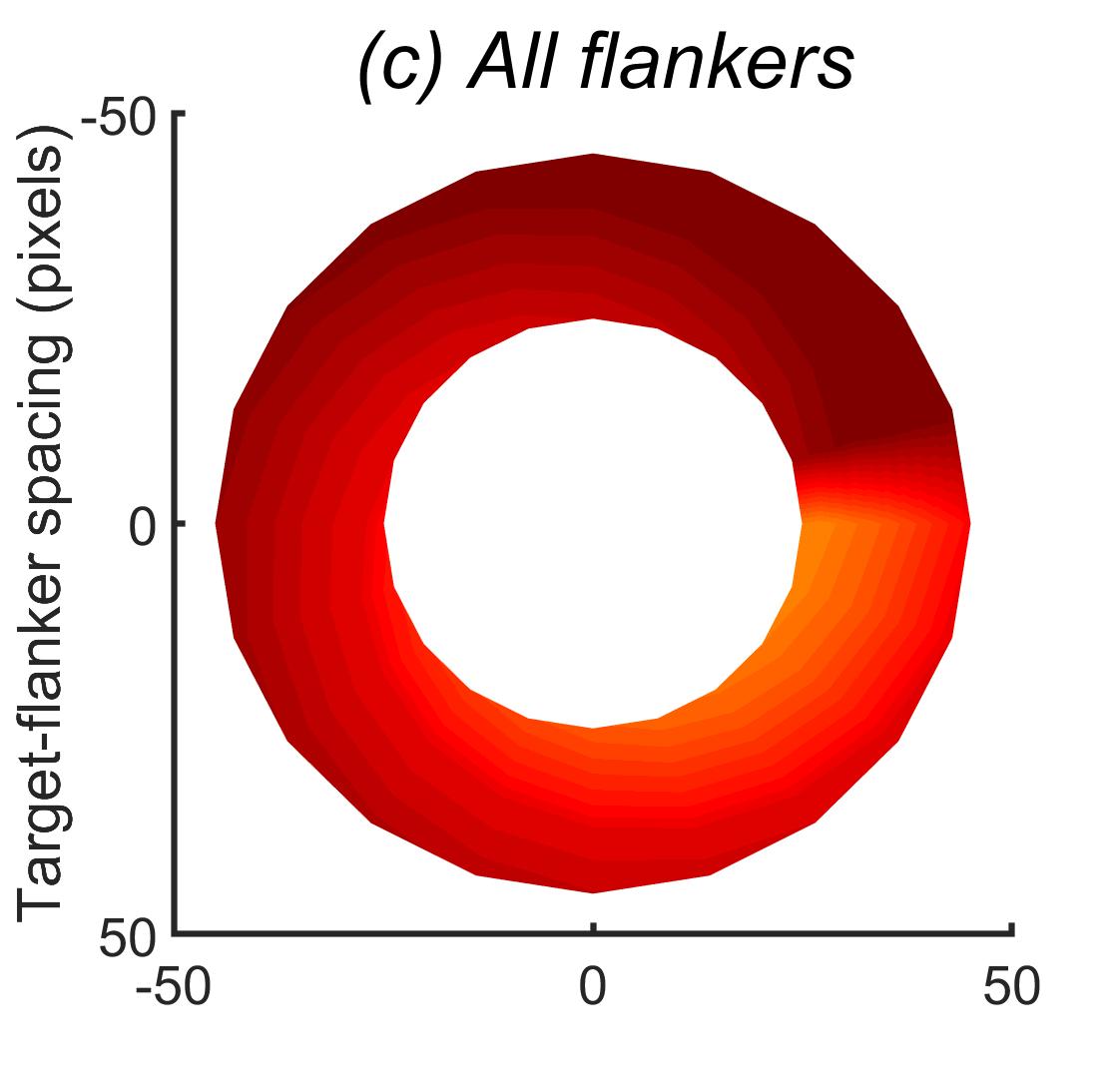}
		\end{subfigure}
		\begin{subfigure}{0.25\textwidth}
			\includegraphics[width=\textwidth]{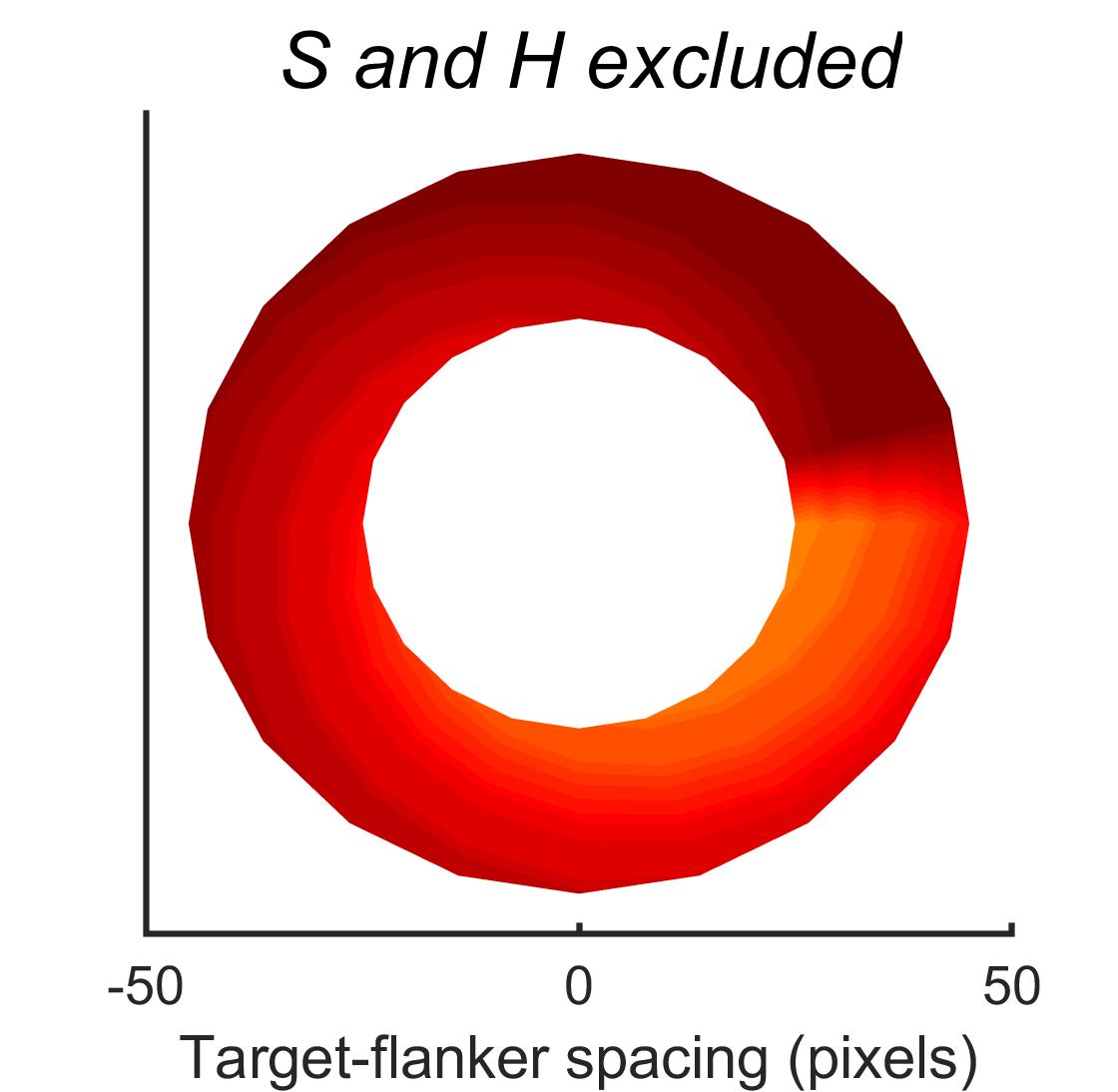}
		\end{subfigure}
		\begin{subfigure}{0.25\textwidth}
			\includegraphics[width=\textwidth]{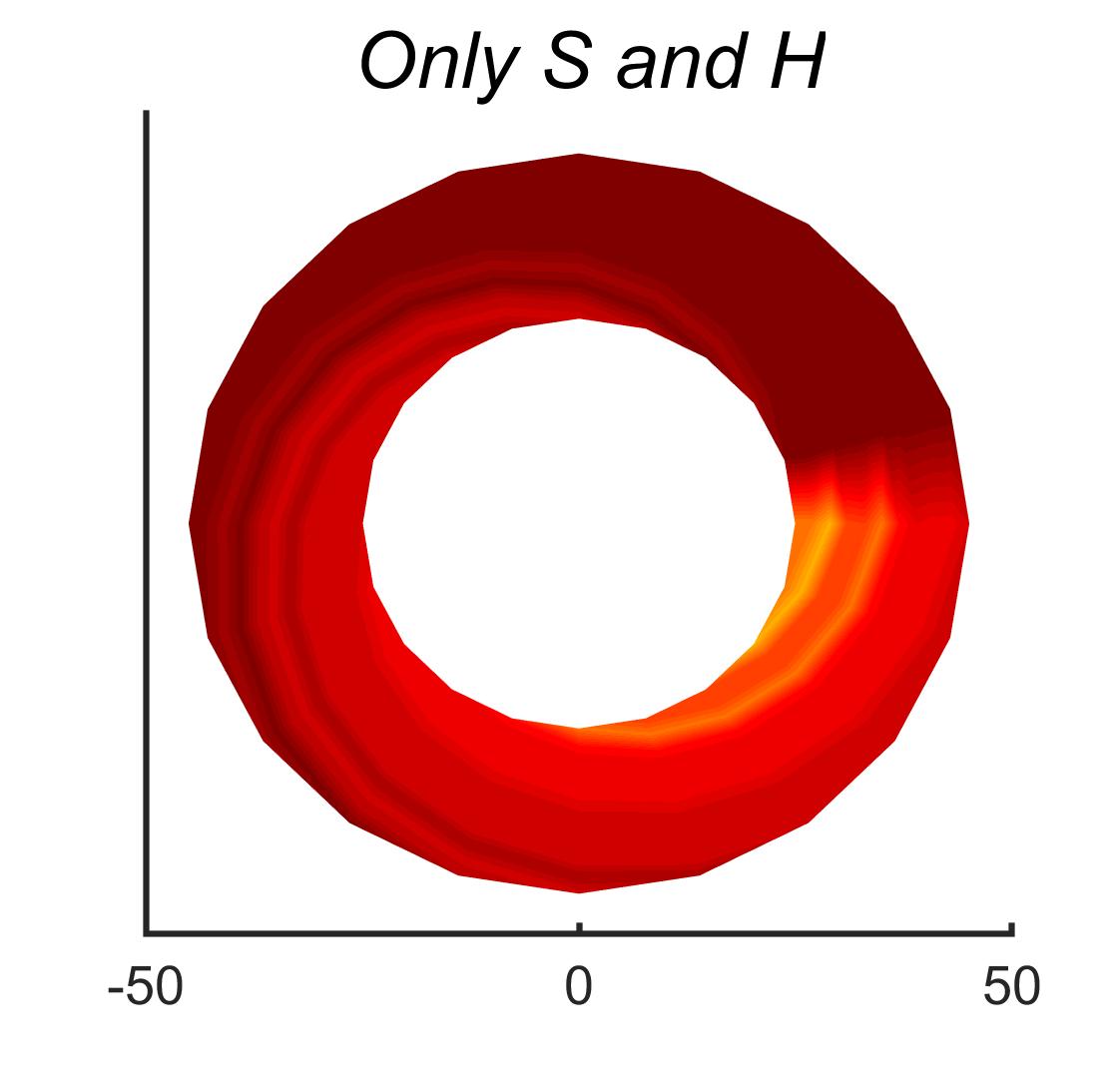}
		\end{subfigure}
		\begin{subfigure}{0.06\textwidth}
			\includegraphics[width=\textwidth]{other/colorbar.jpg}
		\end{subfigure}
		\caption{Accuracy of letter identification of the DenseNet-121 for stimuli that were not reduced in acuity. ILSVRC initialisation with single flankers. Testing and training were done without acuity loss. Model accuracy without flankers was 99.34\%.}
		\label{dnoacuitylossimagenetsingle}
	\end{figure}

\begin{figure}
		\centering
		\begin{subfigure}{0.9\textwidth}
			\includegraphics[width=\textwidth]{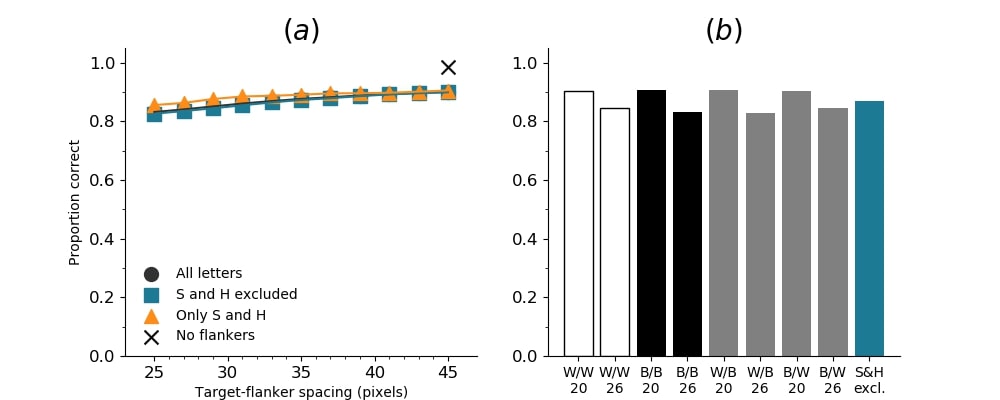}
		\end{subfigure}
		\begin{subfigure}{0.2562\textwidth}
			\includegraphics[width=\textwidth]{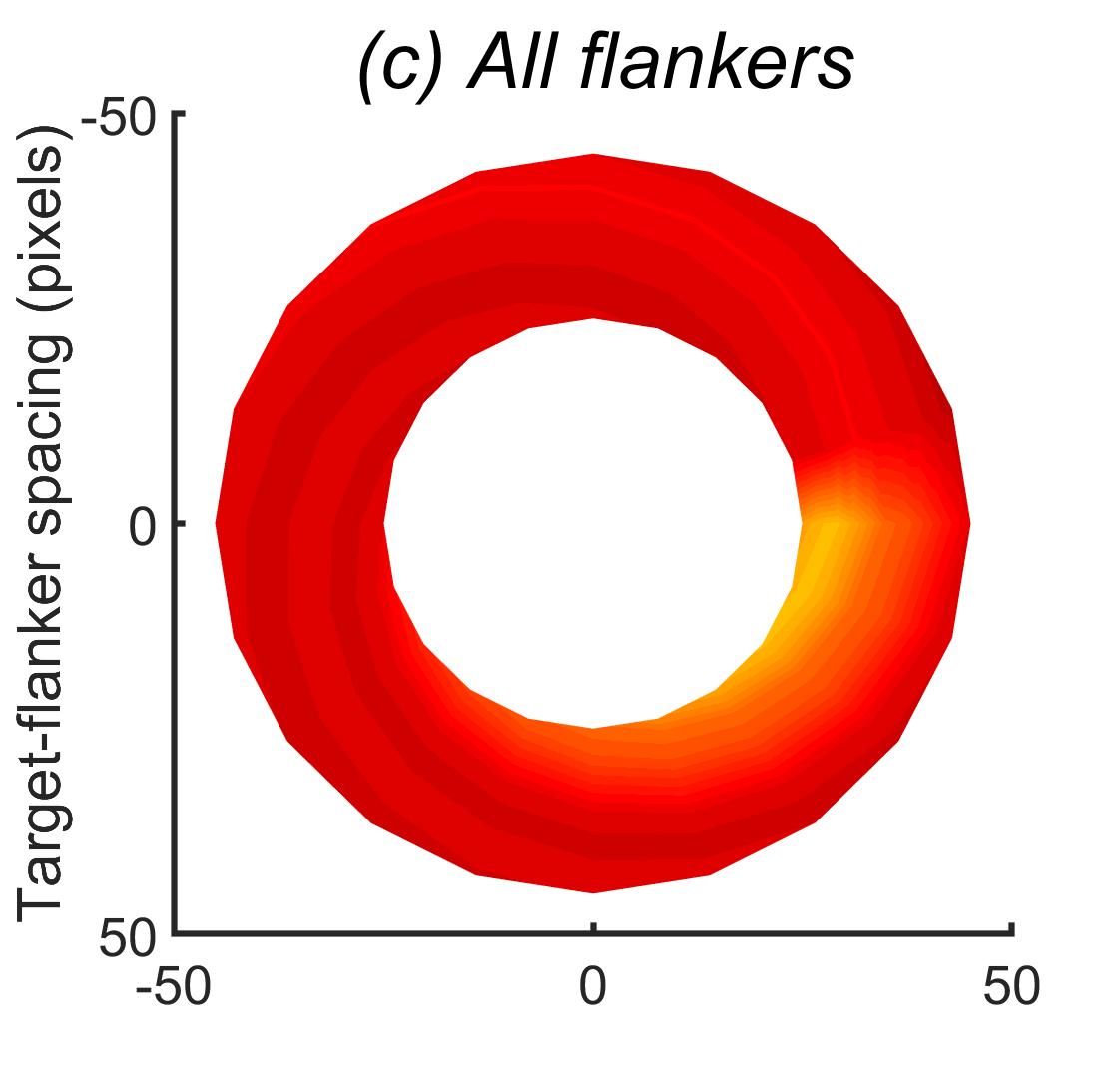}
		\end{subfigure}
		\begin{subfigure}{0.25\textwidth}
			\includegraphics[width=\textwidth]{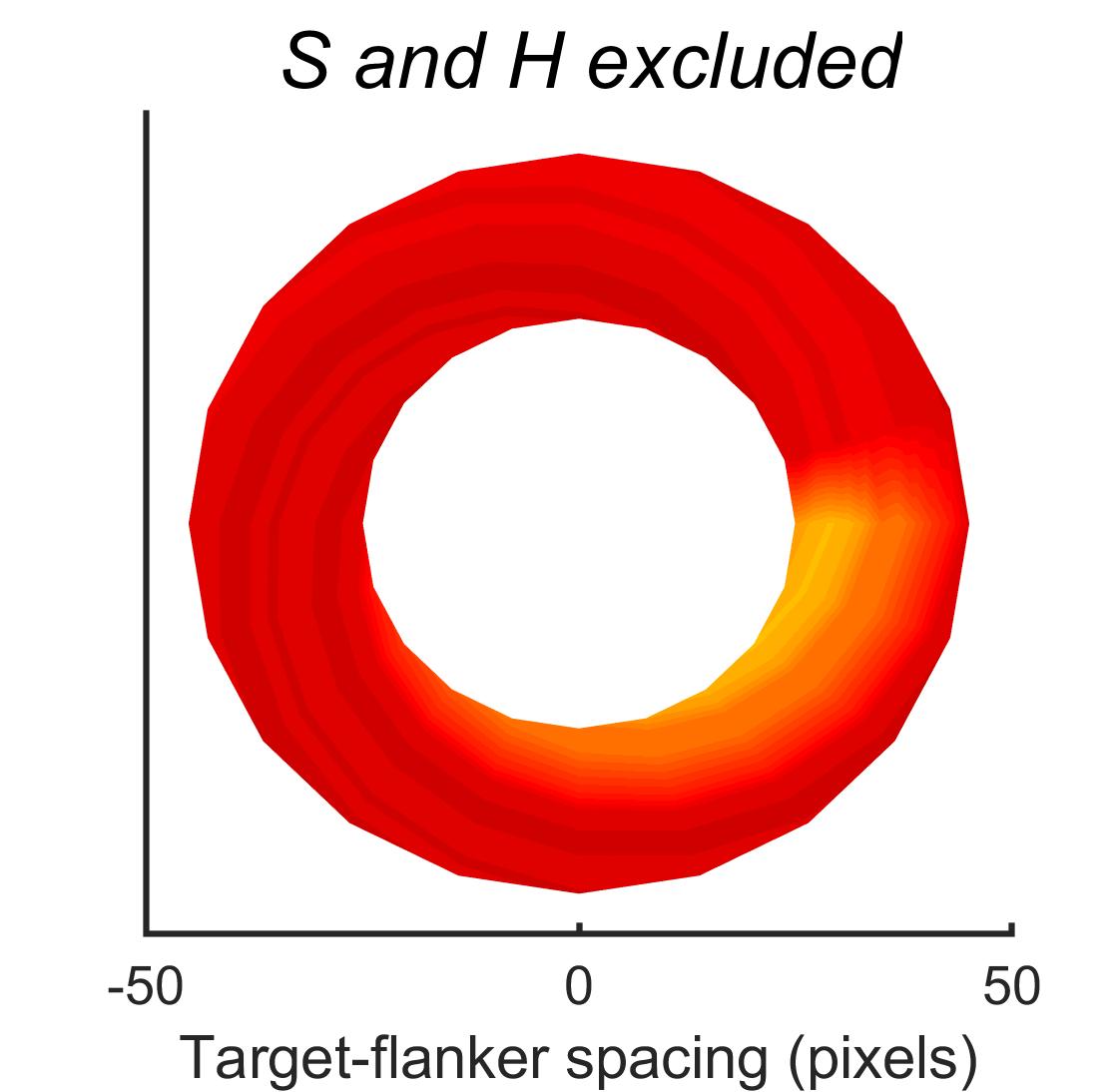}
		\end{subfigure}
		\begin{subfigure}{0.25\textwidth}
			\includegraphics[width=\textwidth]{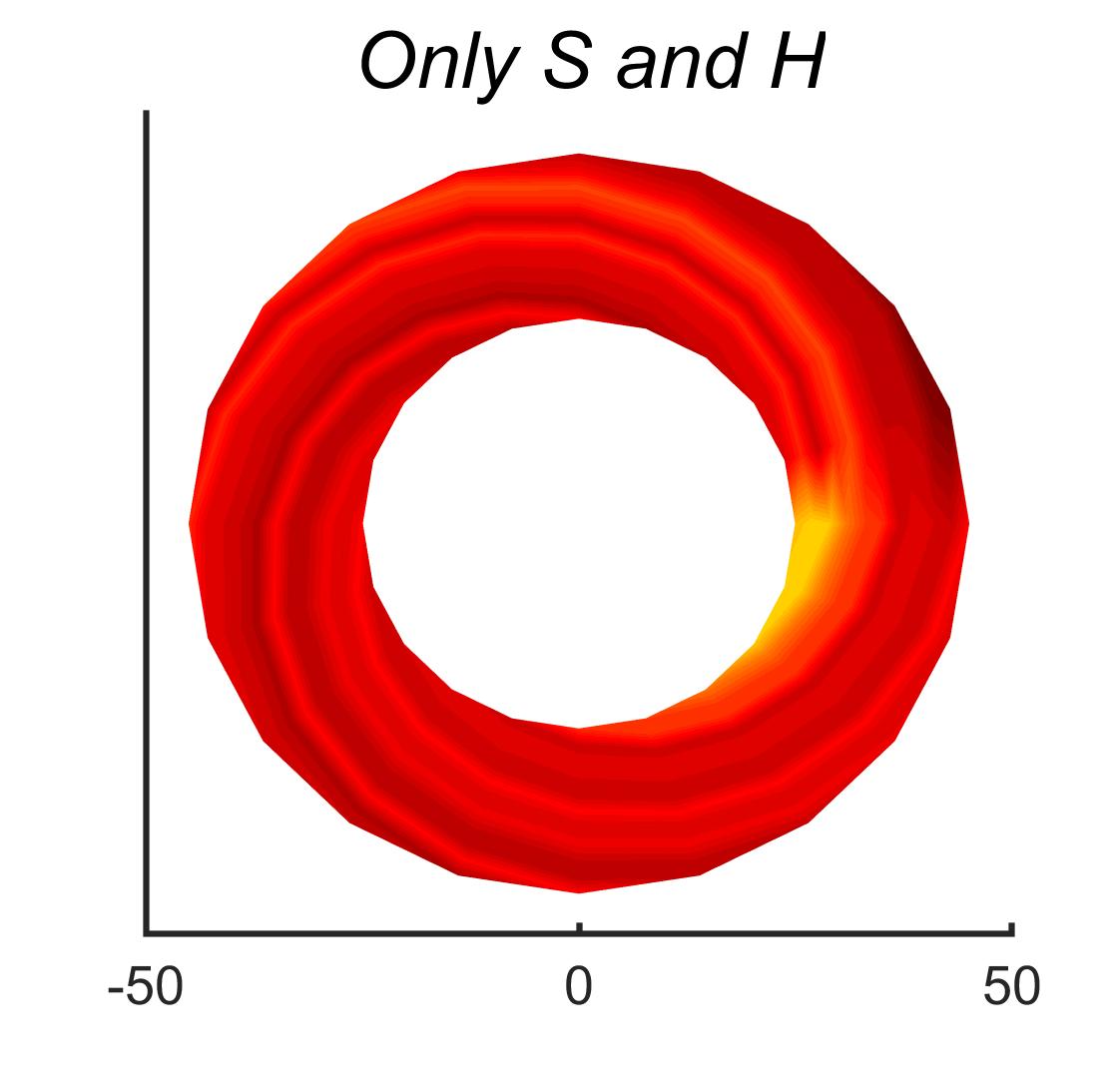}
		\end{subfigure}
		\begin{subfigure}{0.06\textwidth}
			\includegraphics[width=\textwidth]{other/colorbar.jpg}
		\end{subfigure}
		\caption{Accuracy of letter identification for the ILSVRC-initialised model with single flankers. We find that regardless of weight initialisation, crowding behaves similarly. Training and testing was done with acuity loss. Accuracy without flankers was 98.52\%.}
		\label{dimagenetsingle}
	\end{figure}

\begin{figure}
		\centering
		\begin{subfigure}{0.9\textwidth}
			\includegraphics[width=\textwidth]{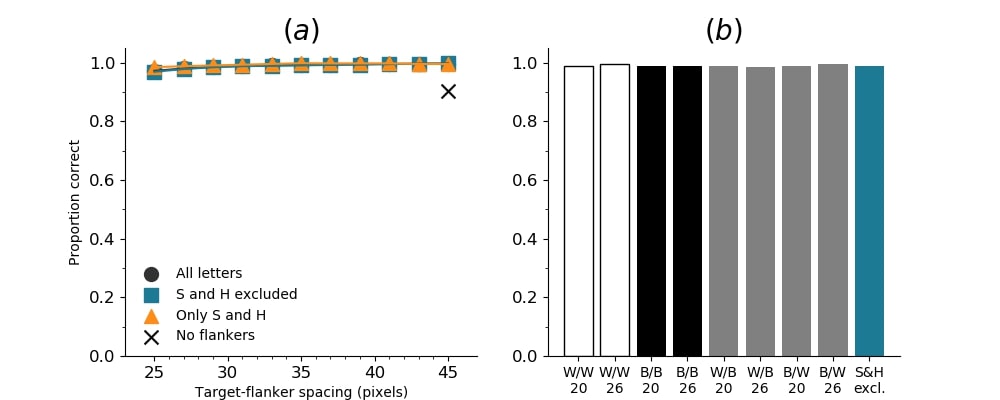}
		\end{subfigure}
		\begin{subfigure}{0.2562\textwidth}
			\includegraphics[width=\textwidth]{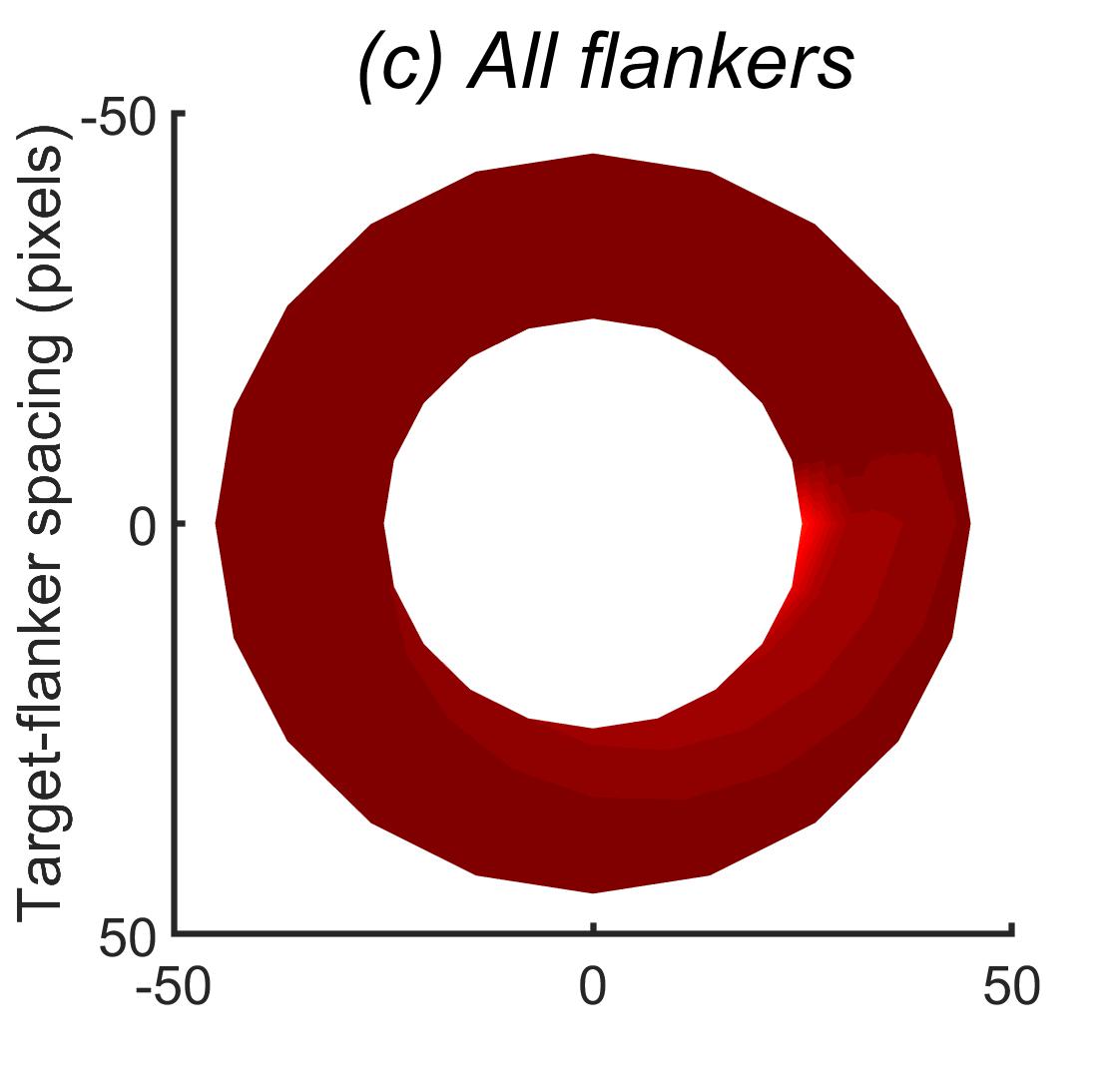}
		\end{subfigure}
		\begin{subfigure}{0.25\textwidth}
			\includegraphics[width=\textwidth]{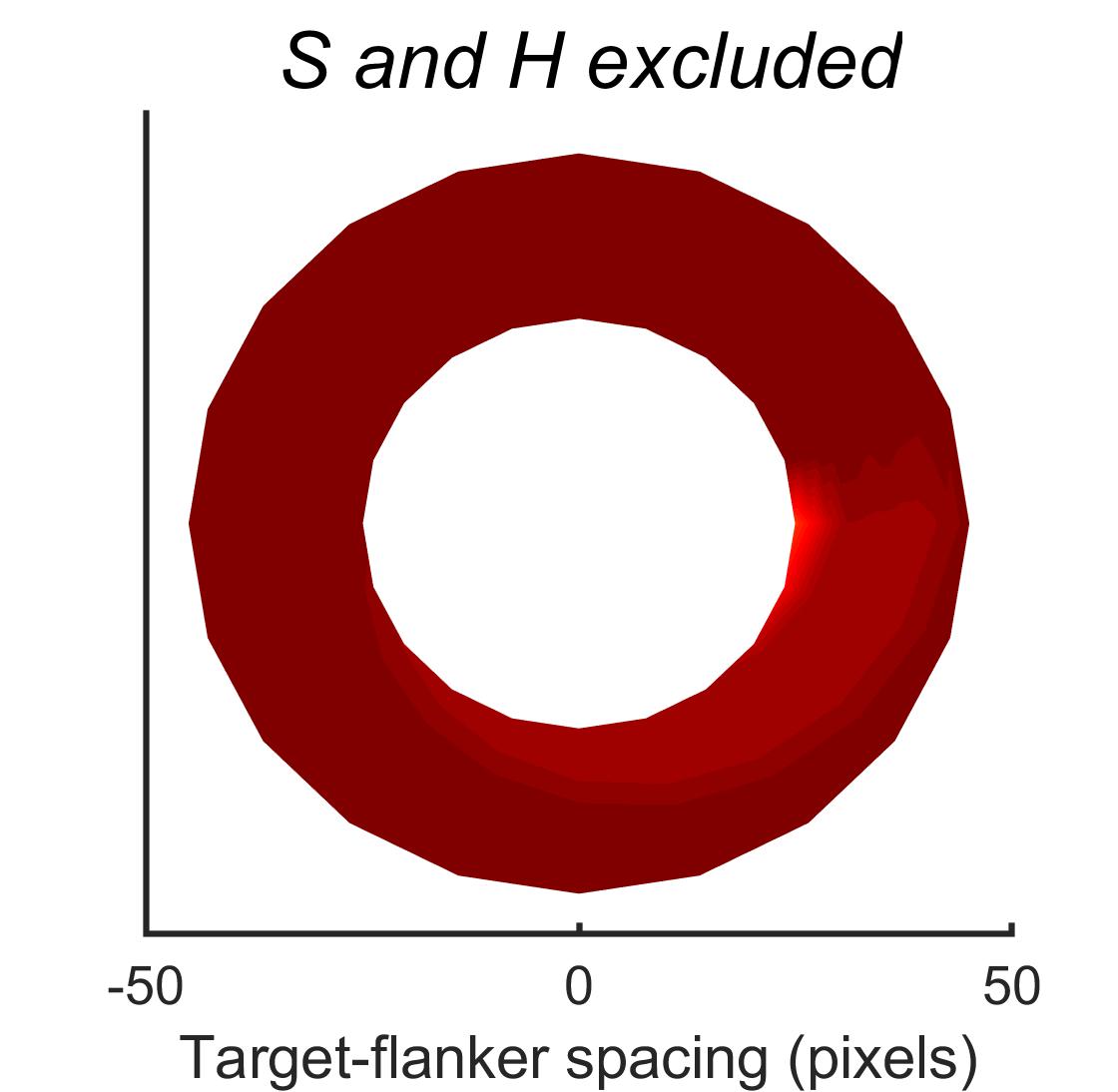}
		\end{subfigure}
		\begin{subfigure}{0.25\textwidth}
			\includegraphics[width=\textwidth]{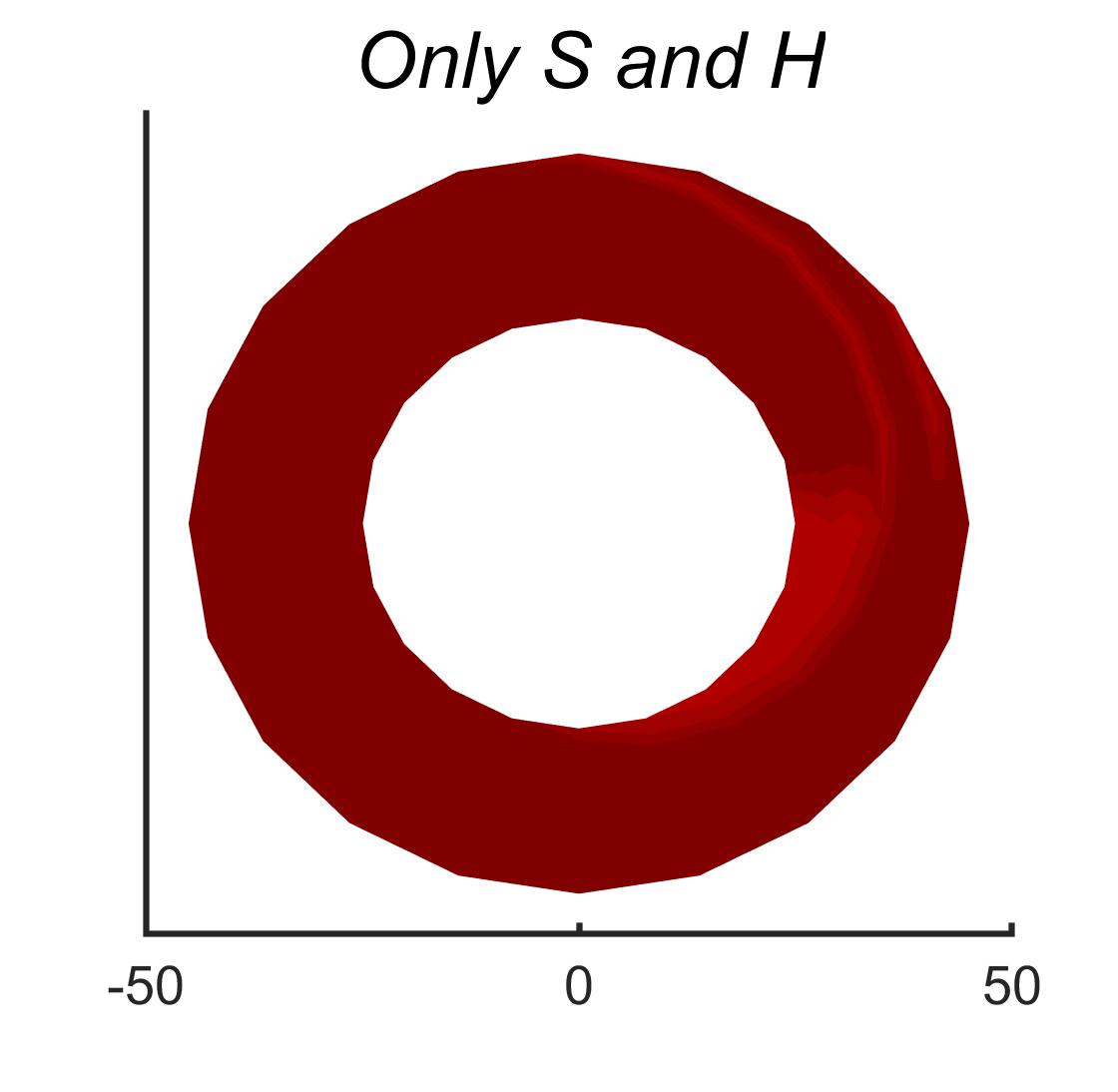}
		\end{subfigure}
		\begin{subfigure}{0.06\textwidth}
			\includegraphics[width=\textwidth]{other/colorbar.jpg}
		\end{subfigure}
		\caption{Accuracy of letter identification of the DenseNet-121 with random weight initialisation and no acuity loss with single flankers. Model base accuracy was 90.25\%. Note that accuracy is increased by adding a flanker---the only position that does not exhibit this behaviour is the same position that causes the most crowding in almost all of our other tests.}
		\label{dnoacuitylossrandominitsingle}
	\end{figure}

\begin{figure}
		\centering
		\begin{subfigure}{0.9\textwidth}
			\includegraphics[width=\textwidth]{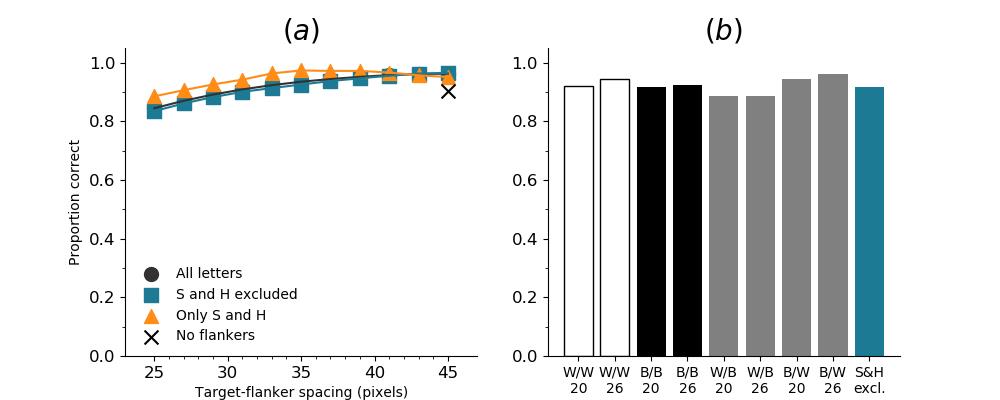}
		\end{subfigure}
		\begin{subfigure}{0.2562\textwidth}
			\includegraphics[width=\textwidth]{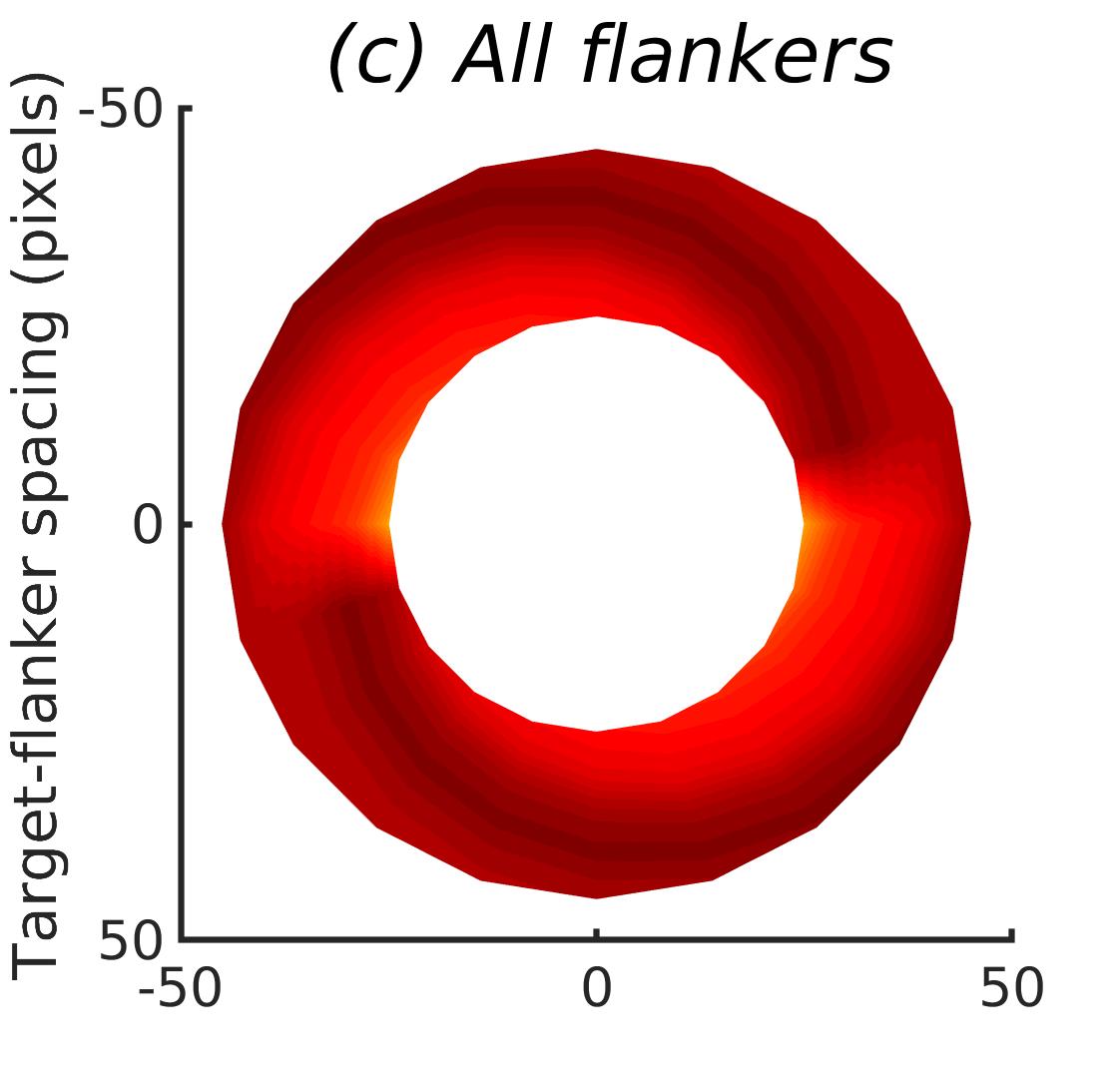}
		\end{subfigure}
		\begin{subfigure}{0.25\textwidth}
			\includegraphics[width=\textwidth]{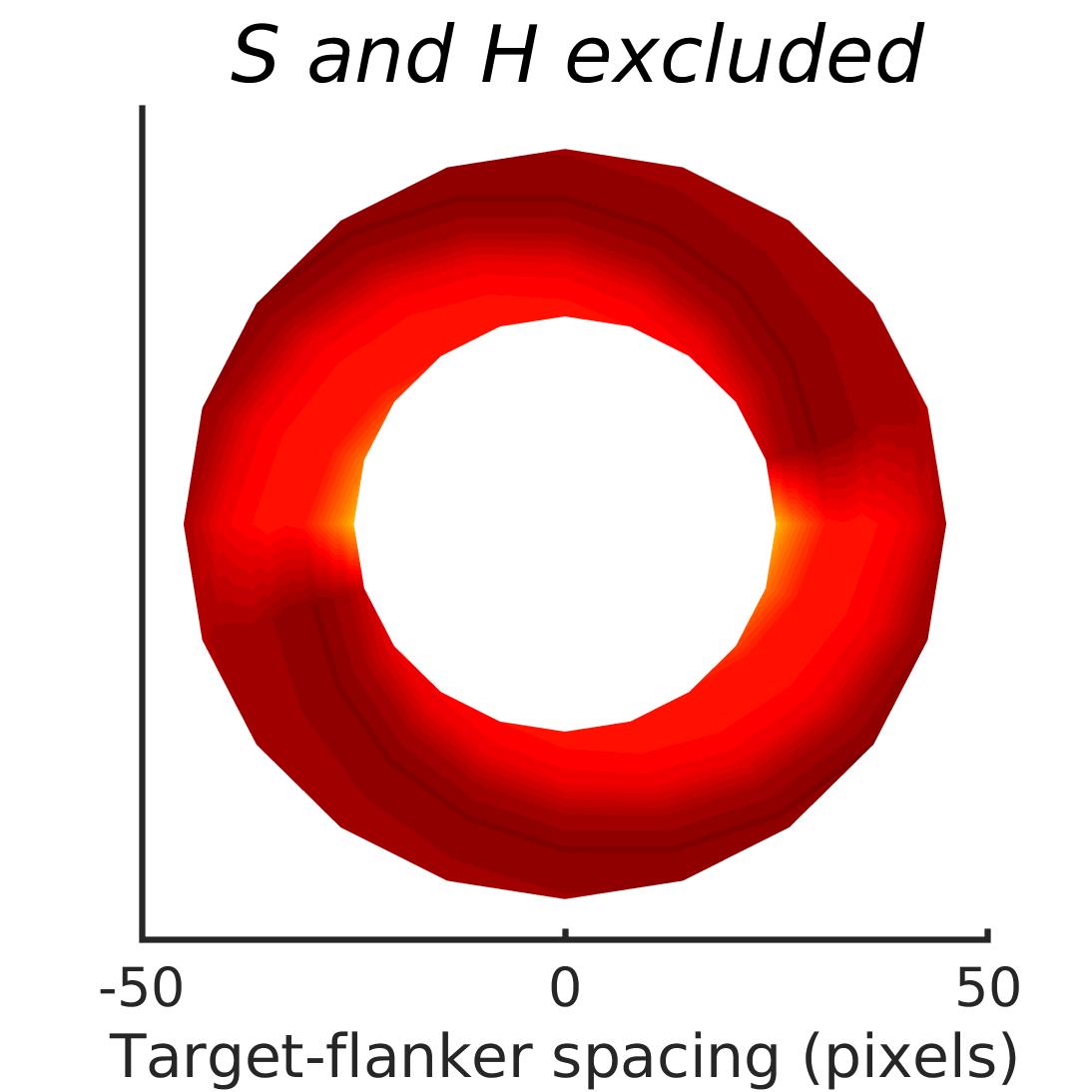}
		\end{subfigure}
		\begin{subfigure}{0.25\textwidth}
			\includegraphics[width=\textwidth]{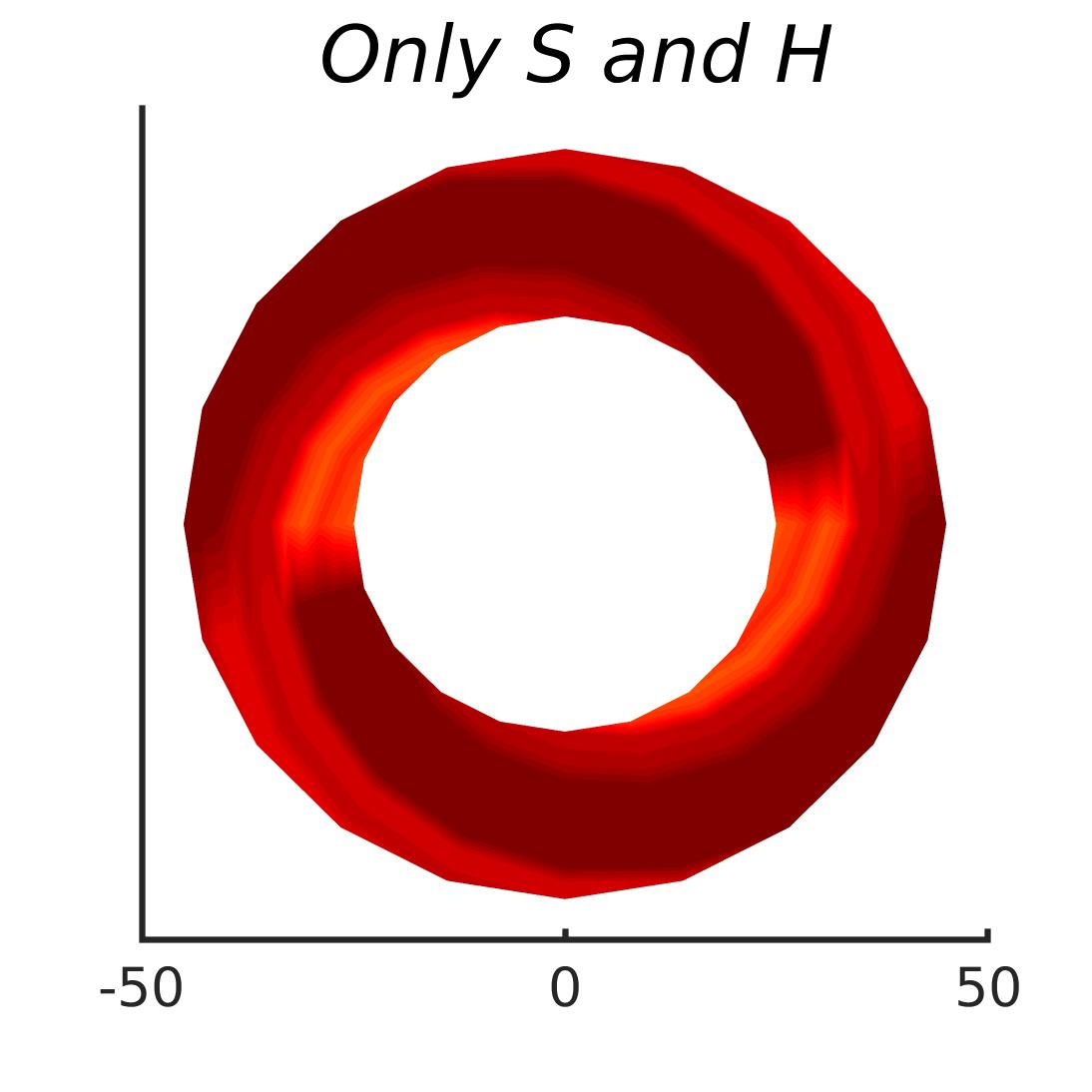}
		\end{subfigure}
		\begin{subfigure}{0.06\textwidth}
			\includegraphics[width=\textwidth]{other/colorbar.jpg}
		\end{subfigure}
		\caption{Accuracy of letter identification of the DenseNet-121 by distance and colour for pair flankers with random weight initialisation and no acuity loss. Accuracy without flankers was 90.25\%. Note that as this is the same model as presented in Figure \ref{dnoacuitylossrandominitsingle}---some positions of flankers also increase accuracy.}
		\label{dnoacuitylossrandominitpair}
	\end{figure}

	In Figures \ref{drandominit}-\ref{dflippedrandominitsingle}, and Supplementary Figures \ref{dimagenet}-\ref{dnoacuitylossrandominitpair}, panel \textit{(a)} plots accuracy as a function of target-flanker spacing in pixels, collapsed across all stimulus manipulations. In addition, we show unflanked accuracy, which is near-perfect for almost all experiments. Panel \textit{(b)} plots accuracy for each target-flanker colour combination for both sizes collapsed over all the target-flanker distances. For example, \textit{W/B 20} denotes a white target, a black flanker, with letter size 20 points. Additionally, collapsed data when the letters S and H are excluded is shown. Panel \textit{(c)} plots the shape of crowding---accuracy at each position of the flanker, where the origin of the plot is centred on the target, collapsed over all size and contrast polarity combinations. Accuracy is shown with all flankers, accuracy with the flankers S and H excluded, and accuracy using only the flankers S and H. Separated effects of the the letters S and H are shown as they were not a part of training, and therefore serve as 'novel' flankers.
	
	In general, letter recognition performance improved with target-flanker distance as expected from human studies, indicating that networks experience at least some form of crowding. However, unlike in humans, this trend was mild, and we also observed peculiar patterns in many of our experiments. We call these peculiar patterns \textit{anomalies of crowding}, or simply anomalies. An anomaly usually took the form of an unexpected change in performance (e.g. poor accuracy at large target-flanker spacing and better accuracy at short spacing for specific target-flanker configurations). These anomalies were found to be caused primarily by the untrained letters S and H as flankers. However, even after these letters were excluded from analysis, such anomalies persisted. Our findings suggest that only by training and testing a model several times, and by averaging results, can such anomalies be mitigated. Also unlike in humans, the current results showed a strong pattern of crowding along the top-left -- bottom-right diagonal in all tests with paired flankers. Interestingly, throughout our experiments no clear pattern of the effect of size or contrast polarity was found. Many models performed better for letter size 20 than for 26, but some exhibited the opposite behaviour. In humans, size has no effect on the strength or extent of crowding, and colour (or similarity) has a strong effect, with different colour flankers causing less crowding than same colour flankers (see Section \ref{introduction}).
	
	\subsection{SimpleNet with random initialisation}
	We trained five independent SimpleNets with unaltered images (no 'acuity loss') to test the sensitivity of small convolutional networks to different flanker configurations. The primary reason we tested SimpleNet was because of the claimed correspondence between such networks and the primate visual system \citep{cichy_dynamics_2017, guclu_deep_2015, kheradpisheh_deep_2016}. If the two architecture (DCNNs and biological systems) achieve object recognition in comparable ways, then we should see evidence of human-like crowding in SimpleNet. We found that while the networks learned to perform the letter identification task with high accuracy, they suffered greatly from flanker presence. In other words, such networks do suffer from crowding: flankers substantially degrade target identification performance. However, there are noticeable differences between the crowding observed in humans and in the SimpleNets. 
	\begin{figure}
		\centering
		\begin{subfigure}{0.9\textwidth}
			\includegraphics[width=\textwidth]{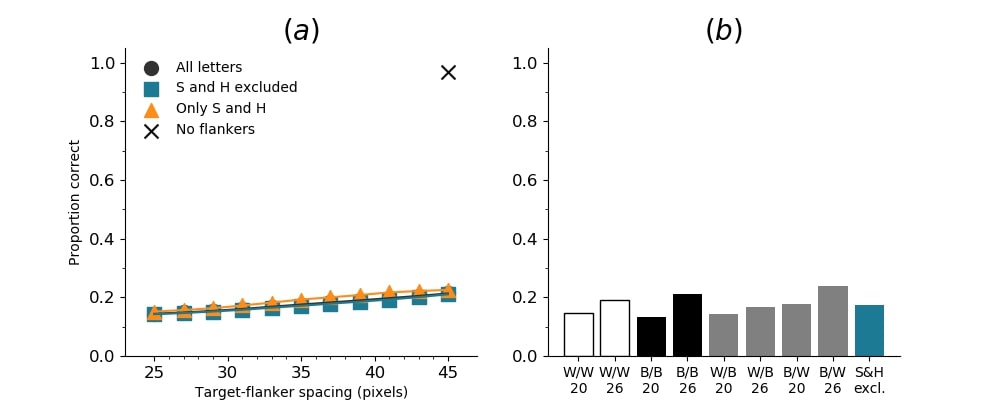}
		\end{subfigure}
		\begin{subfigure}{0.2562\textwidth}
			\includegraphics[width=\textwidth]{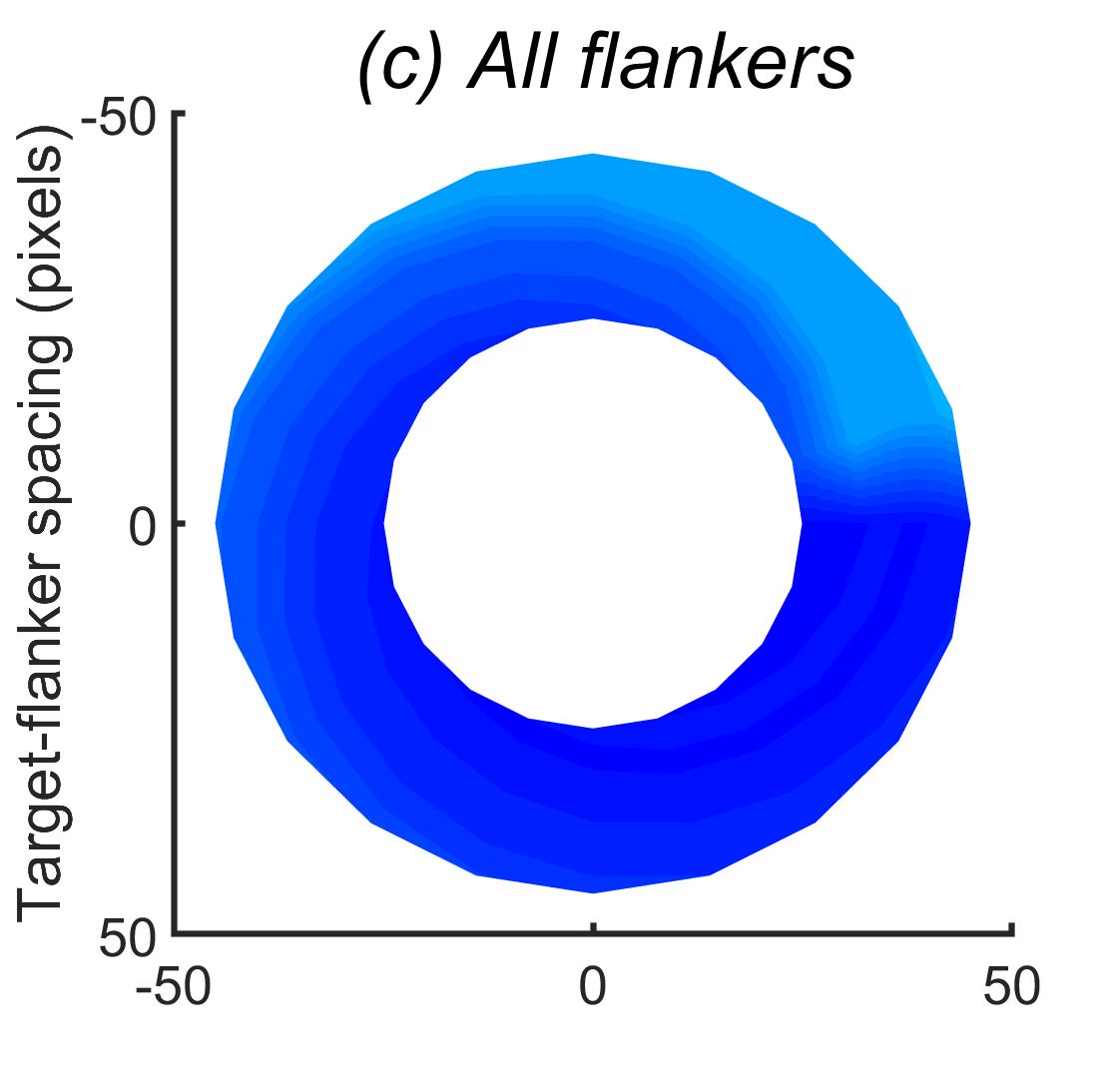}
		\end{subfigure}
		\begin{subfigure}{0.25\textwidth}
			\includegraphics[width=\textwidth]{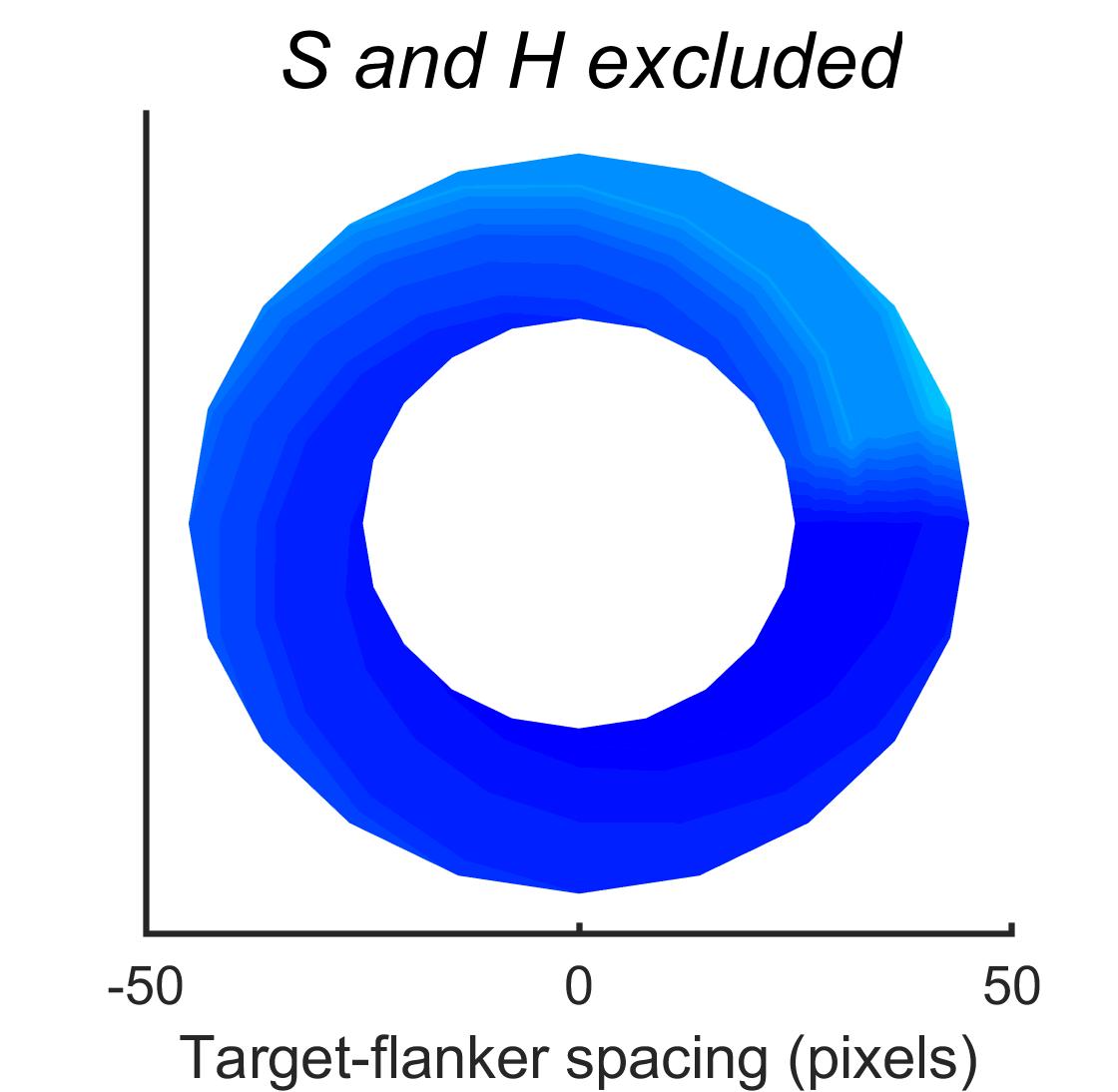}
		\end{subfigure}
		\begin{subfigure}{0.25\textwidth}
			\includegraphics[width=\textwidth]{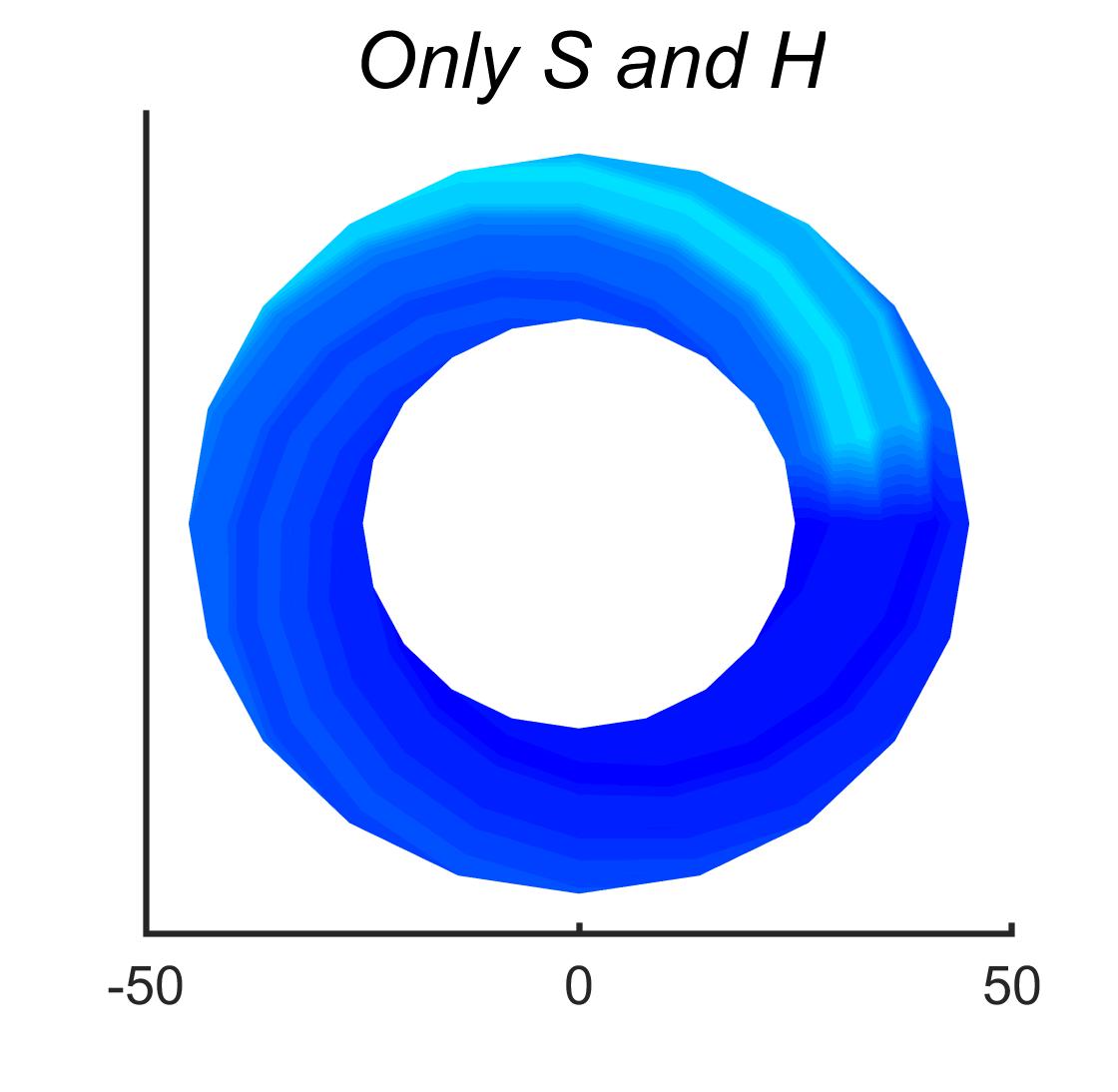}
		\end{subfigure}
		\begin{subfigure}{0.06\textwidth}
			\includegraphics[width=\textwidth]{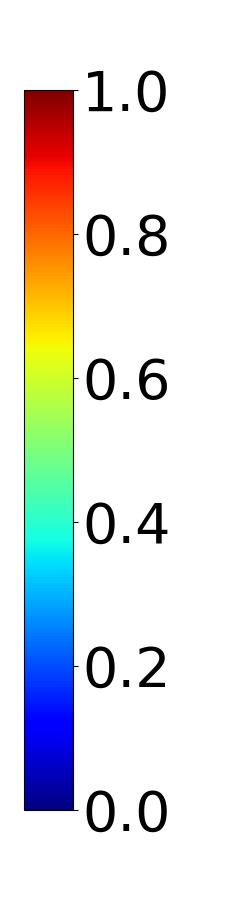}
		\end{subfigure}
		\caption{Accuracy of letter identification of the randomly initialised small 5-layer convolutional network with single flankers ('SimpleNet'). A total of five independent training sessions and test sessions are averaged in this figure. Results of individual runs are shown in Supplementary Figures \ref{smallrandominit1}-\ref{smallrandominit5}. Training and testing of all five models in this figure was done without unaltered images. Average accuracy without flankers was 96.97, as shown in panel \textit{(a)}\%. The letters S and H are excluded in \textit{(c) middle} and shown exclusively in \textit{(c) right}, as the network had not seen the letters in testing. We also tested the SimpleNets with acuity-reduced stimuli---Figure \ref{smallrandominitacuitylossconcat} shows the averaged results, and Figures \ref{smallrandominitacuityloss1}-\ref{smallrandominitacuityloss5} show individual models' results. We found that accuracy reached near-chance levels (12.5\%), and as such little can be inferred from these results.}
		\label{smallrandominitconcat}
	\end{figure}
	
	Unflanked targets were identified with high accuracy (96.97\%), but the presence of a single flanker even at at a large distance from the flanker reduced performance substantially (flanked performance was 35\% or lower). In contrast, crowding in humans is quite weak in the presence of a single flanker and is rapidly alleviated with spacing between the target and the flanker \citep{petrov_asymmetries_2011}. However, for SimpleNet, the overall reduction was dramatic with hardly any improvement with spacing. In fact, extrapolating from the data, the target-flanker distance at which there would be no crowding (where performance is the same as in the unflanked condition) would be 218 pixels, which is approximately the entire width of the image. That is, the model is strongly crowded at almost all distances. In addition, when the flankers presented to the model were untrained (the letters S and H), the pattern of crowding became unpredictable; at certain angular locations, flankers further away caused more crowding than those closer. These are examples of anomalies of crowding, described above. This effect does not occur in humans \citep{huckauf_lateral_1999}\footnote{However, as mentioned above, we believe that it is possible that such anomalies disappear entirely when a model is trained a large number of times and the results averaged.}. These results indicate that DCNNs suffer from crowding in the periphery, that they suffer from crowding up to a much greater distance than humans, and that the effect of target-flanker spacing is weaker than in humans, at least in the range of distances we tested. We attempted to fit psychometric curves (see the Appendix, Figure \ref{psychometric}), but the fits were unsuccessful in producing meaningful results, primarily due to the anomalies and a lack of a clear upper asymptote. We also note that all five instantiations of SimpleNet displayed the same pattern of crowding, indicating that these findings were reproducible and not an artefact of the initial settings.
	
	\subsection{VGG-16 with ILSVRC initialisation}
\begin{figure}
		\centering
		\begin{subfigure}{0.9\textwidth}
			\includegraphics[width=\textwidth]{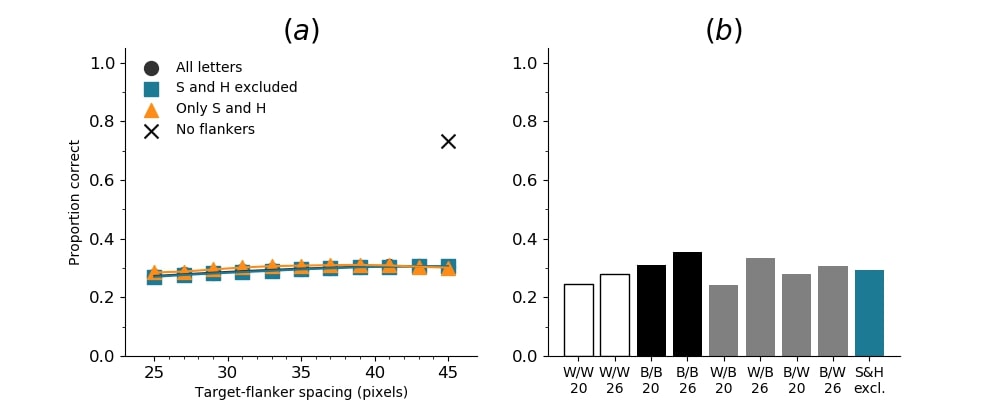}
		\end{subfigure}
		\begin{subfigure}{0.2562\textwidth}
			\includegraphics[width=\textwidth]{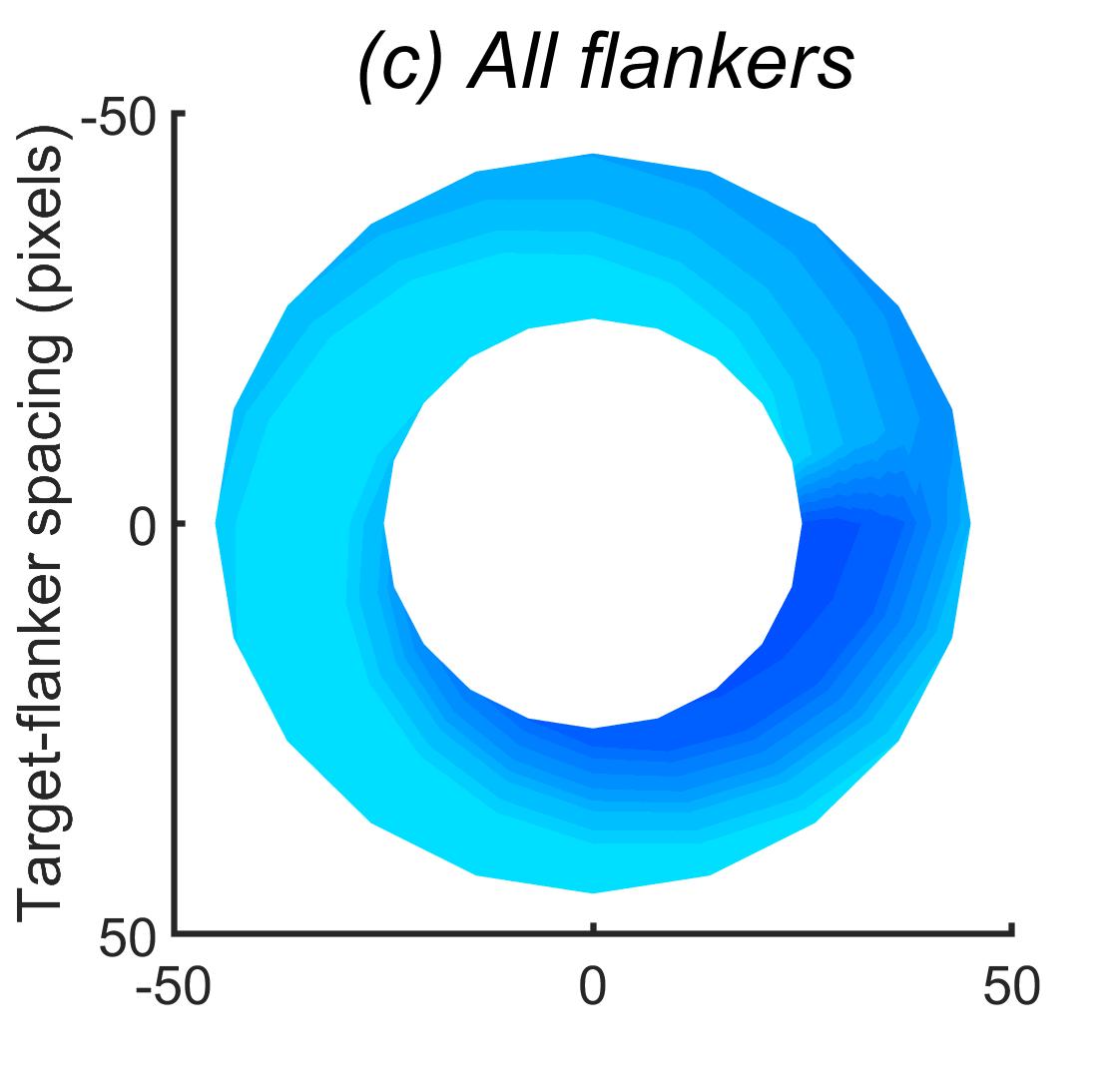}
		\end{subfigure}
		\begin{subfigure}{0.25\textwidth}
			\includegraphics[width=\textwidth]{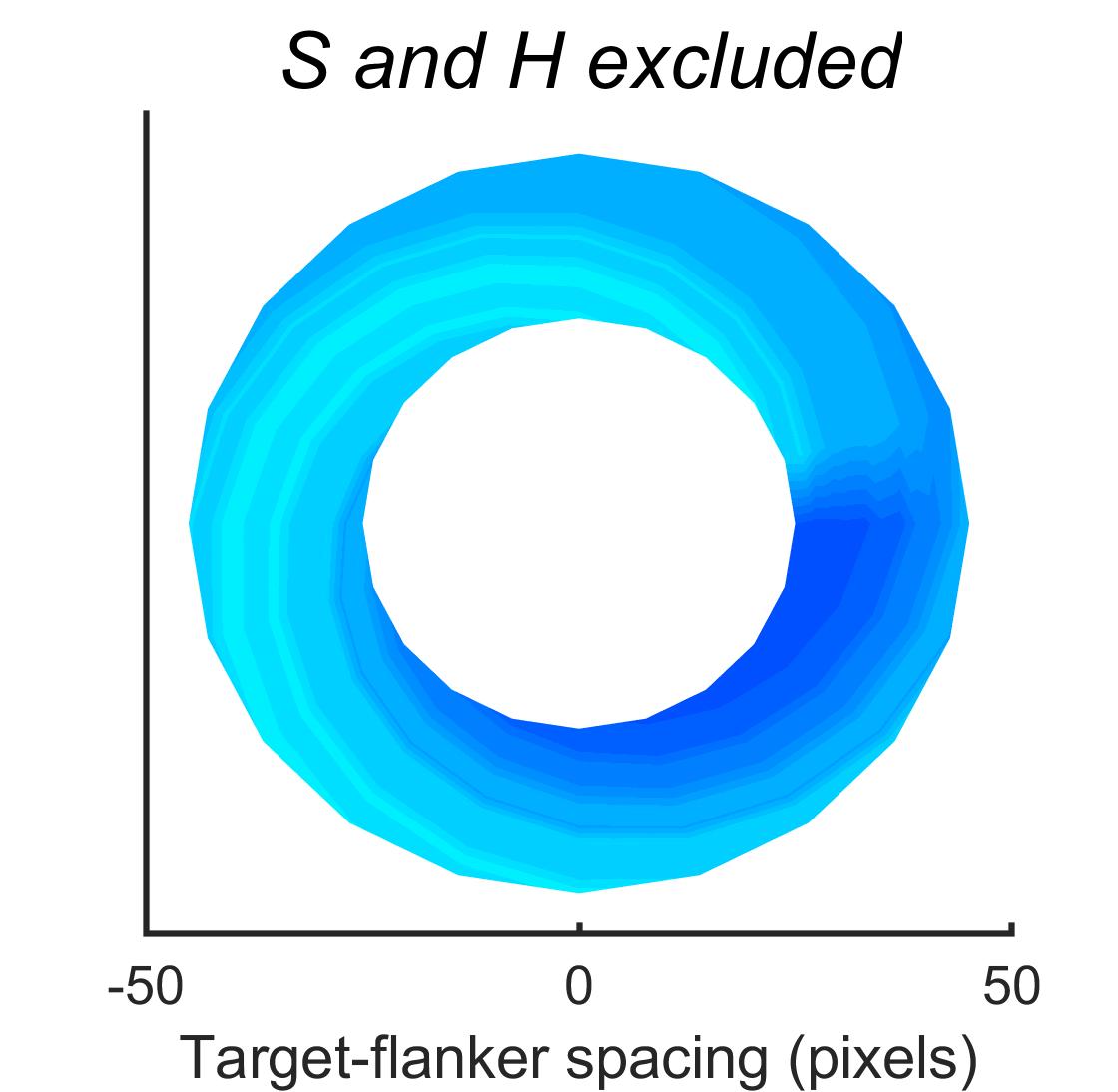}
		\end{subfigure}
		\begin{subfigure}{0.25\textwidth}
			\includegraphics[width=\textwidth]{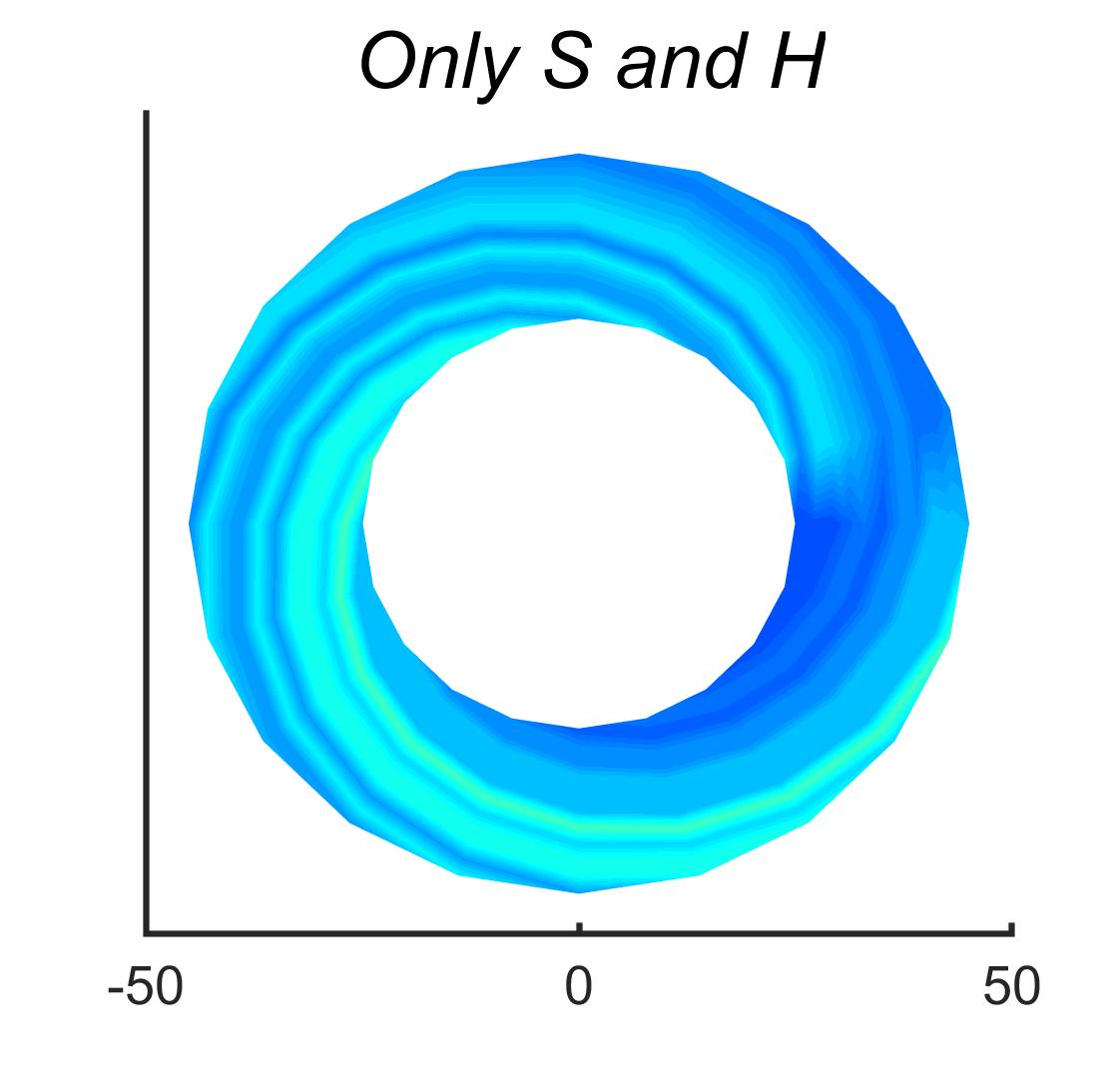}
		\end{subfigure}
		\begin{subfigure}{0.06\textwidth}
			\includegraphics[width=\textwidth]{other/colorbar.jpg}
		\end{subfigure}
		\caption{Accuracy of letter identification for the ILSVRC-initialised VGG-16 model with single flankers. Even though the accuracy for unflanked targets in the VGG-16 was lower than for the SimpleNets, it appears to be more robust to clutter; that is, flanked performance is higher. Nevertheless, the general pattern of crowding remains the same. VGG-16 accuracy without flankers was 73.46\%.}
		\label{vggimagenet}
	\end{figure}

	We also trained a different architecture of network, the VGG-16 \citep{simonyan_very_2014}, to test whether our results are specific to the SimpleNet (and to the DenseNet-121, see below) architecture. We found that while the VGG-16 performed somewhat better in our task (Figure \ref{vggimagenet}), it exhibited the same general patterns and behaviour of crowding as the SimpleNet. This implies that the presence and characteristics of crowding, and by implication, object recognition in DCNNs, is a property of the basic building blocks of DCNNs and not caused by a particular network architecture. Note that in our study VGG-16 was initialised in a completely different way compared to the SimpleNets. Yet, the pattern was the same, with slightly better robustness to flankers\footnote{We attempted two runs of this model, but only one converged.}. The VGG-16 with ILSVRC initialisation can be considered to be more 'experienced' with visual stimuli. However, both VGG-16 and our SimpleNet were highly sensitive to the presence of clutter and were insensitive to a large extent to the spacing between the target and the clutter.
	
	\subsection{DenseNet-121 with random initialisation}
	The previous two architectures of models we have tested contained only simple combinations of convolutional, max-pooling, and densely connected layers. To test whether our results are specific to such configurations, or apply more generally to more sophisticated architectures, we tested the DenseNet-121 \citep{huang_densely_2016}, a recent architecture that takes advantage of batch normalisation and skip connections. Further, as noted above, DenseNets and ResNest (from which DenseNets are derived) have been argued to have a higher correspondance to the primate visual system than earlier networks such as AlexNet and VGG \citep{SchrimpfKubilius2018BrainScore, nkriegeskorte_coarser_2017}. We trained and tested the DenseNet network extensively under various network and stimulus configurations in order to assess if a highly successful model suffers from crowding and if this crowding is comparable to that in humans, given the claimed correspondence.
	
\begin{figure}
		\centering
		\begin{subfigure}{0.9\textwidth}
			\includegraphics[width=\textwidth]{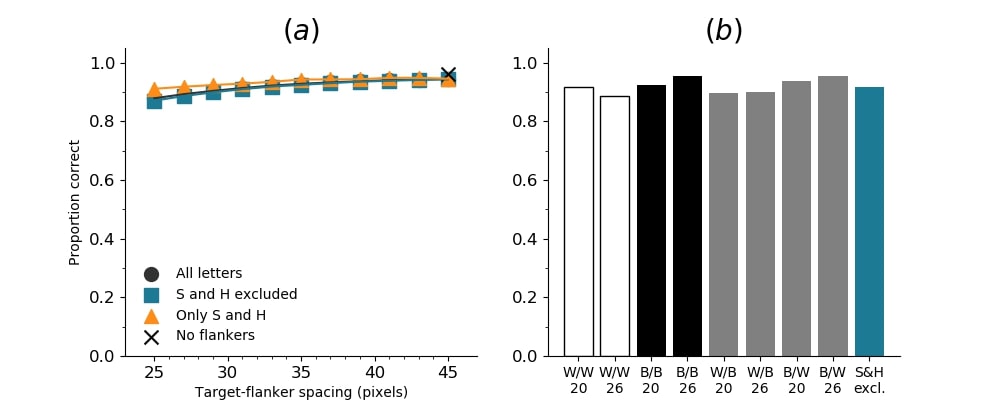}
		\end{subfigure}
		\begin{subfigure}{0.2562\textwidth}
			\includegraphics[width=\textwidth]{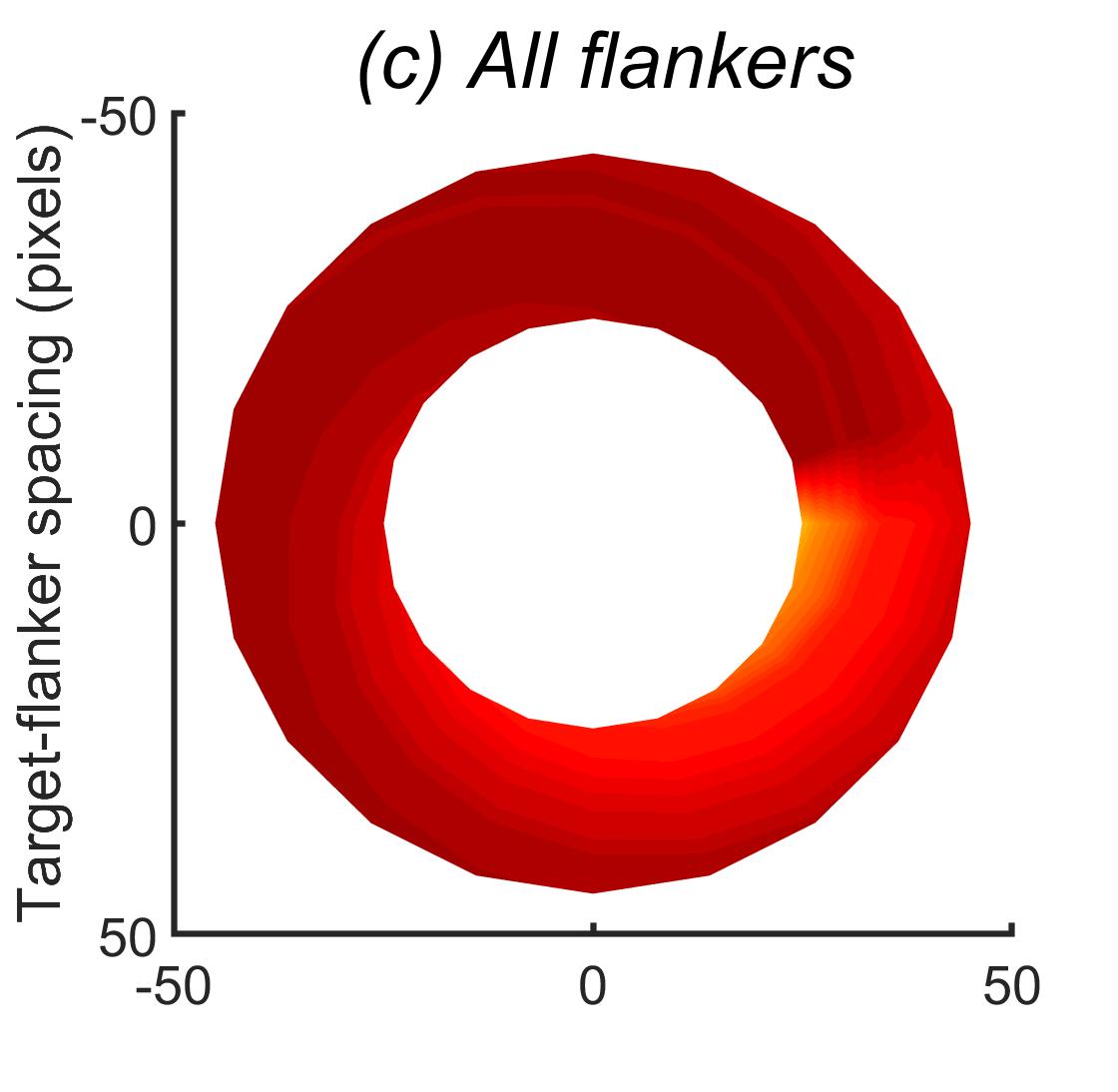}
		\end{subfigure}
		\begin{subfigure}{0.25\textwidth}
			\includegraphics[width=\textwidth]{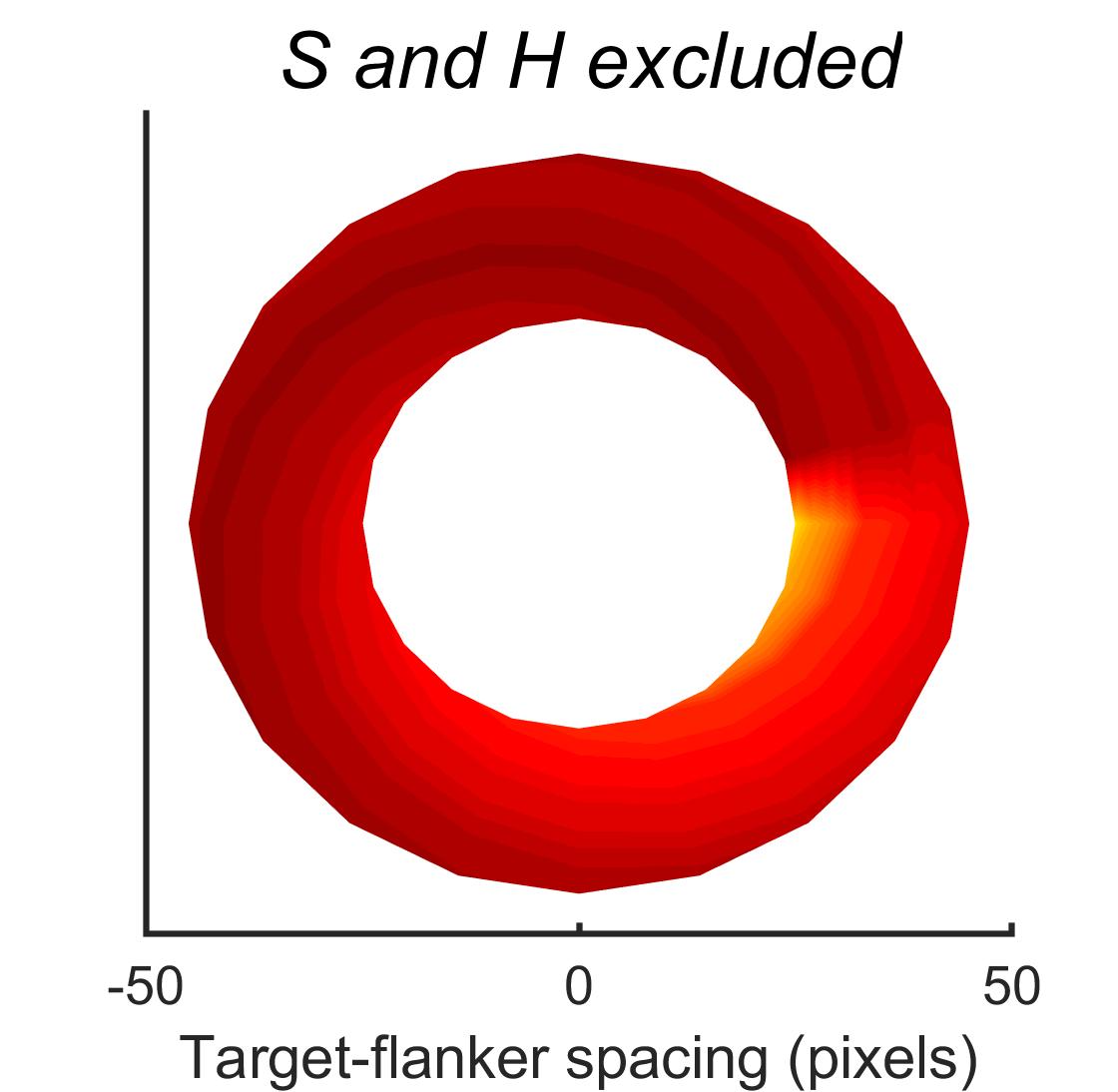}
		\end{subfigure}
		\begin{subfigure}{0.25\textwidth}
			\includegraphics[width=\textwidth]{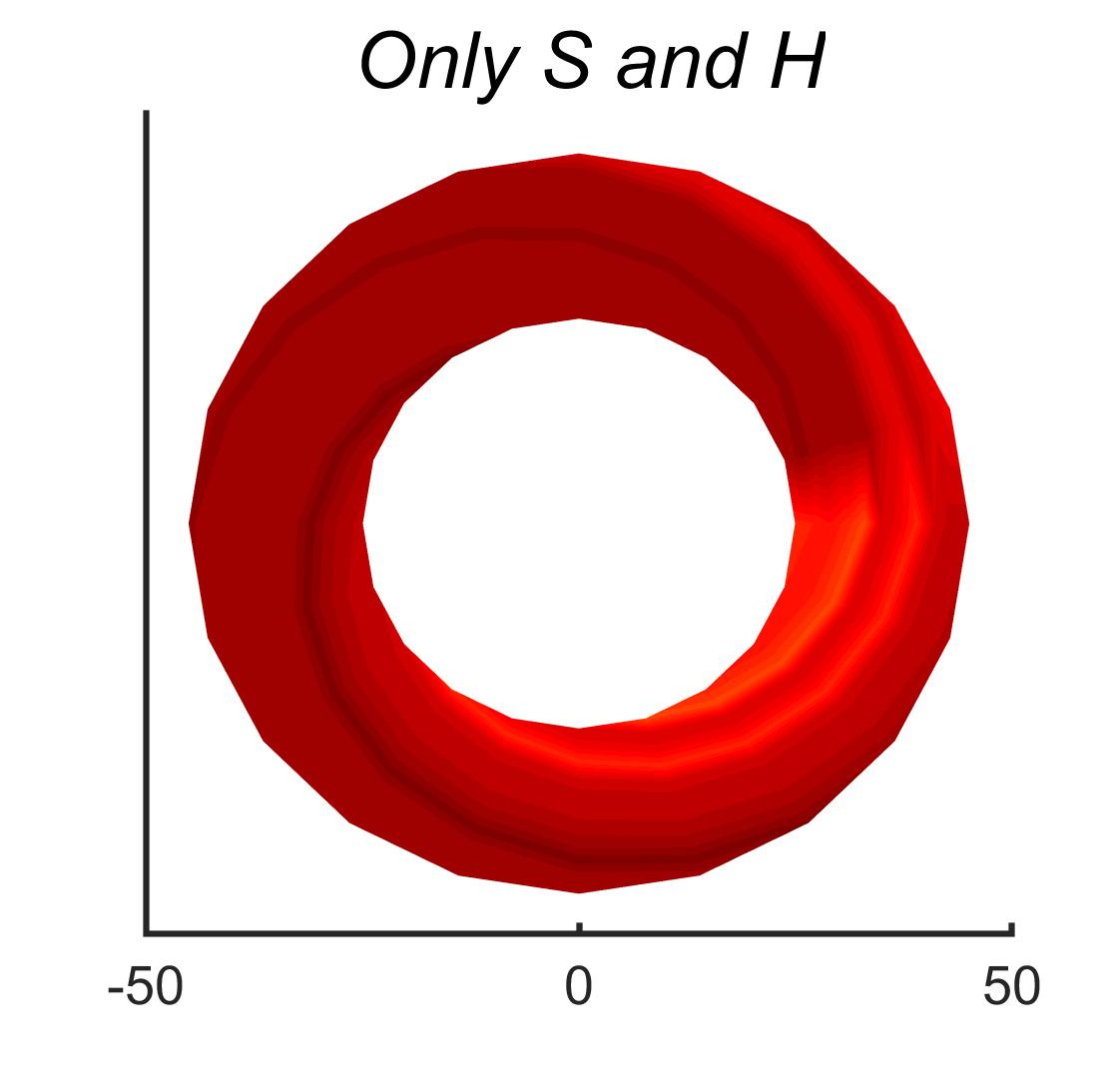}
		\end{subfigure}
		\begin{subfigure}{0.06\textwidth}
			\includegraphics[width=\textwidth]{other/colorbar.jpg}
		\end{subfigure}
		\caption{Accuracy of letter identification of the randomly initialised DenseNet-121 in the presence of single flankers. Training and testing was done with acuity loss. The DenseNet-121 is much more robust to the presence of flankers than the VGG-16 or SimpleNet models. Accuracy without flankers was 96.11\%.}
		\label{drandominitsingle}
	\end{figure}

	Figure \ref{drandominitsingle} shows the results when the network was initialised with random weights and stimuli were degraded to match perceptual input to the human visual system. We found that the DenseNet-121 is much more robust to clutter than the VGG-16 or the SimpleNet. The presence of a single flanker reduces target identification performance, but the drop is not dramatic. Further, increasing the spacing of the flanker from the target ameliorates crowding to the extent that far flankers do not interfere with target identification. The performance of this model is reminiscent of human performance. Interestingly, however, as with the SimpleNet and VGG-16, the strongest interference by a single flanker is not where its acuity is the lowest (the outermost position on the left of the background image along the horizontal meridian), but instead remains on the bottom-diagonal of the target towards the centre of the image. Note that in humans, the strongest interference is observed when the flanker is placed at this outermost location, and not by a flanker closer to the centre of the image \citep{petrov_asymmetries_2011}.
	
	\begin{figure}
		\centering
		\begin{subfigure}{0.9\textwidth}
			\includegraphics[width=\textwidth]{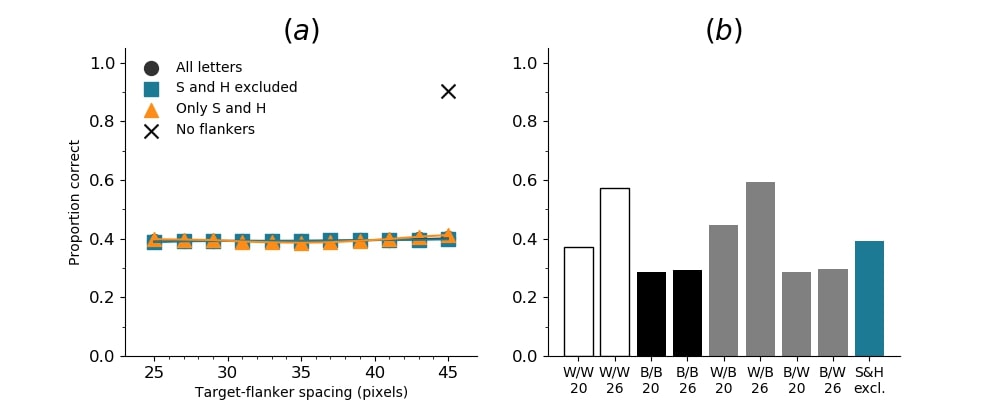}
		\end{subfigure}
		\begin{subfigure}{0.2562\textwidth}
			\includegraphics[width=\textwidth]{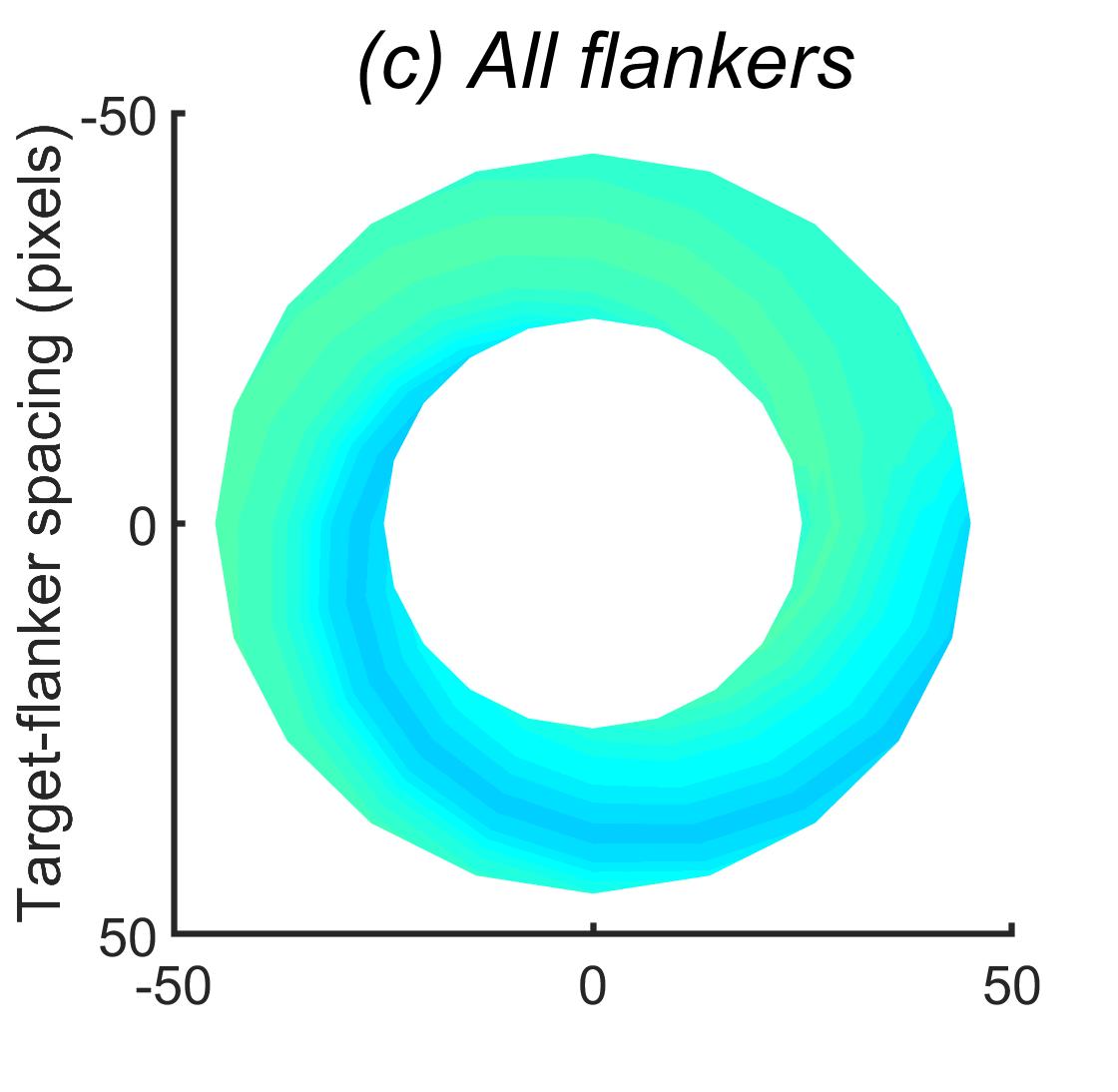}
		\end{subfigure}
		\begin{subfigure}{0.25\textwidth}
			\includegraphics[width=\textwidth]{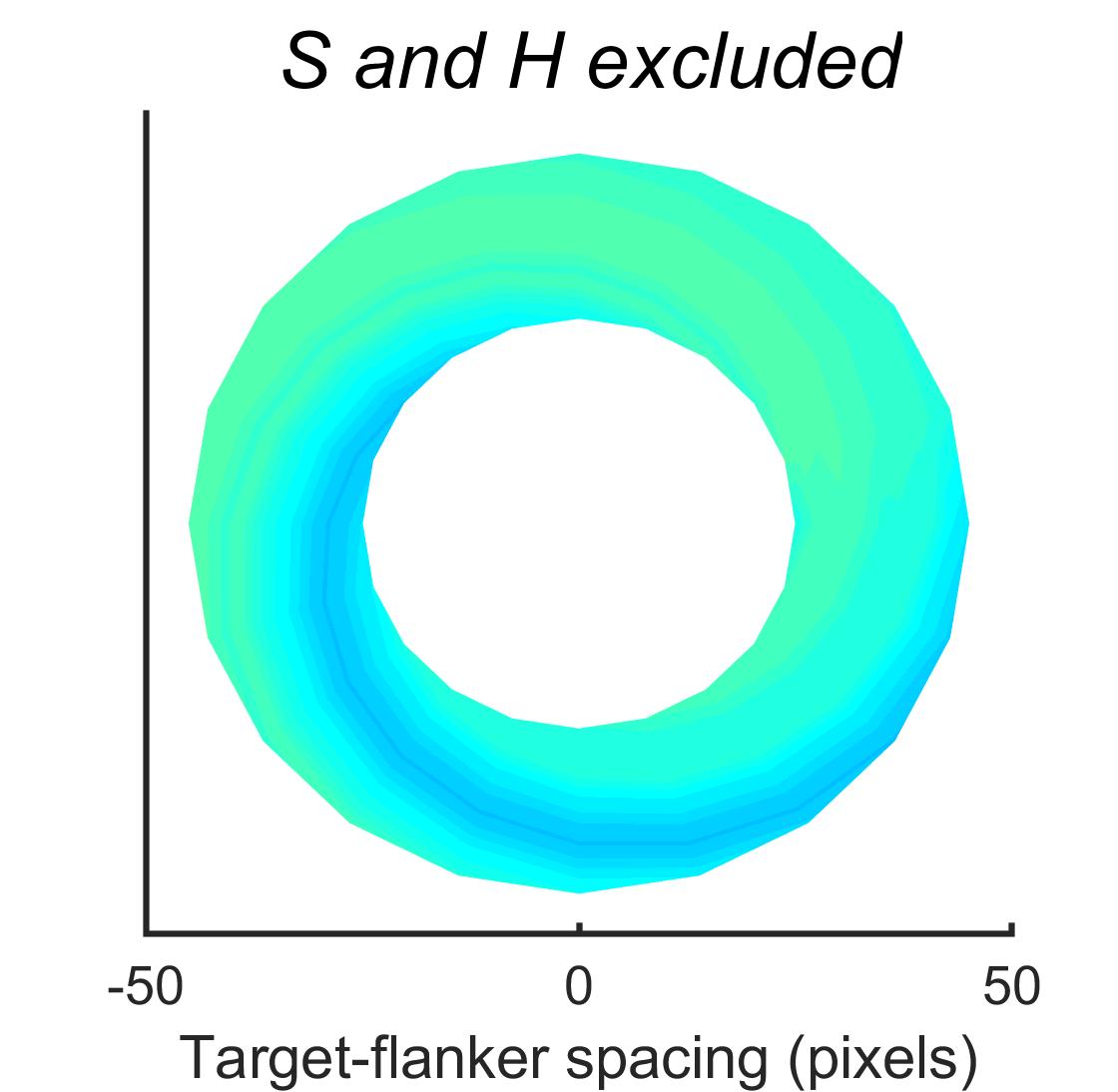}
		\end{subfigure}
		\begin{subfigure}{0.25\textwidth}
			\includegraphics[width=\textwidth]{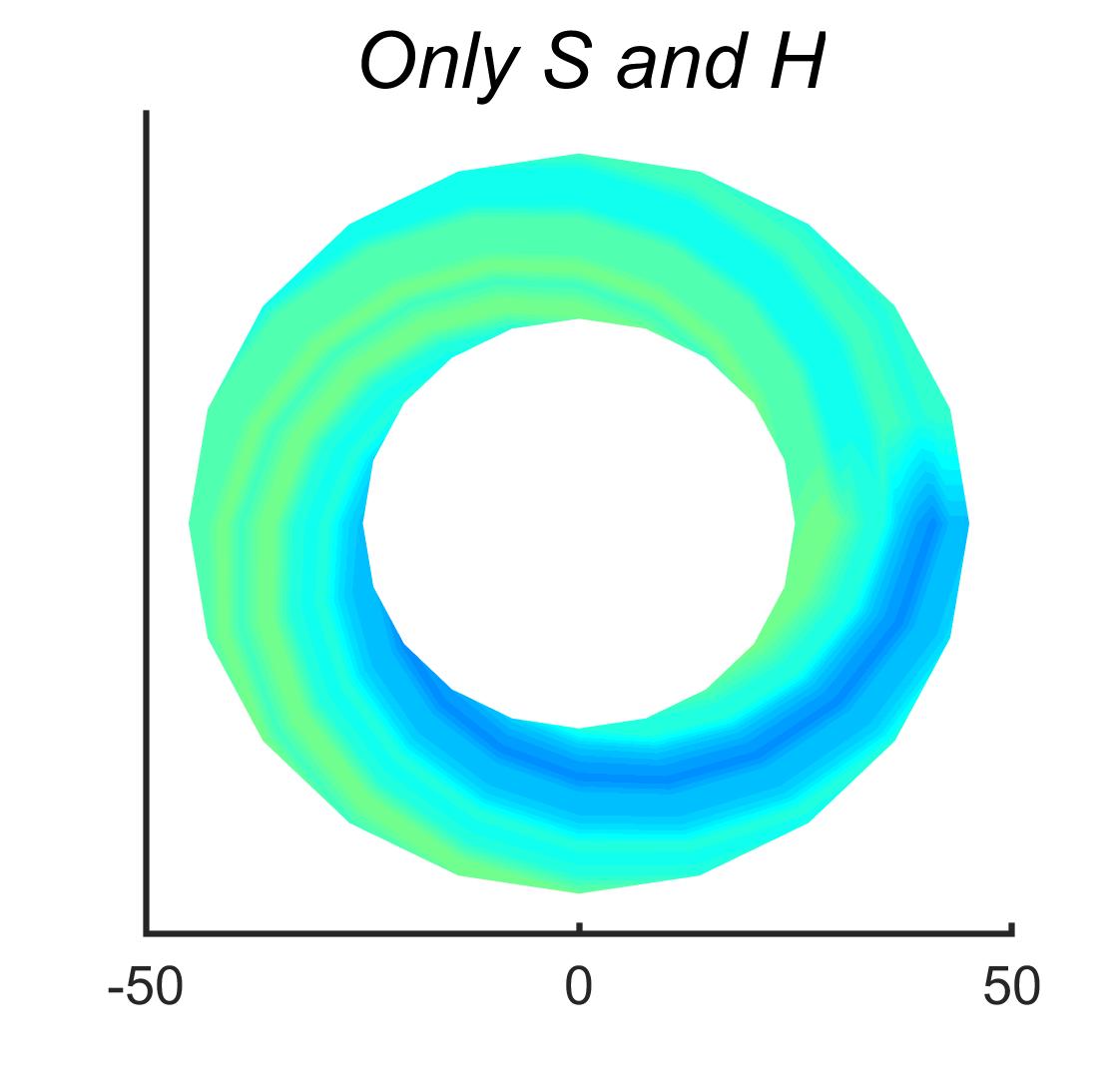}
		\end{subfigure}
		\begin{subfigure}{0.06\textwidth}
			\includegraphics[width=\textwidth]{other/colorbar.jpg}
		\end{subfigure}
		\caption{Accuracy of letter identification for the DenseNet-121 with single flankers when the target and flankers are placed on the right-hand side of the image, instead of the left-hand side. Training and testing was done with acuity loss. The area of most crowding on average shifts to the left-hand side of the target, towards the centre of the image, showing evidence that a higher acuity flanker will crowd the target more than a lower acuity flanker. Additionally, distance has little effect on crowding. Model accuracy without flankers was 90.37\%.}
		\label{drandominitsingleright}
	\end{figure}
	
\begin{figure}
		\centering
		\begin{subfigure}{0.9\textwidth}
			\includegraphics[width=\textwidth]{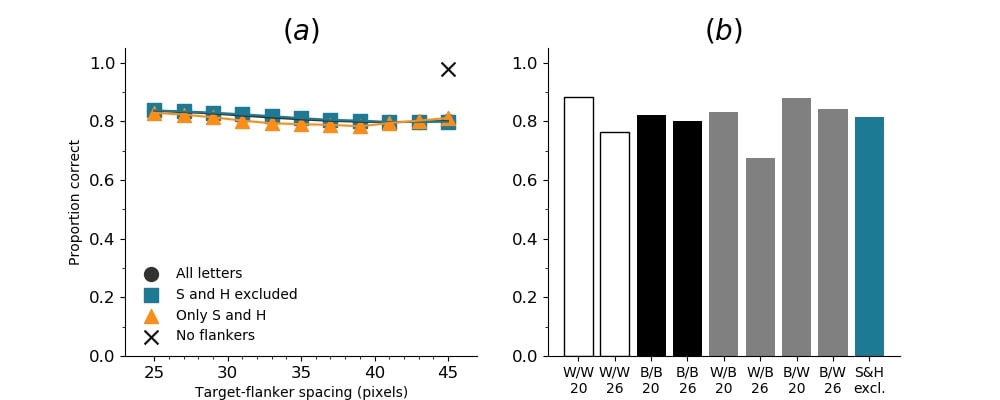}
		\end{subfigure}
		\begin{subfigure}{0.2562\textwidth}
			\includegraphics[width=\textwidth]{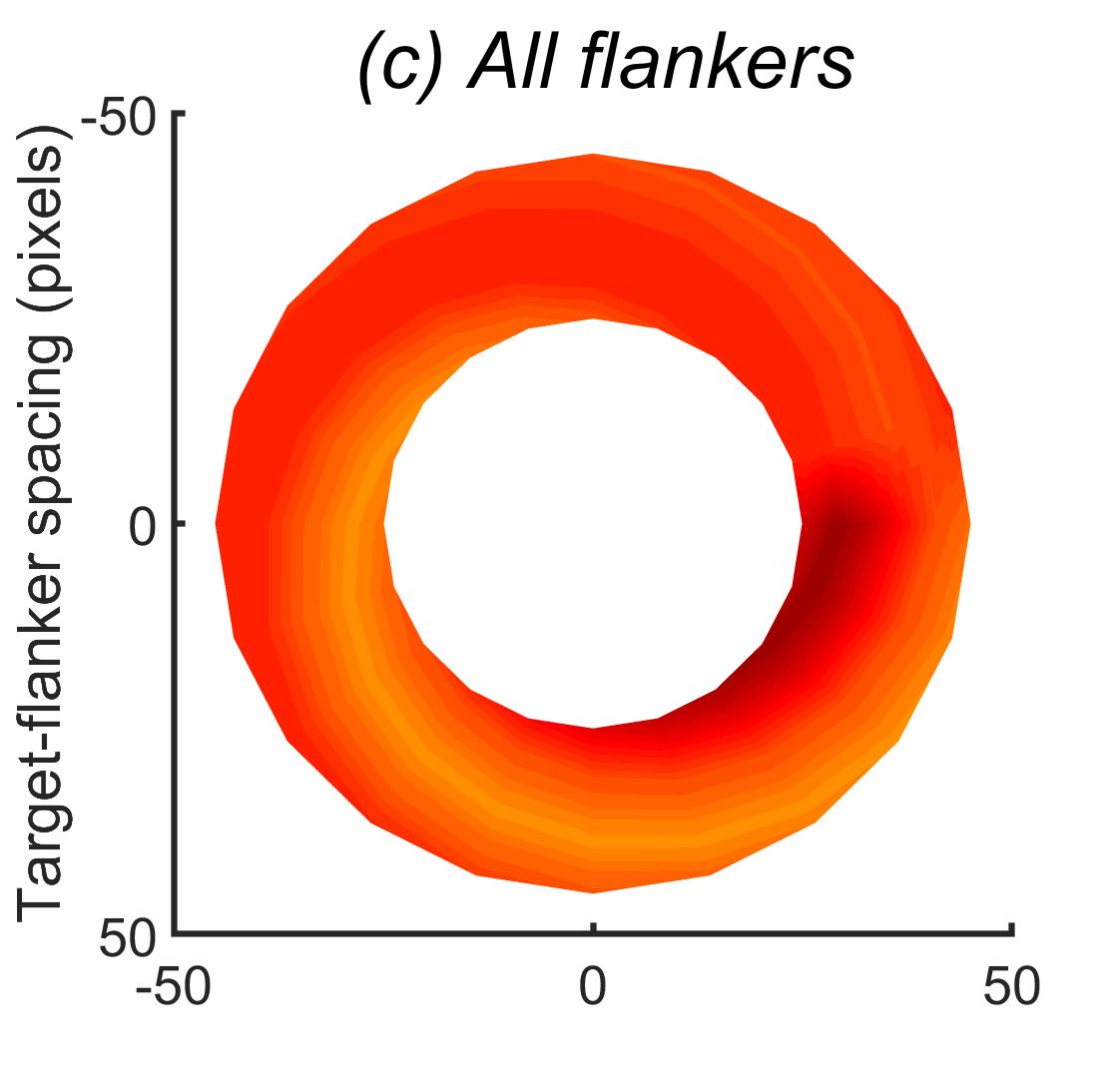}
		\end{subfigure}
		\begin{subfigure}{0.25\textwidth}
			\includegraphics[width=\textwidth]{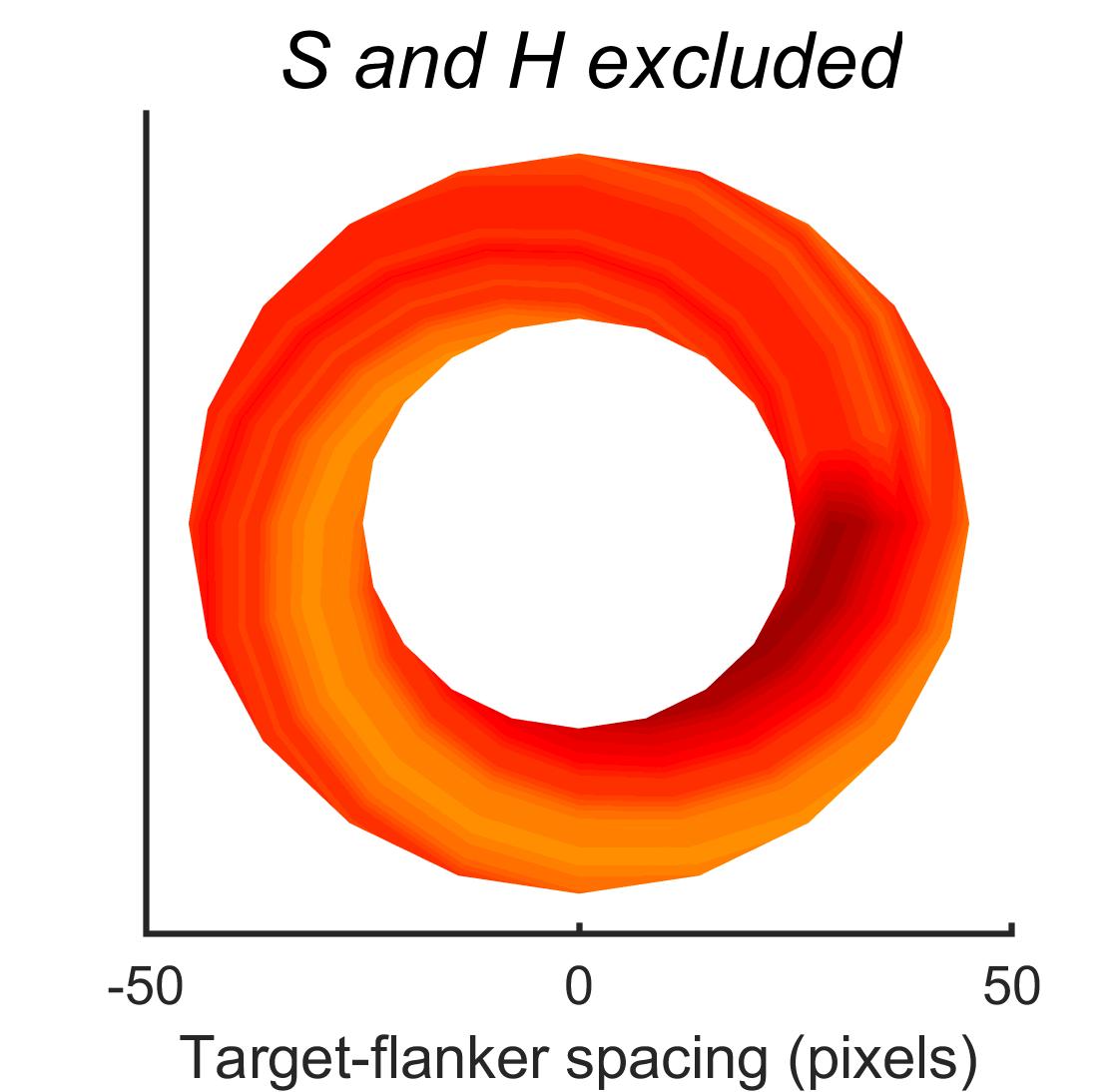}
		\end{subfigure}
		\begin{subfigure}{0.25\textwidth}
			\includegraphics[width=\textwidth]{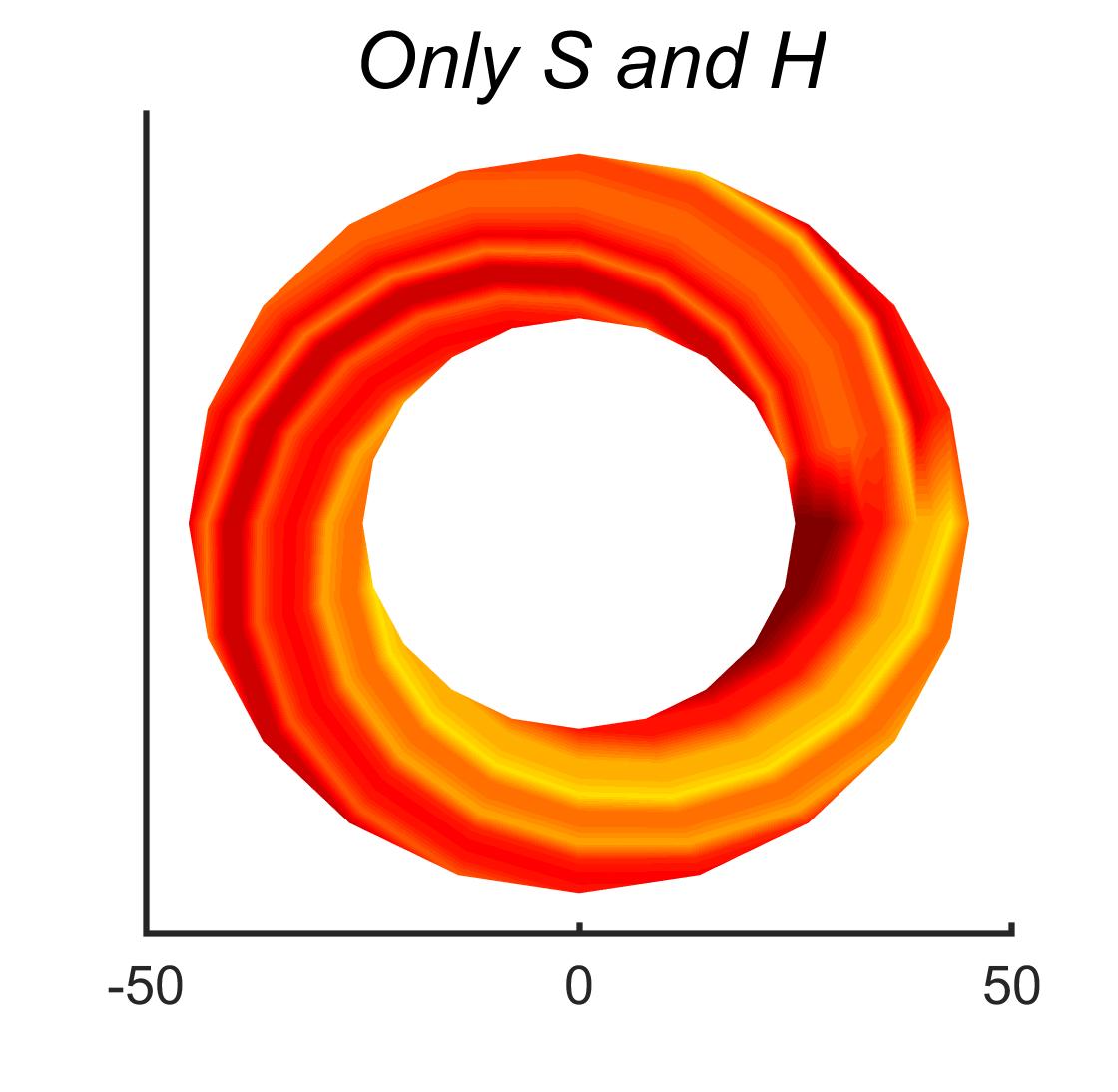}
		\end{subfigure}
		\begin{subfigure}{0.06\textwidth}
			\includegraphics[width=\textwidth]{other/colorbar.jpg}
		\end{subfigure}
		\caption{Accuracy of letter identification for the DenseNet-121 with single flankers when the target and flankers are placed on the right-hand side of the image, instead of the left-hand side. Training and testing was done with acuity loss. This is an additional run of the model presented in Figure \ref{drandominitsingleright} to verify results. Notice that target-flanker spacing decreases accuracy. Model accuracy without flankers was 97.92\%.}
		\label{drandominitsingleright2}
	\end{figure}

	To determine if the higher interference by a flanker placed along the top-left to bottom-right diagonal and close to the image centre was an artefact of the stimuli used and the training procedure, we trained a new network with the same parameters as before but with the letter stimuli presented on the right side instead of the left. As can be seen in Figure \ref{drandominitsingleright}, the shape of crowding flips across the vertical axis. That is, it is not the absolute top-left to bottom-right axis that matters, but the presence of a flanker close to the centre of the image, but in the lower visual field that causes the greatest disruption. To replicate these findings, we trained a new model using identical configuration (Figure \ref{drandominitsingleright2}). The general characteristics of crowding in this model remained the same, but it appears that this instantiation of the model (see Figure \ref{drandominitsingleright2}) is more robust to clutter and exhibits minimally reduced performance with increased target-flanker spacing. The reason for this discrepancy is unclear to us. Importantly, however, it is clear from these models and models trained with unaltered images that the pattern of crowding remains the same even with large changes in network and stimulus characteristics, even if the magnitude changes. This magnitude difference can be partially attributed to the image manipulations ('acuity loss'), but not the pattern of results.
	
	\subsection{DenseNet-121 with random and ILSVRC initialisations with paired flankers}
	\label{denseandimagenetsection}
\begin{figure}
		\centering
		\begin{subfigure}{0.9\textwidth}
			\includegraphics[width=\textwidth]{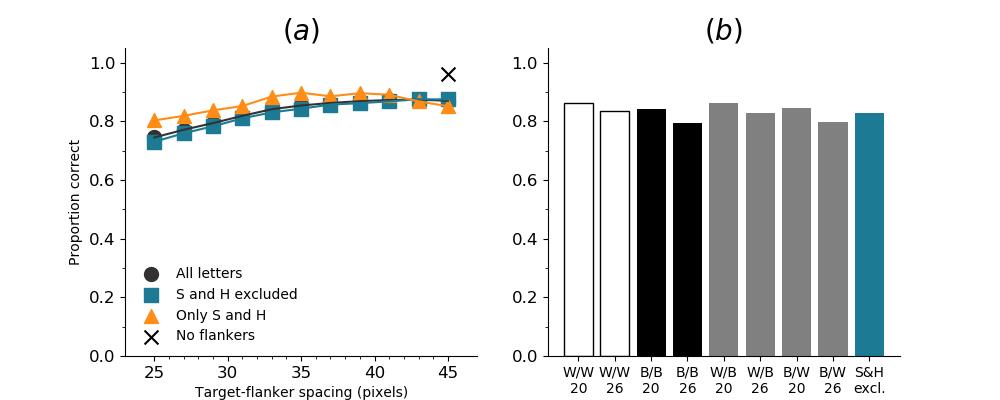}
		\end{subfigure}
		\begin{subfigure}{0.2562\textwidth}
			\includegraphics[width=\textwidth]{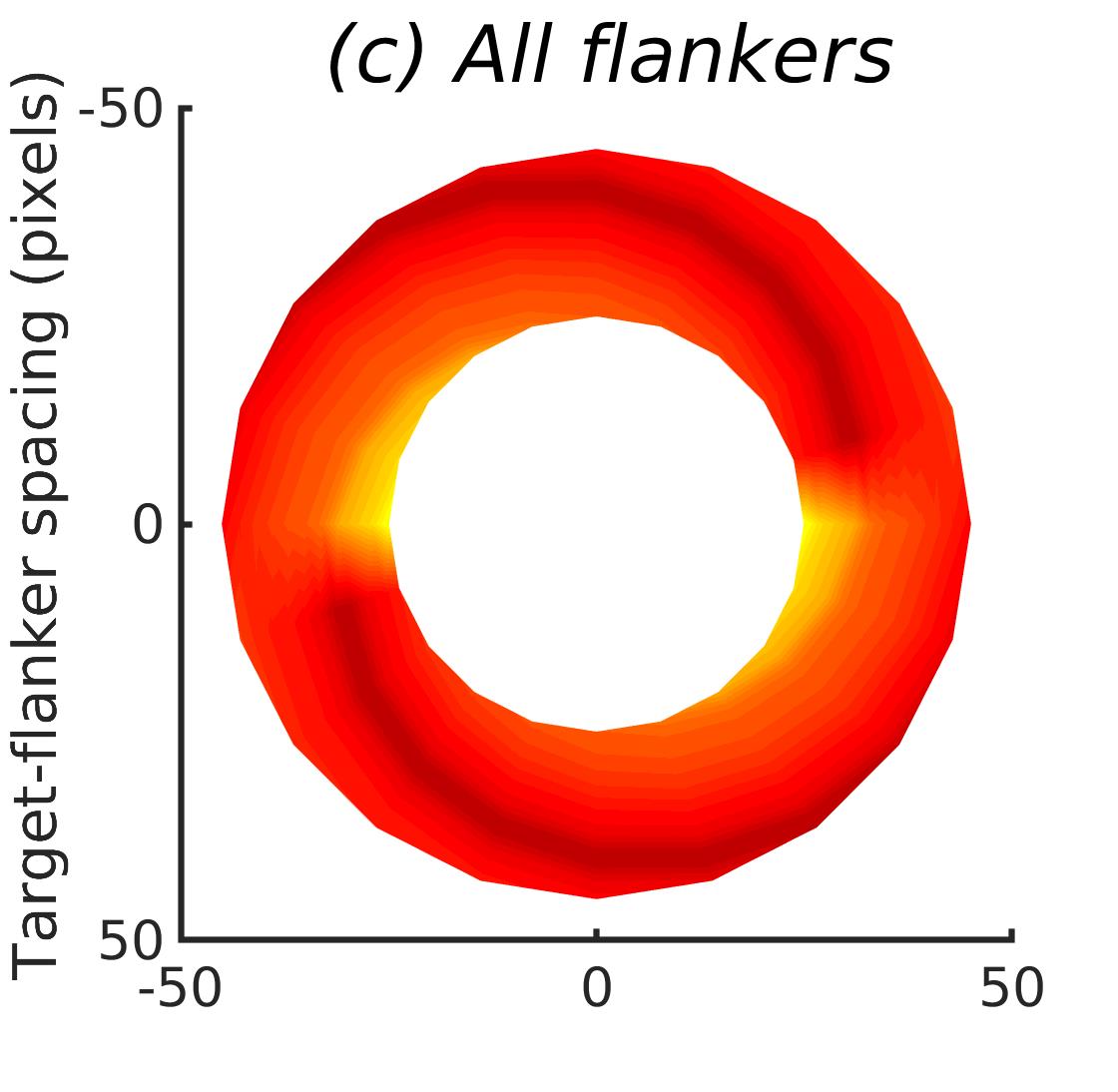}
		\end{subfigure}
		\begin{subfigure}{0.25\textwidth}
			\includegraphics[width=\textwidth]{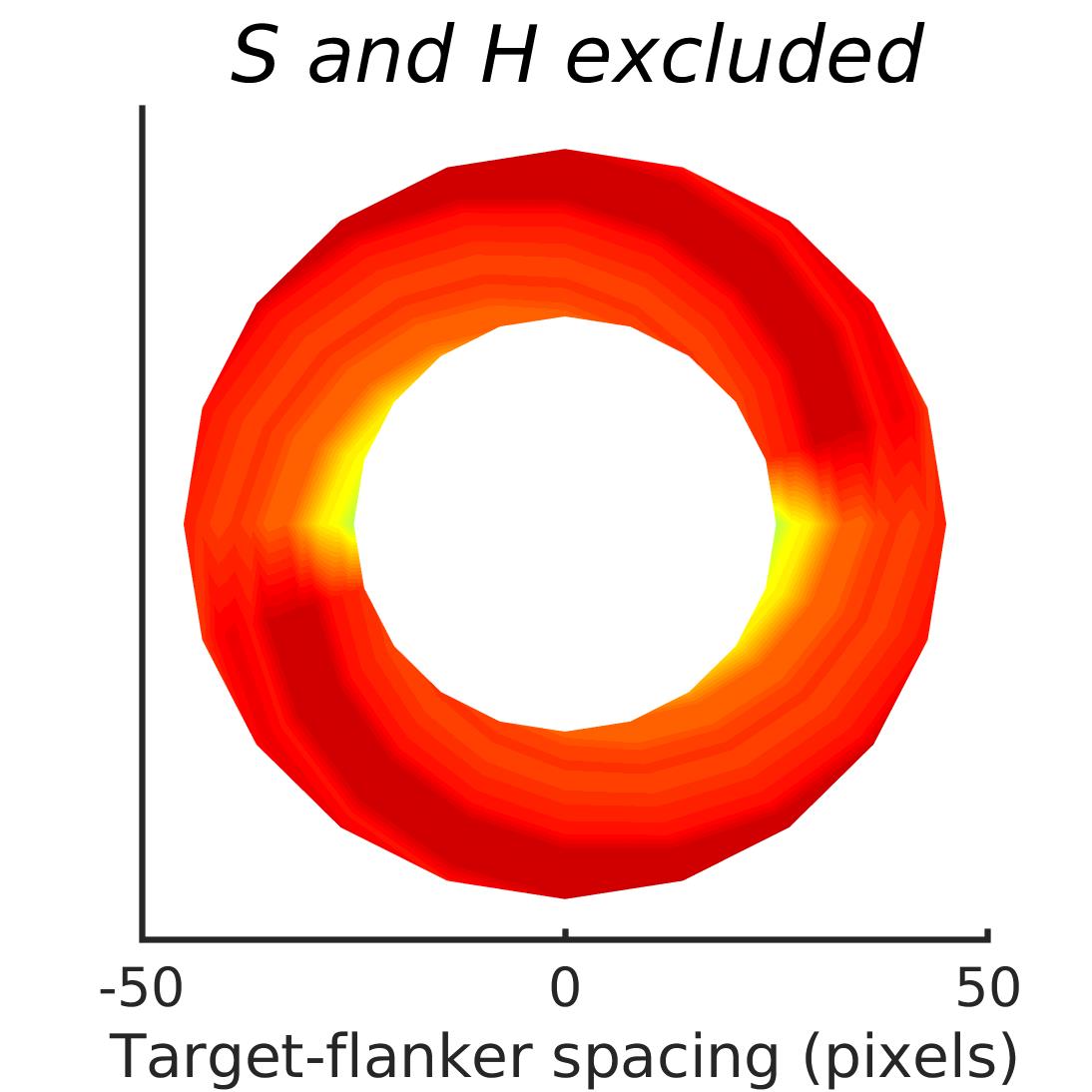}
		\end{subfigure}
		\begin{subfigure}{0.25\textwidth}
			\includegraphics[width=\textwidth]{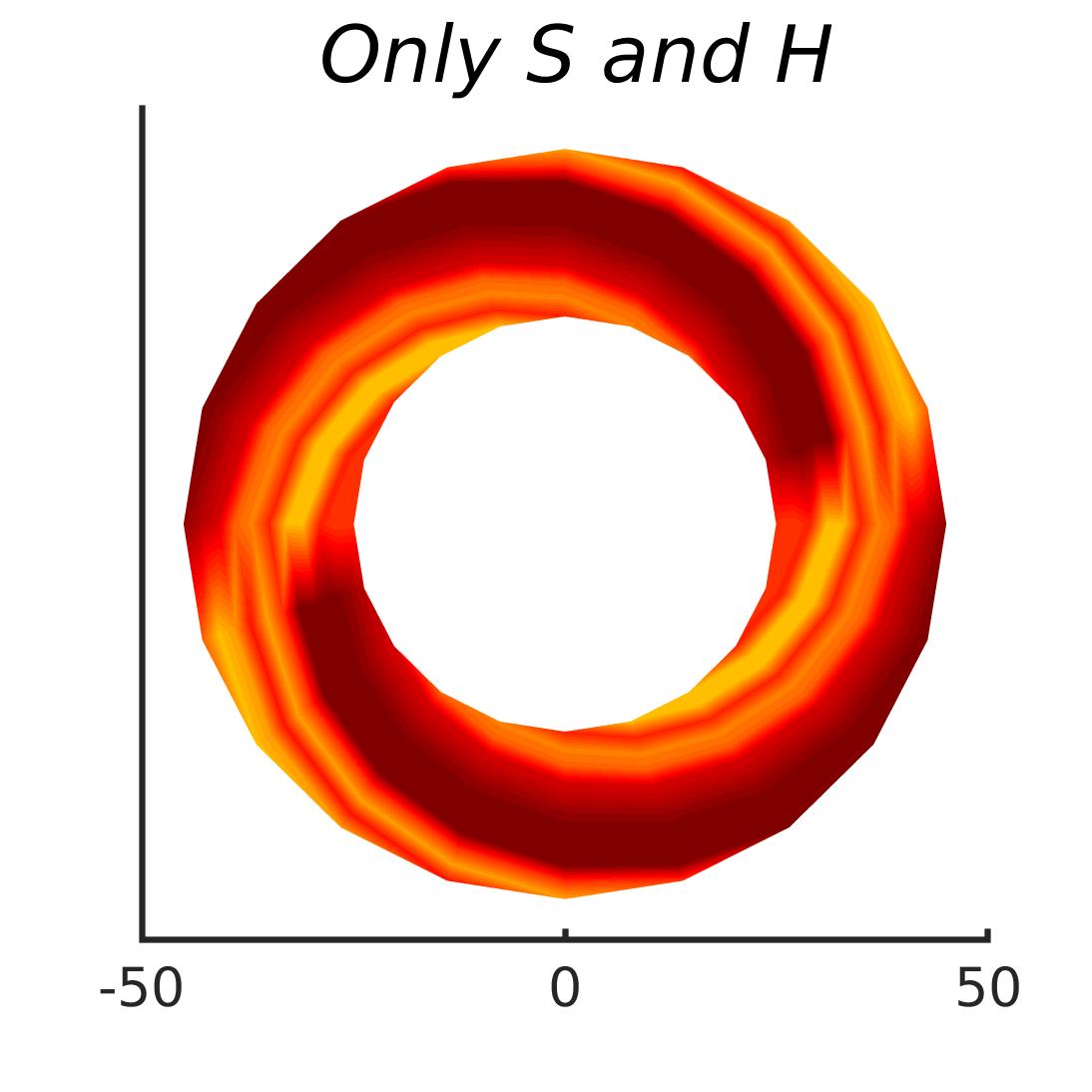}
		\end{subfigure}
		\begin{subfigure}{0.06\textwidth}
			\includegraphics[width=\textwidth]{other/colorbar.jpg}
		\end{subfigure}
		\caption{Accuracy of letter identification for the randomly initialised DenseNet-121 tested on paired flankers. The model was trained and tested with acuity loss, and its accuracy without flankers is 96.11\%.}
		\label{drandominit}
	\end{figure}

\begin{figure}
		\centering
		\begin{subfigure}{0.9\textwidth}
			\includegraphics[width=\textwidth]{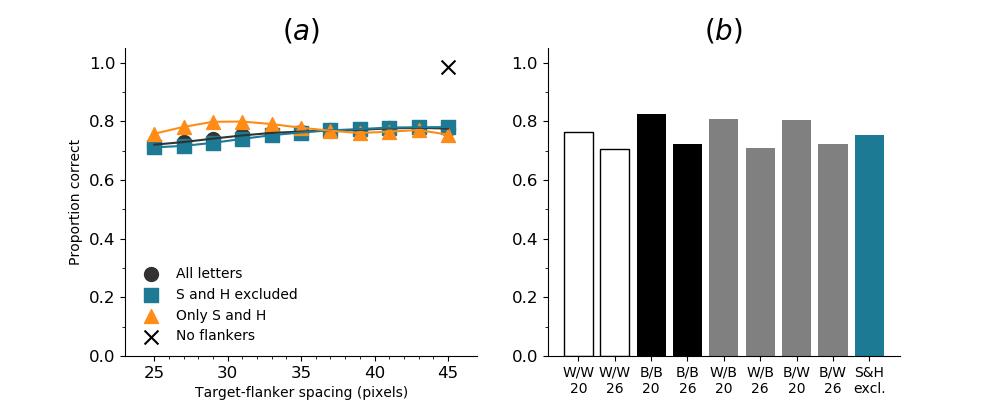}
		\end{subfigure}
		\begin{subfigure}{0.2562\textwidth}
			\includegraphics[width=\textwidth]{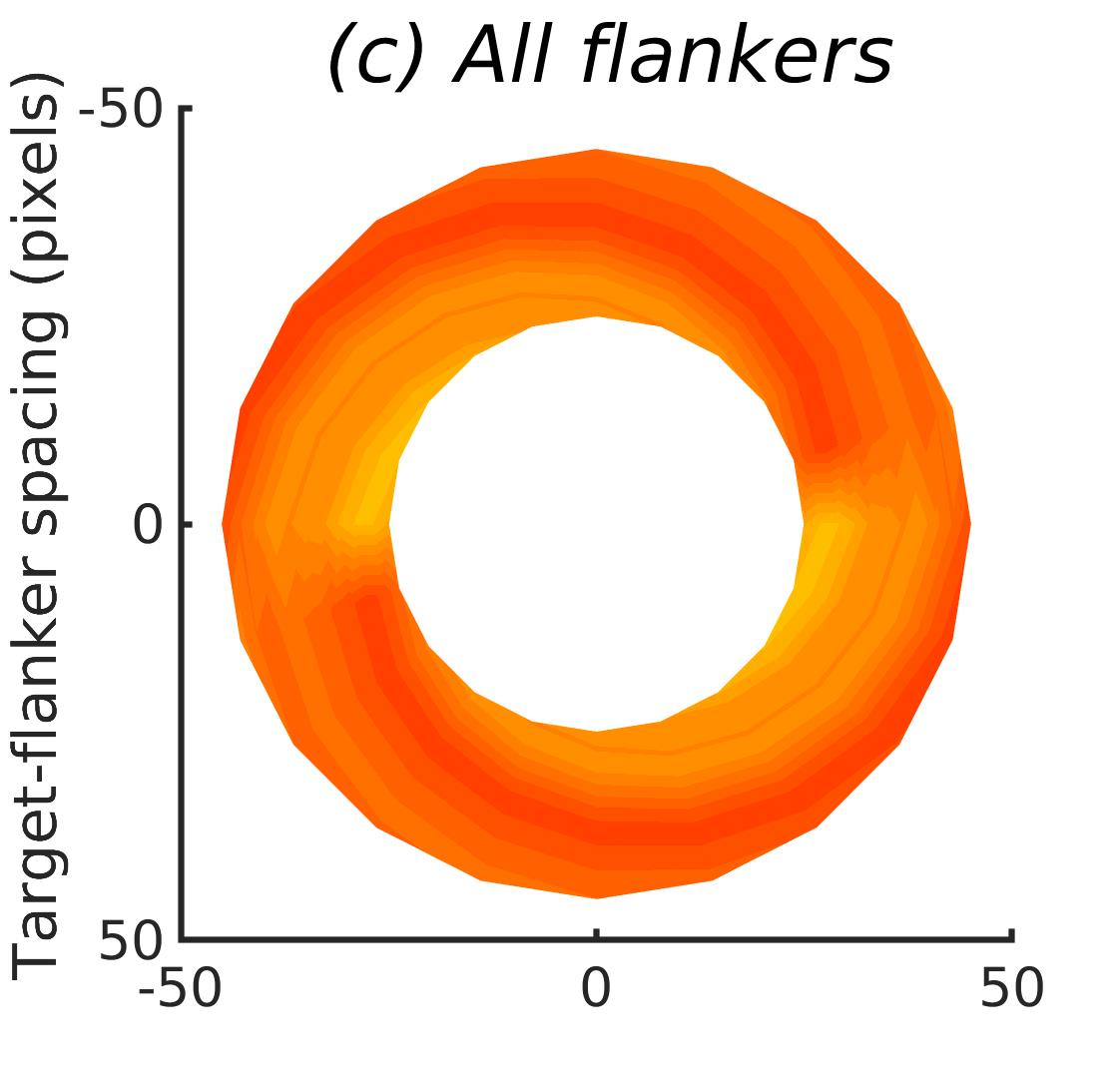}
		\end{subfigure}
		\begin{subfigure}{0.25\textwidth}
			\includegraphics[width=\textwidth]{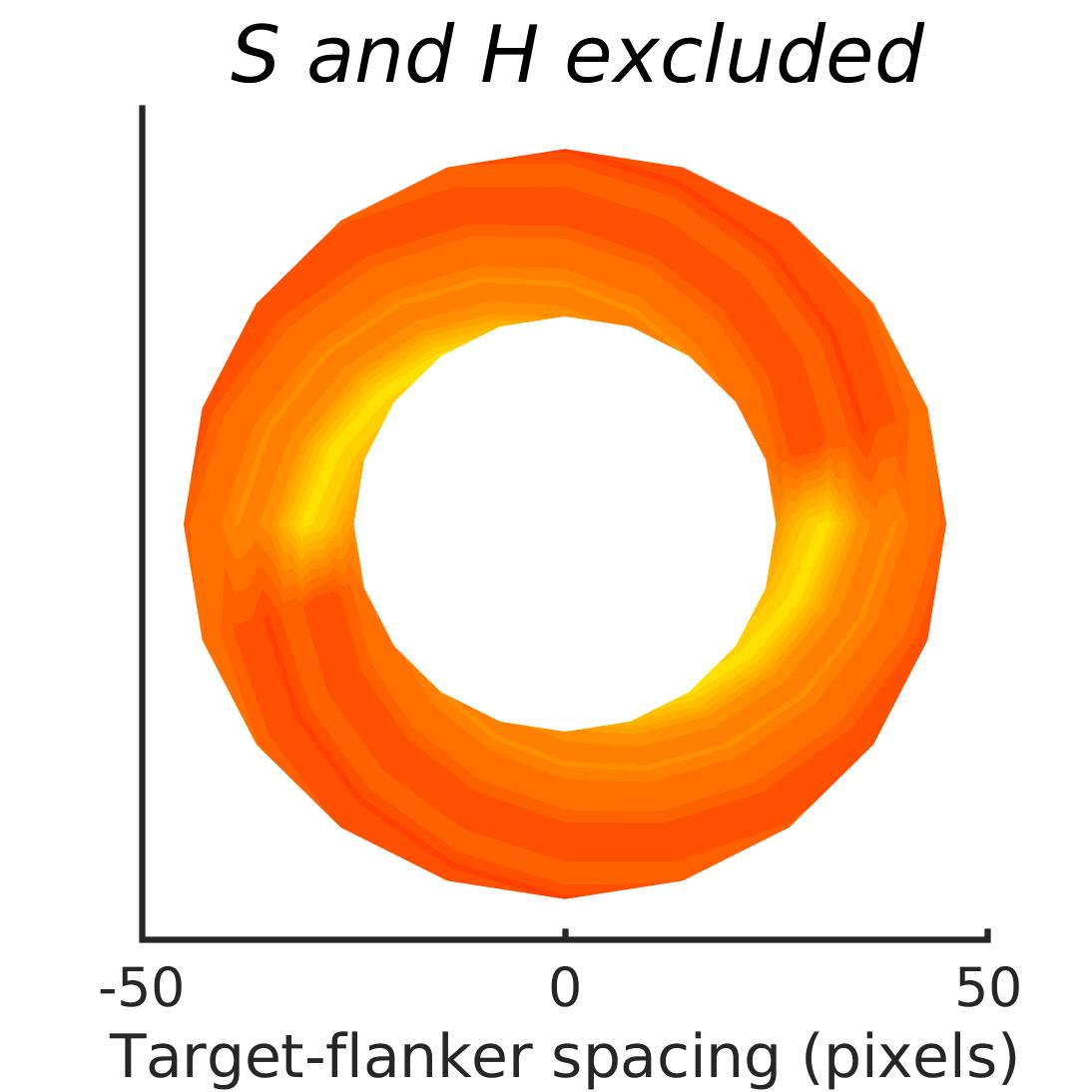}
		\end{subfigure}
		\begin{subfigure}{0.25\textwidth}
			\includegraphics[width=\textwidth]{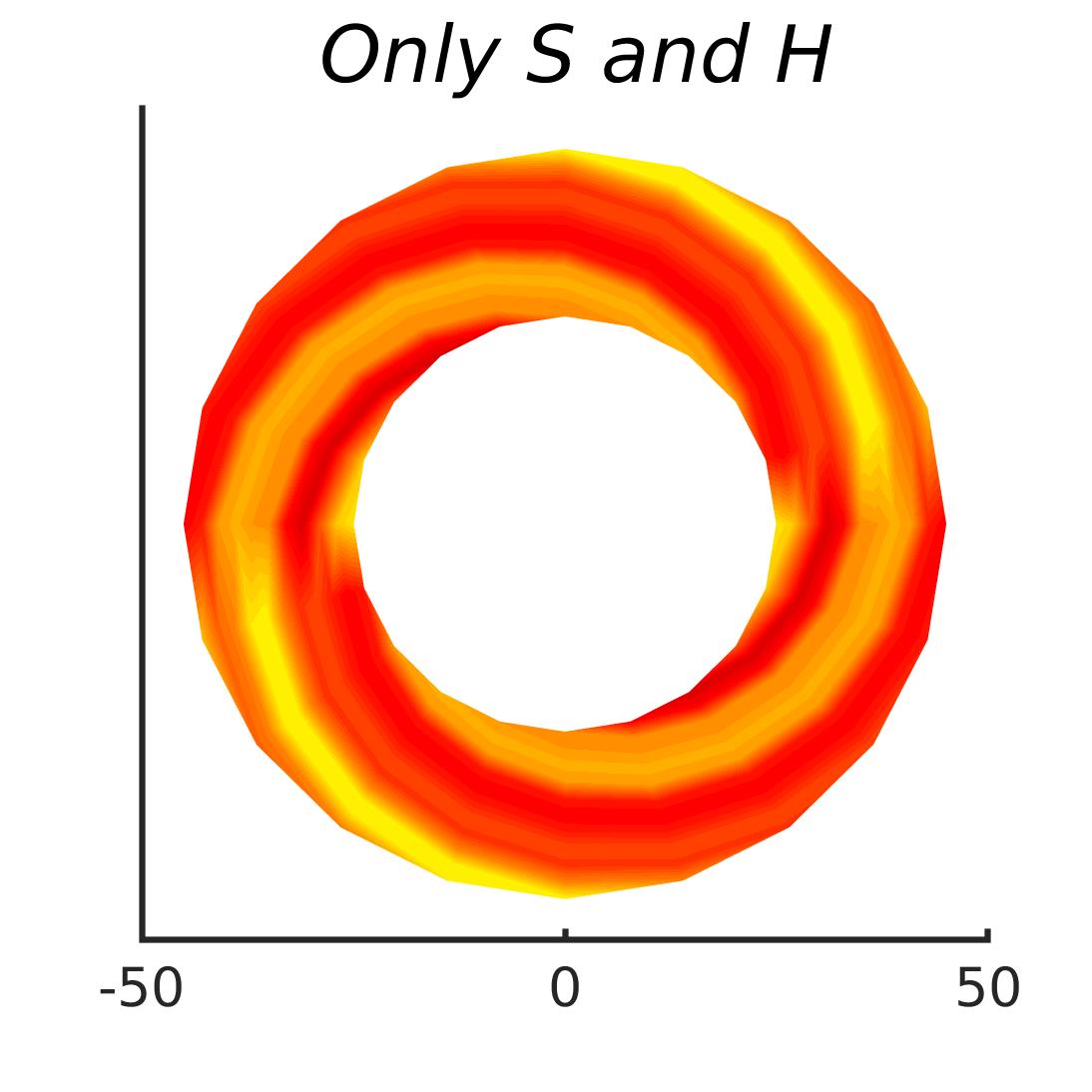}
		\end{subfigure}
		\begin{subfigure}{0.06\textwidth}
			\includegraphics[width=\textwidth]{other/colorbar.jpg}
		\end{subfigure}
		\caption{Accuracy of letter identification for the ILSVRC-initialised DenseNet-121. Training and testing was done with acuity loss. Model accuracy without flankers was 98.52\%.}
		\label{dimagenet}
	\end{figure}

	Psychophysical experiments in humans on crowding are often performed with a pair of flankers, one on either side of the target, rather than a single one (e.g., \citet{bouma_interaction_1970}, \citet{freeman_substitution_2012}). Hence, we also tested the DenseNet-121 with paired flankers. In these experiments we trained the DenseNet-121 initialised with random and ILSVRC weights, separately. Results are shown in Figures \ref{drandominit} and \ref{dimagenet}, respectively. We found that the bottom-right flanker that dominates crowding in single-flanker experiments causes the general pattern of crowding to replicate across the horizontal axis (along the top-left to bottom-right axis). It is interesting to note that the model is crowded more by paired flankers at all distances than by single flankers, and does not reach near-unflanked accuracy even at the furthest target-flanker distance. In humans, paired flankers are more effective in interfering with performance than single flankers, and have a larger range of interference. That is, they are more effective even at larger distances. DenseNet-121 appears to mirror that characteristic. Nevertheless, in humans, crowding is eliminated, under similar circumstances, if the distance between the target and the flankers is greater than half the target eccentricity (the distance between the centre of the image and the target), as codified in the 'Bouma Law' \citep{pelli_uncrowded_2008}. This is equivalent to a spacing of about 29 pixels in our setup. That is, performance should be the same as in the unflanked condition if flankers are separated from the target by about 29 pixels. This is not the case with the DenseNet model. 
	
	\subsection{Effect of acuity loss manipulation}
	As the DenseNet-121 was more robust to flanker interference, we trained and tested the DenseNet-121 with the same hyperparameters on images that had not been reduced in acuity (unaltered images). We found that the general shape of crowding remained the same in all tests but one (pair flankers with ILSVRC initialisation: Figure \ref{dnoacuitylossimagenet}), and barring that experiment the effect of flankers was dramatically reduced. This suggests that the image manipulation cannot explain the pattern of our results, apart from its magnitude. These results also point to the proposal that the pattern of results observed here is inherent to DCNNs. 
	
	In the special case of ILSVRC initialisation with a pair of flankers, the performance was much lower than expected (roughly 40\%), whereas for most other experiments with acuity loss this ranged from 60-85\%. This strange behaviour may have been caused by differences in convergence of the network. In addition to poor performance in the test, the axis of crowding flipped compared to all other experiments. These results closely mimic the effects seen in the SimpleNet tests; robustness to clutter of the DenseNet is higher, and regardless of whether the acuity loss procedure is used, the general characteristics of crowding remain.
	
	We found that while using unaltered images in training and testing can lead to some unpredictable results, such as massive performance drops or improvements with flankers, the general shapes of crowding tended to stay the same. We also found that in the case of experiments trained and tested with unaltered image, anomalies of crowding, described in Section \ref{denseandimagenetsection}, largely disappeared when the flankers S and H were excluded from analysis.
	
\begin{figure}
		\centering
		\begin{subfigure}{0.9\textwidth}
			\includegraphics[width=\textwidth]{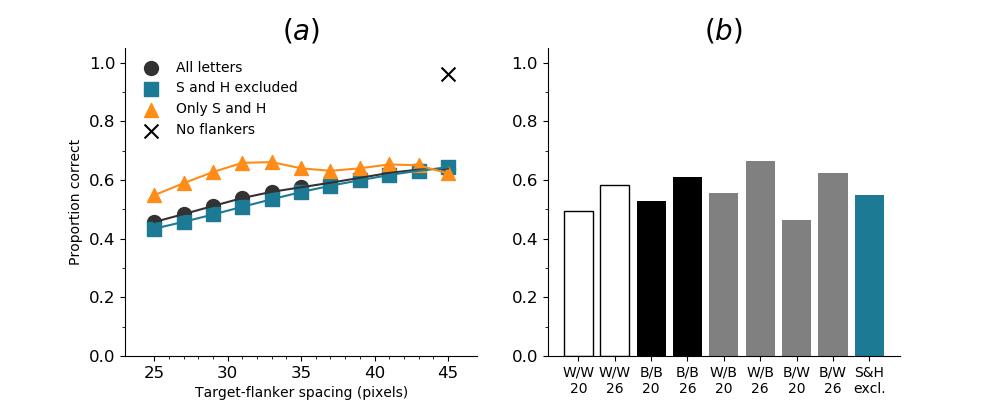}
		\end{subfigure}
		\begin{subfigure}{0.2562\textwidth}
			\includegraphics[width=\textwidth]{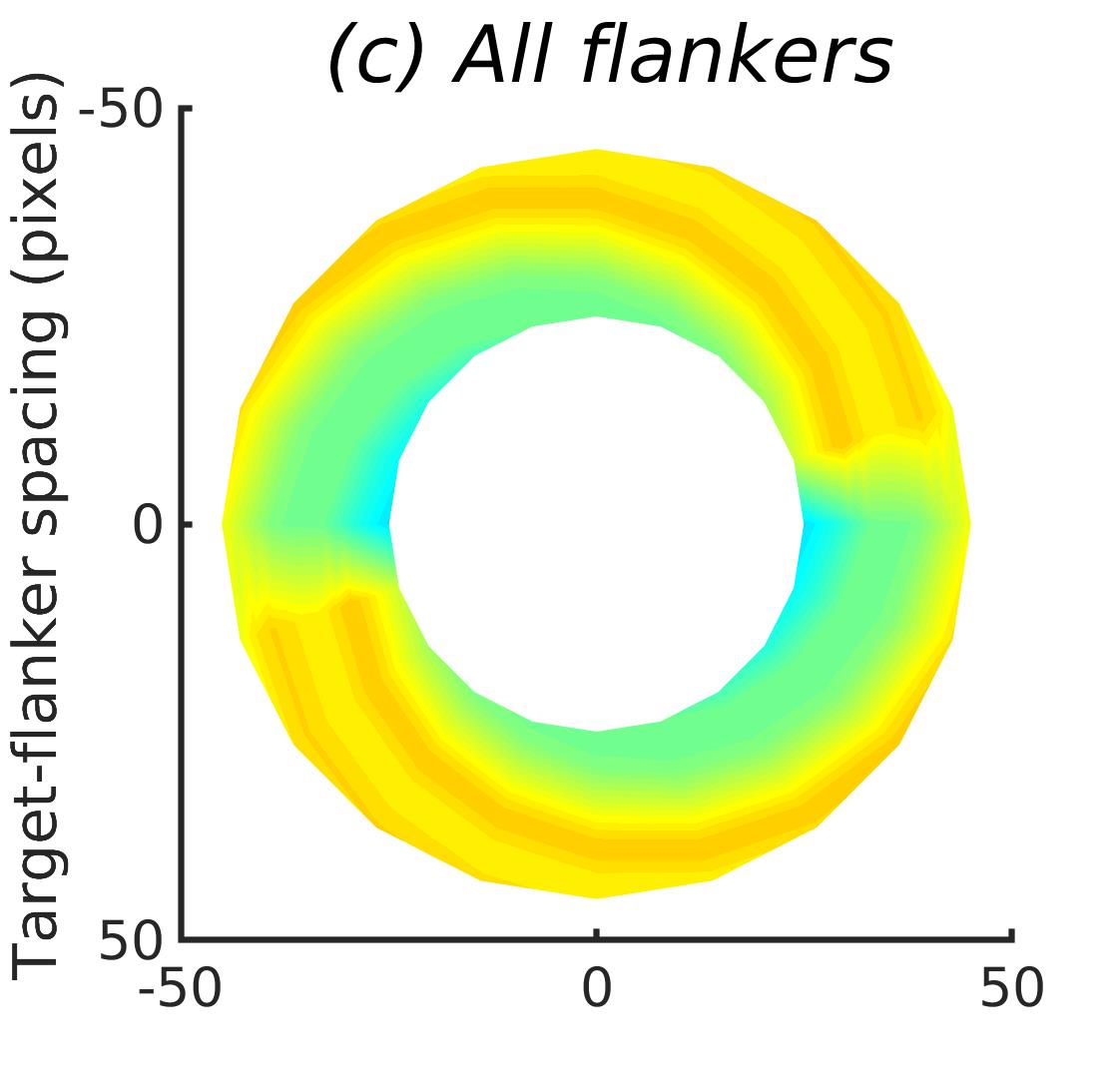}
		\end{subfigure}
		\begin{subfigure}{0.25\textwidth}
			\includegraphics[width=\textwidth]{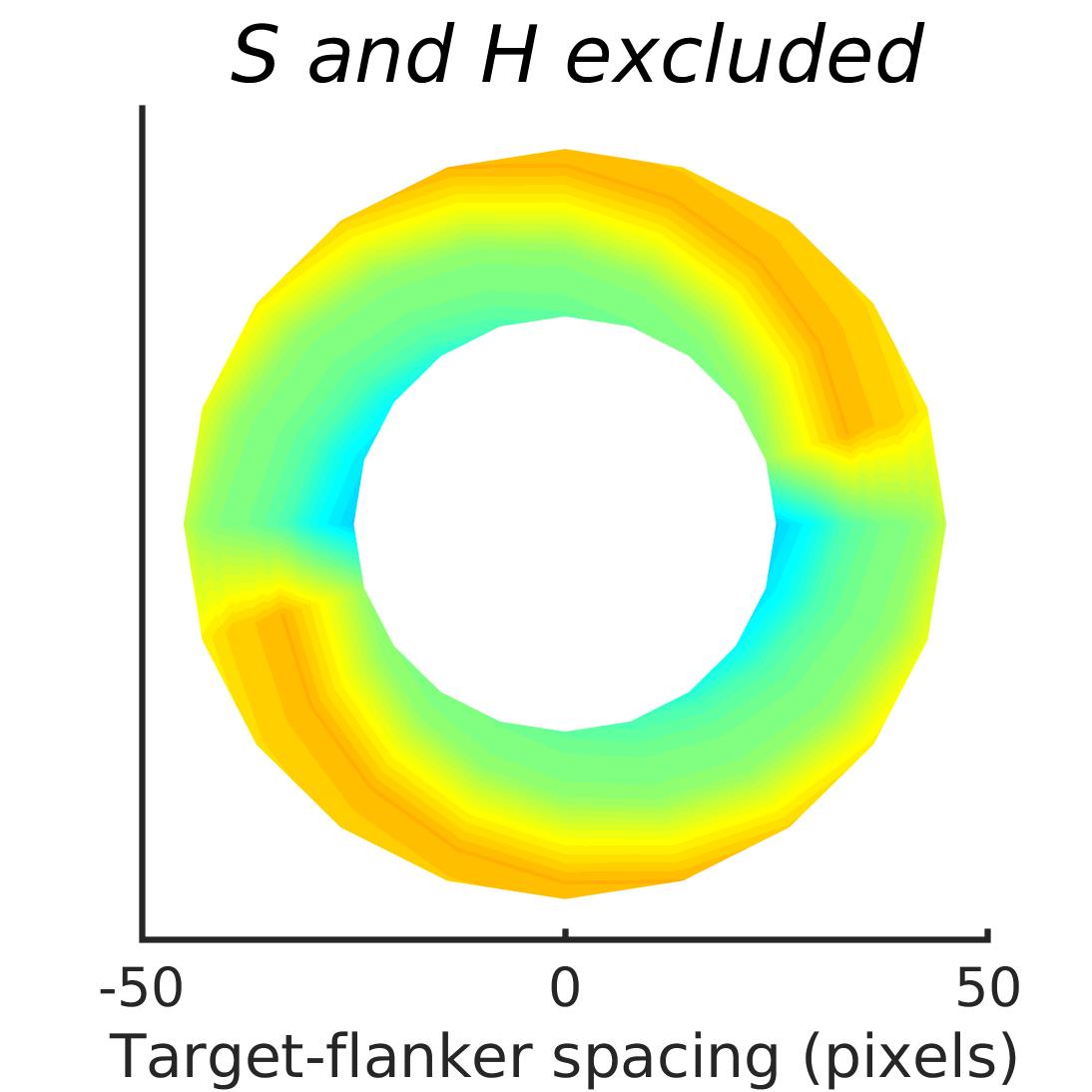}
		\end{subfigure}
		\begin{subfigure}{0.25\textwidth}
			\includegraphics[width=\textwidth]{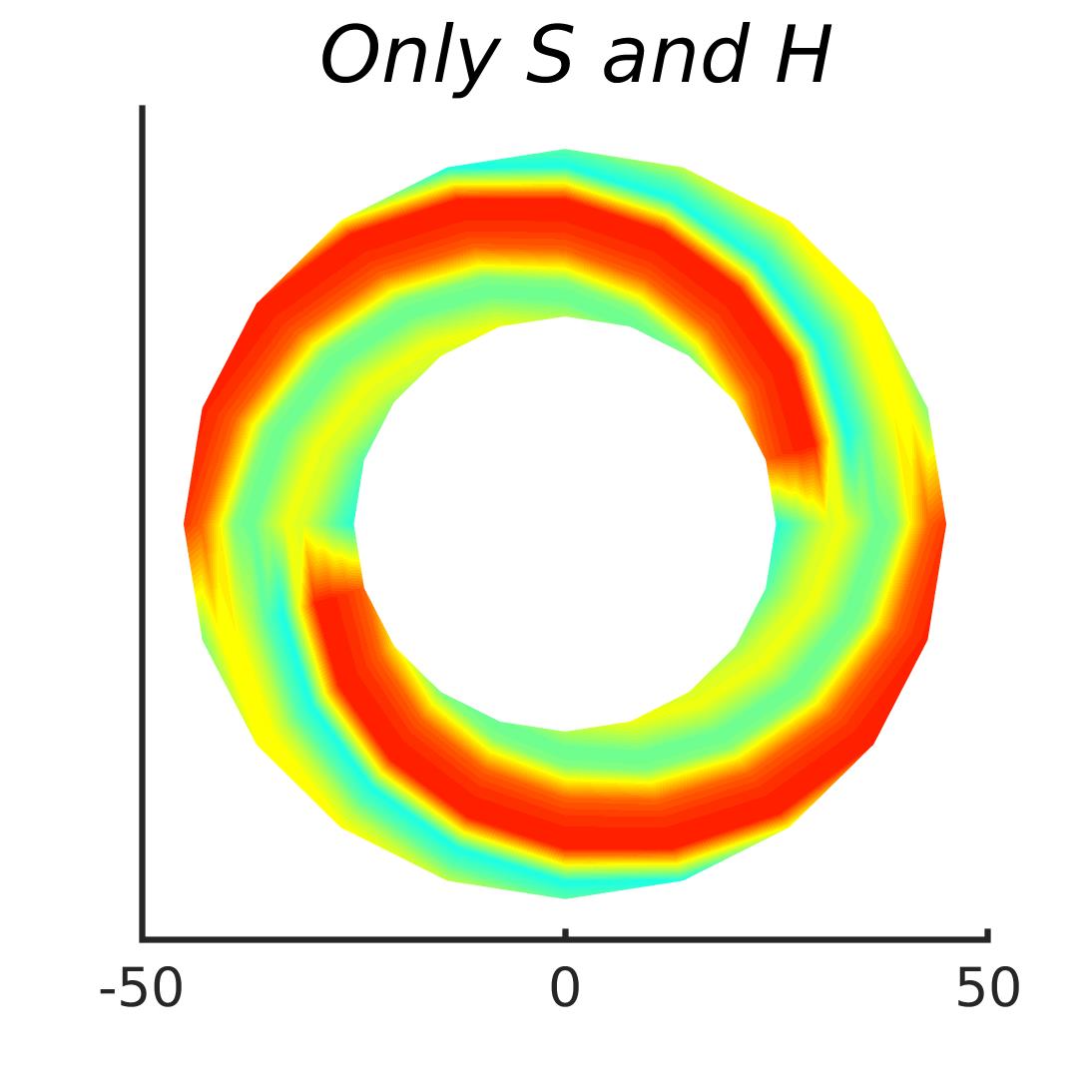}
		\end{subfigure}
		\begin{subfigure}{0.06\textwidth}
			\includegraphics[width=\textwidth]{other/colorbar.jpg}
		\end{subfigure}
		\caption{Accuracy of letter identification of the DenseNet-121 with randomly initialised weights trained with acuity loss and tested with full acuity stimuli. The network was shown pair flankers and did not exhibit a large change in behaviour with access to full acuity. Model accuracy without flankers was 96.11\%.}
		\label{dacuitylosstrainnoacuitylosstestrandominit}
	\end{figure}

	Finally, we tested the randomly initialised DenseNet that was trained with acuity loss to see how behaviour changes when the network gains access to full acuity without additional training. Results are shown in Figure \ref{dacuitylosstrainnoacuitylosstestrandominit}. 
	
	We found that there was no large difference in the general characteristics of crowding regardless of whether the network had reduced acuity during training or during testing. Acuity affected crowding primarily in magnitude, but not in shape or general characteristics, such as the effect of distance on crowding, or the effect of similarity (contrast polarity and size) between the target and flankers.
	
	\subsection{Effect of the amount of useful information in a local region}
	
\begin{figure}
		\centering
		\begin{subfigure}{0.9\textwidth}
			\includegraphics[width=\textwidth]{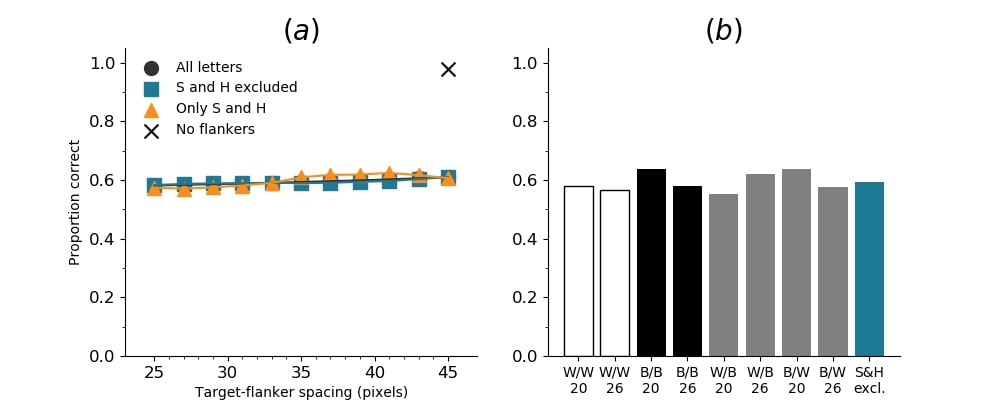}
		\end{subfigure}
		\begin{subfigure}{0.2562\textwidth}
			\includegraphics[width=\textwidth]{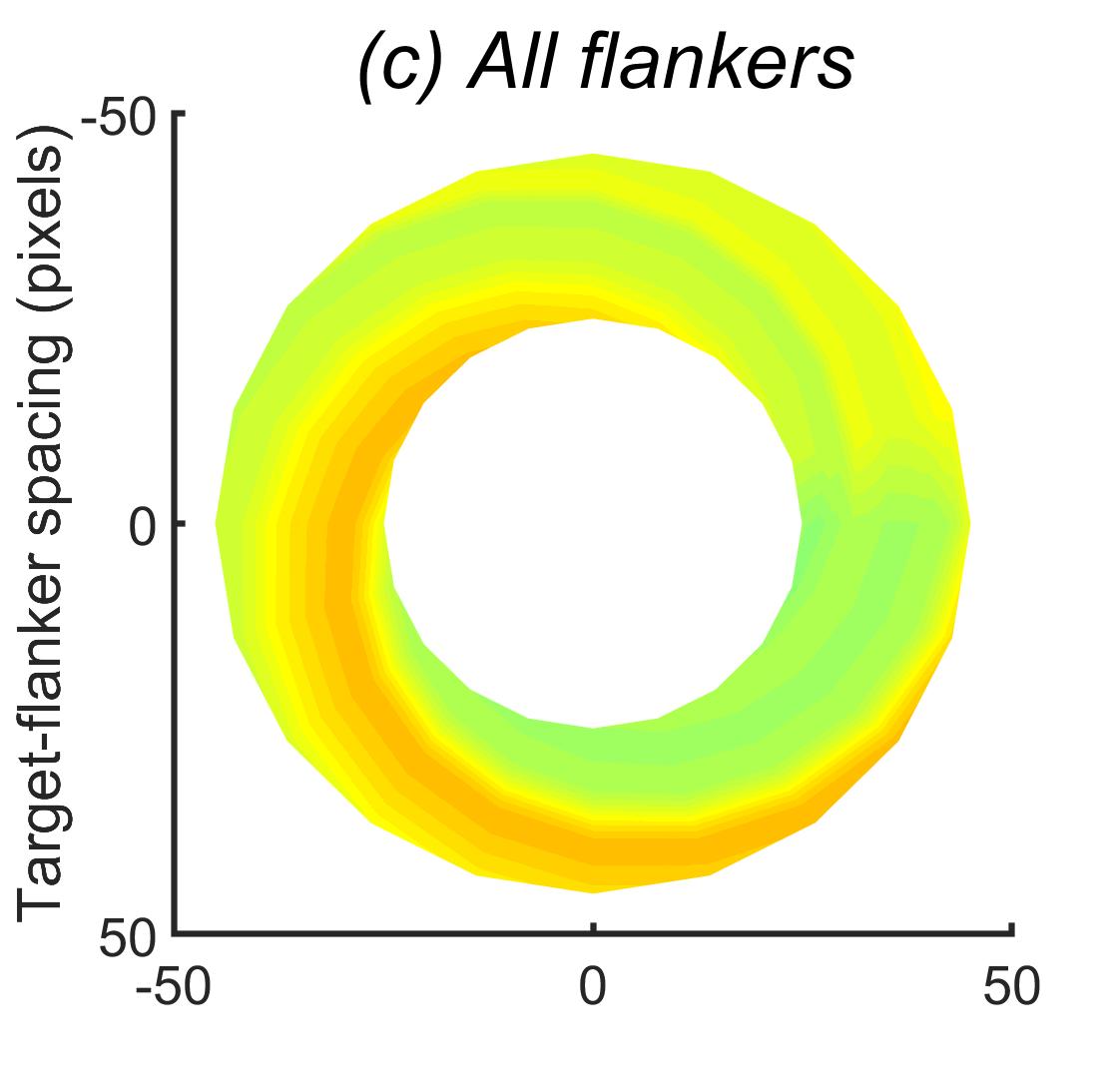}
		\end{subfigure}
		\begin{subfigure}{0.25\textwidth}
			\includegraphics[width=\textwidth]{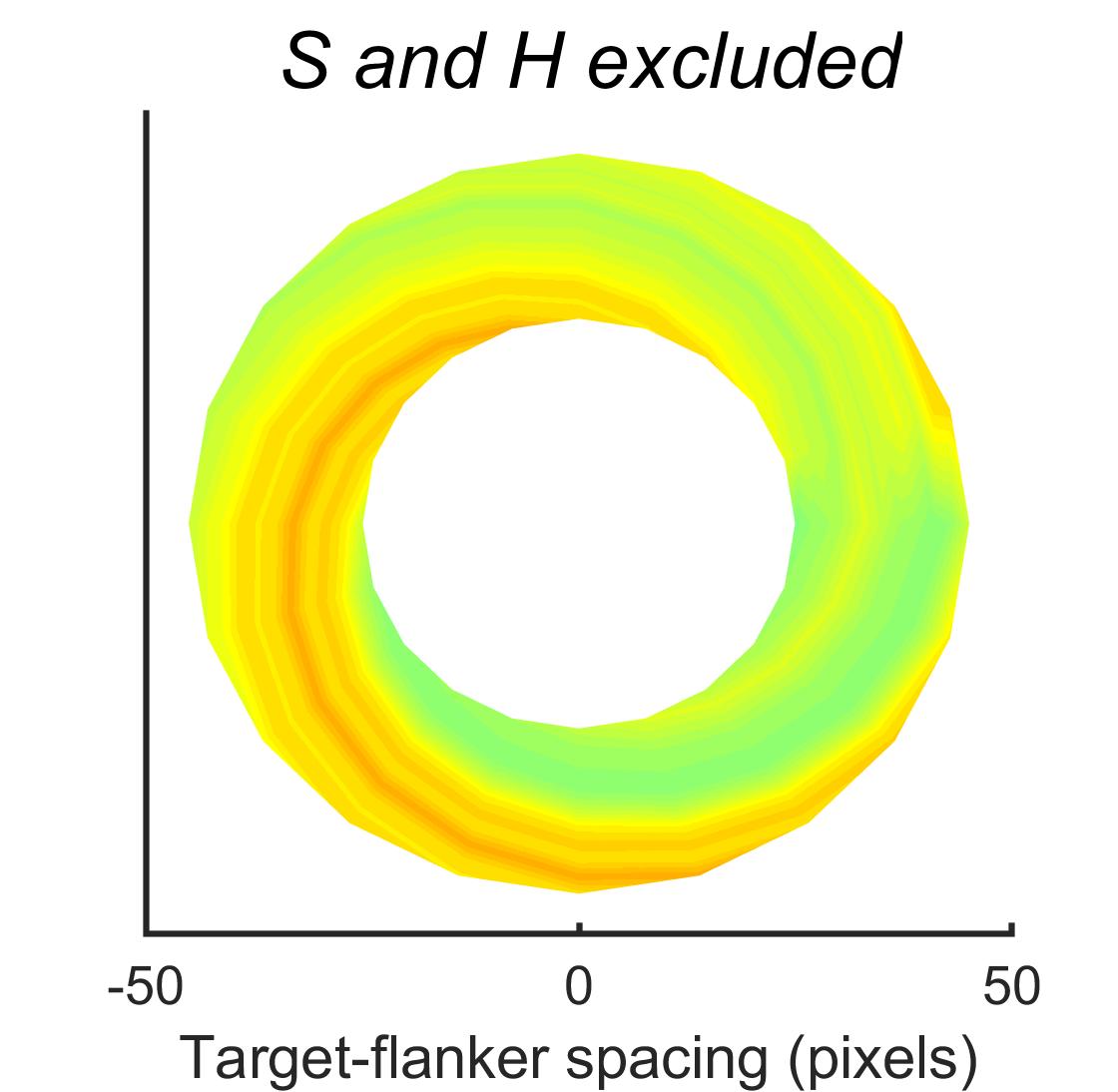}
		\end{subfigure}
		\begin{subfigure}{0.25\textwidth}
			\includegraphics[width=\textwidth]{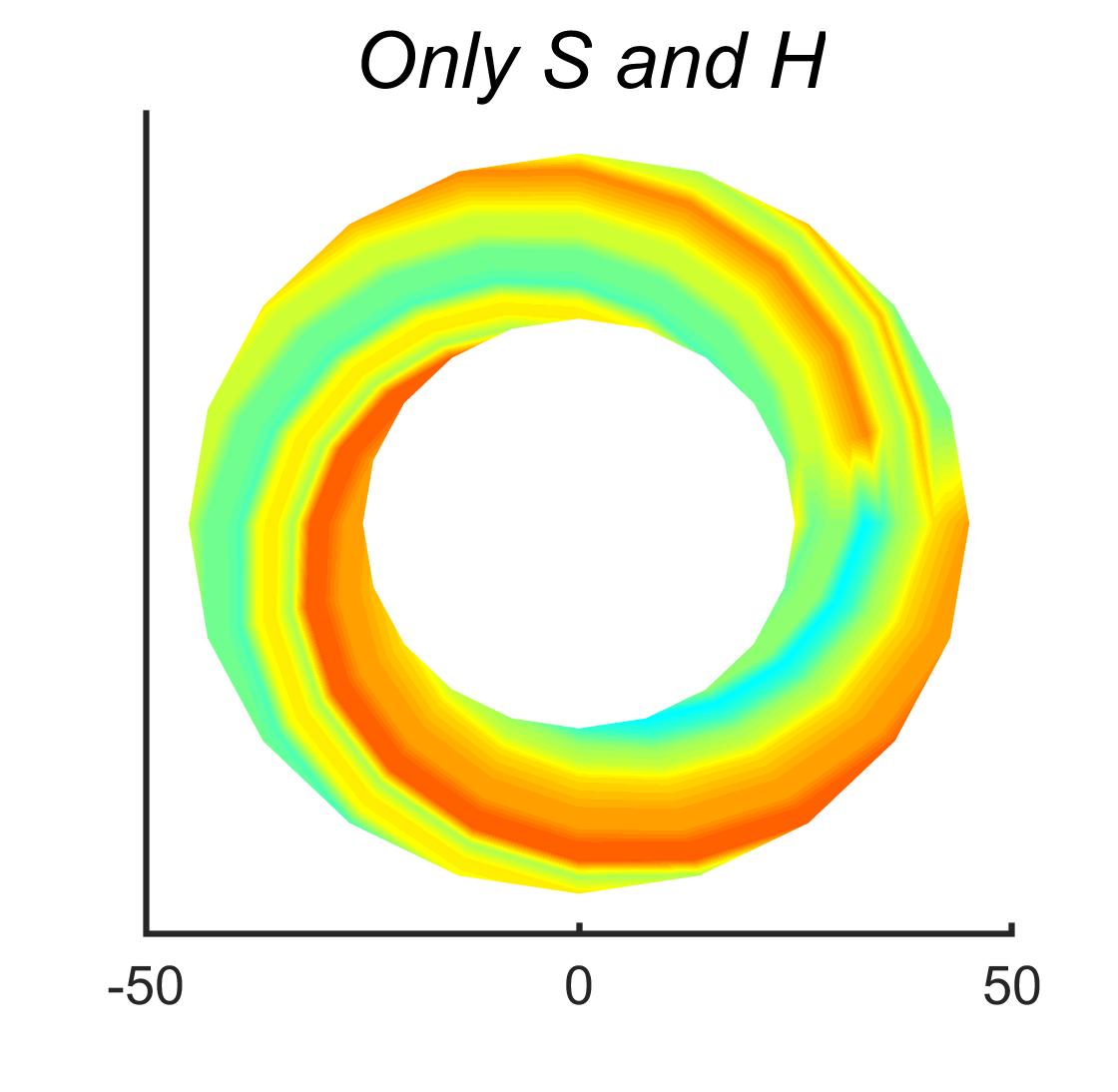}
		\end{subfigure}
		\begin{subfigure}{0.06\textwidth}
			\includegraphics[width=\textwidth]{other/colorbar.jpg}
		\end{subfigure}
		\caption{Accuracy of letter identification for a randomly initialised DenseNet-121 with background stimuli vertically flipped. Training and testing were done with acuity loss. We found that the degree of crowding decreases with distance less than in most of our experiments. Model accuracy without flankers was 97.98\%.}
		\label{dflippedrandominitsingle}
	\end{figure}

	Because our models suffered from crowding to a greater degree in the lower half of the images than in the upper half, we tested whether flipping our background images vertically in training would also flip crowding vertically. It is possible that natural images have diagnostic information in the lower visual field and the network is more sensitive to clutter in that part of the visual field. Results are shown in Figure \ref{dflippedrandominitsingle}. We found that a relatively large portion of crowding does shift to the upper half of the image, practically equalising the amount of crowding on both halves of the image (59.66\% accuracy on the top-half, 59.42\% accuracy on the bottom-half). This suggests that the amount of useful information in local regions of a stimulus plays a contributing role in crowding in DCNNs. This effect is the opposite of what is observed in humans. Humans have greater attentional resolution and lower crowding in the lower half of the visual field \citep{intriligator_spatial_2001}. Our DCNN models do not. We hypothesise that the networks developed to have greater preference for regions with a higher density of useful information for classification and hence flankers placed in such locations caused more crowding. In other words, the data used to train the networks likely contributes to this effect.
	
	It is interesting to note that while in the randomly initialised single-flanker model with upright background images the models exhibited a greater degree of crowding in the lower portion of the image (89.47\% accuracy in the lower half, 95.31\% in the upper half), this effect was not entirely reversed when the background images were vertically flipped; the accuracy between the two halves of the stimuli only equalises. We are unable to explain this phenomenon.
	
	\subsection{Radial-tangential asymmetry}
	
	\begin{figure}
		\centering
		\includegraphics[width=0.5\textwidth, keepaspectratio]{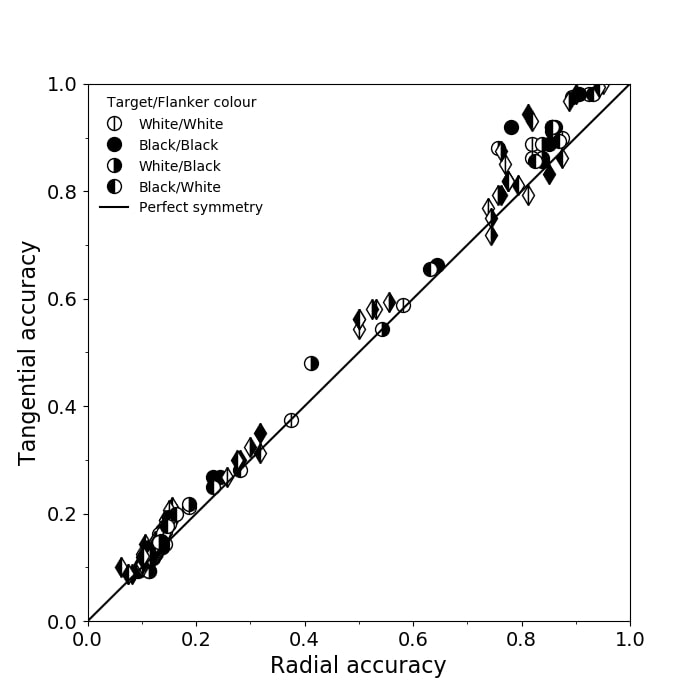}
		\caption{Radial-tangential accuracy for all single-flanker models. Only the horizontal and vertical flankers at 25px centre-centre distance from the target are plotted. Circle-shaped markers denote the letter size 20, and diamond markers denote the letter size 26. The asymmetry is relatively mild---in humans at certain distances effects have greater magnitude.}
		\label{radial-tangential}
	\end{figure}
	
	We examined if the networks demonstrate the radial-tangential asymmetry, where flankers along the radial direction (flankers along the axis connecting the target to the centre of the image) are more influential than tangential flankers. We plotted the accuracy of identification when it was surrounded by the nearest letters along these two axes (Figure \ref{radial-tangential}). We found that crowding is asymmetric in the expected direction---radial flankers crowd more than tangential flankers do, like in humans \citep{toet_two-dimensional_1992, petrov_asymmetries_2011}. We also find that if the overall accuracy is higher, crowding is more asymmetric than if it is lower.
	
	We also note, as described above, that the networks demonstrate an in-out asymmetry, where there is a difference in effect of the 'inner' flanker, or flanker closest to the image centre and 'outer' flanker, or the farthest flanker, The inner flanker appears to be more powerful in disrupting target identification than the outer one. However, the effect is the opposite of what is observed among humans, where the outer flanker is more influential than the inner one.
	
	\subsection{Effect of size and contrast polarity}
\begin{figure}
		\centering
		\begin{subfigure}{0.9\textwidth}
			\includegraphics[width=\textwidth]{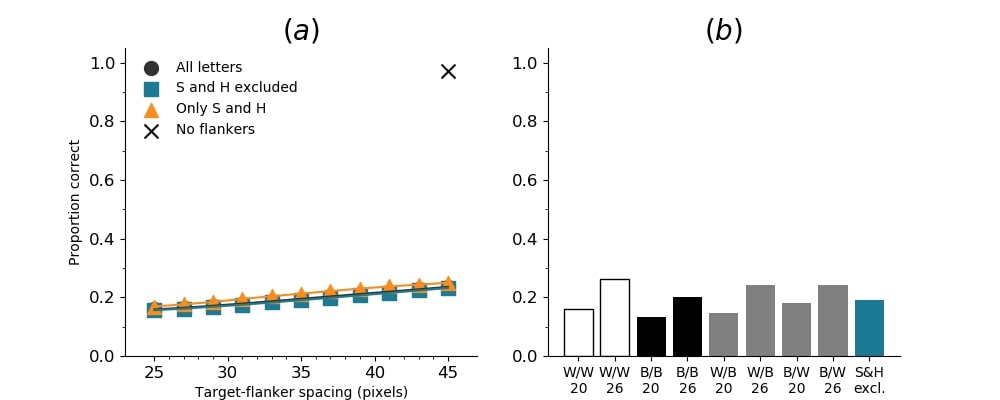}
		\end{subfigure}
		\begin{subfigure}{0.2562\textwidth}
			\includegraphics[width=\textwidth]{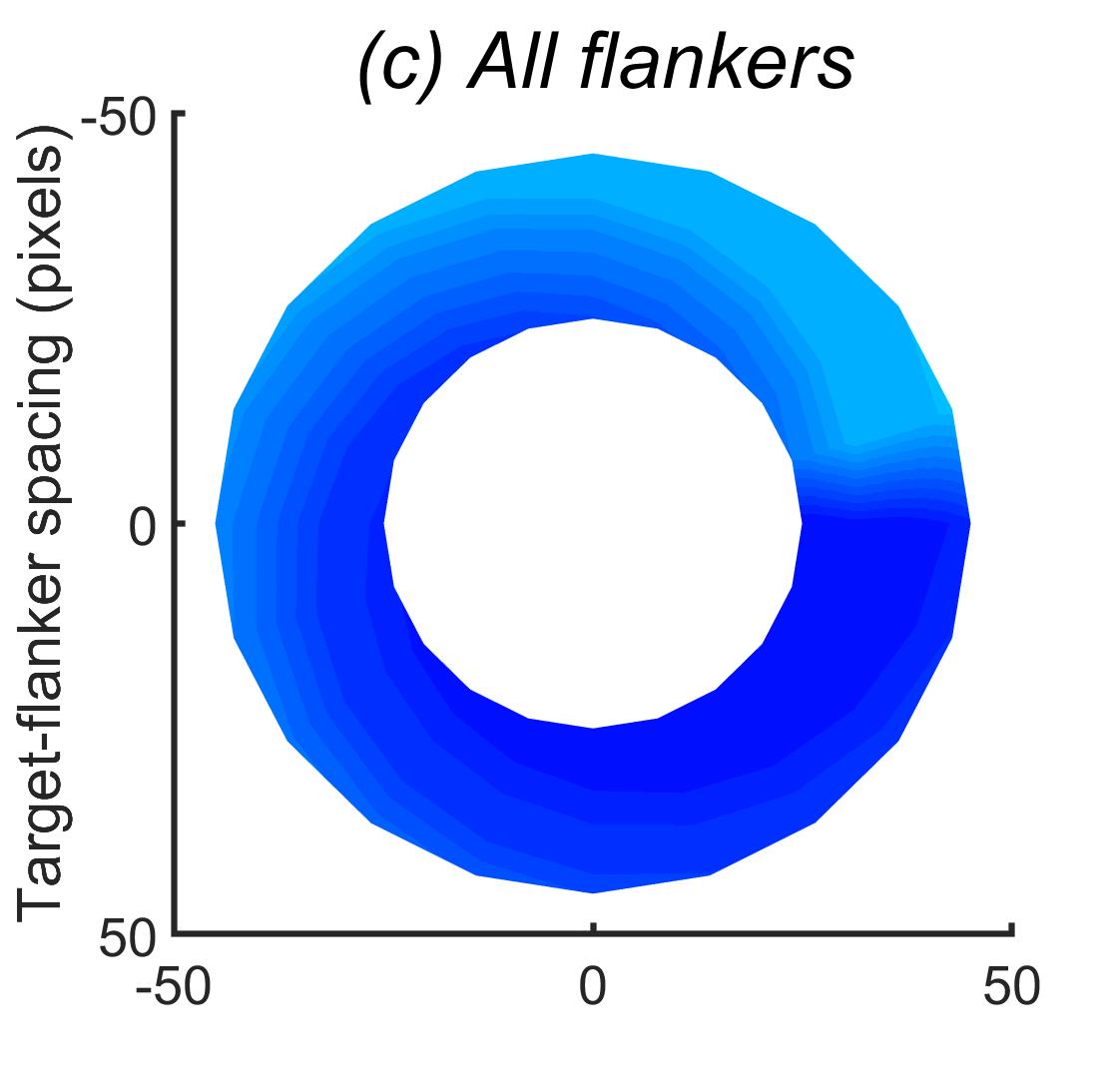}
		\end{subfigure}
		\begin{subfigure}{0.25\textwidth}
			\includegraphics[width=\textwidth]{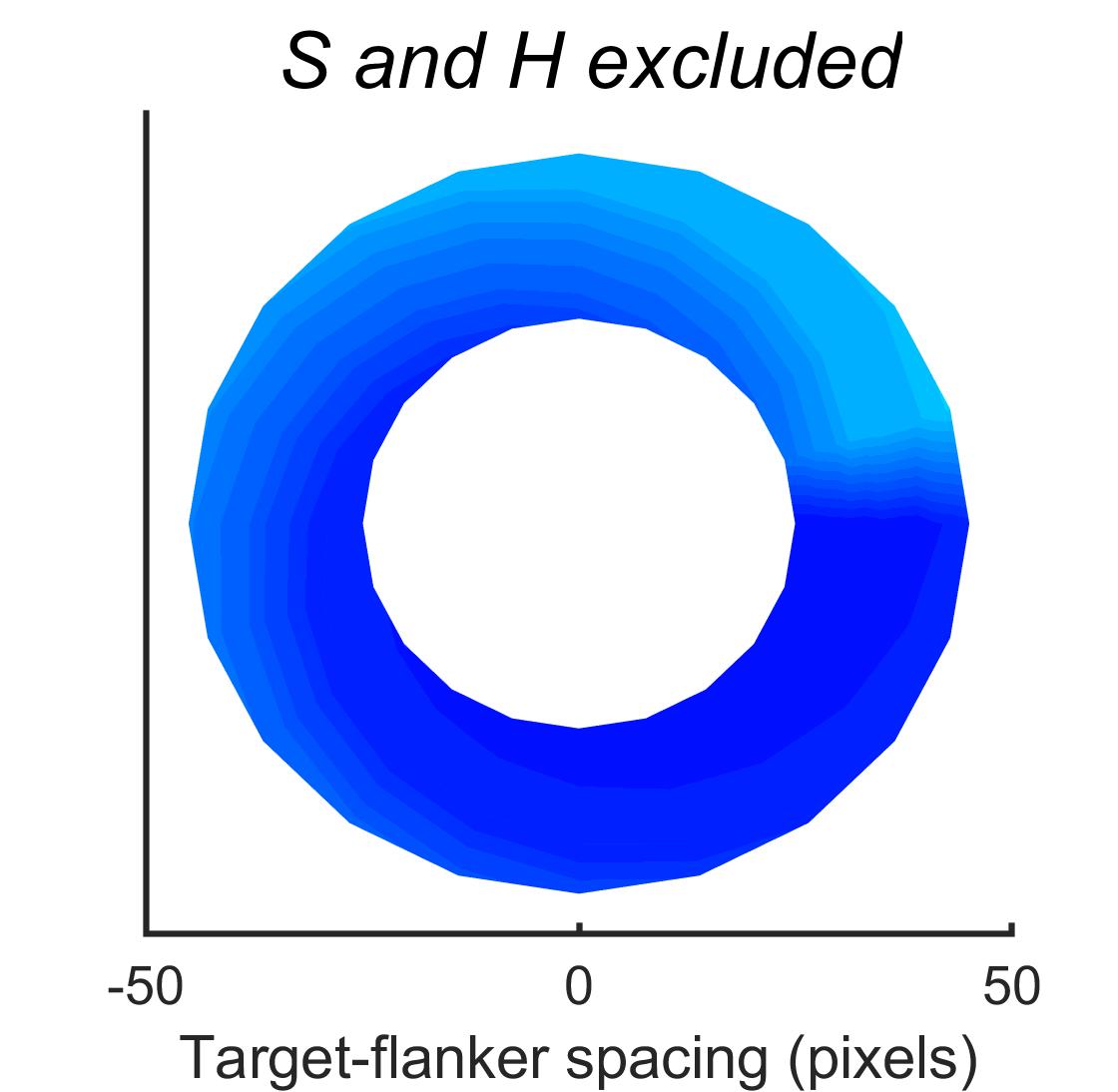}
		\end{subfigure}
		\begin{subfigure}{0.25\textwidth}
			\includegraphics[width=\textwidth]{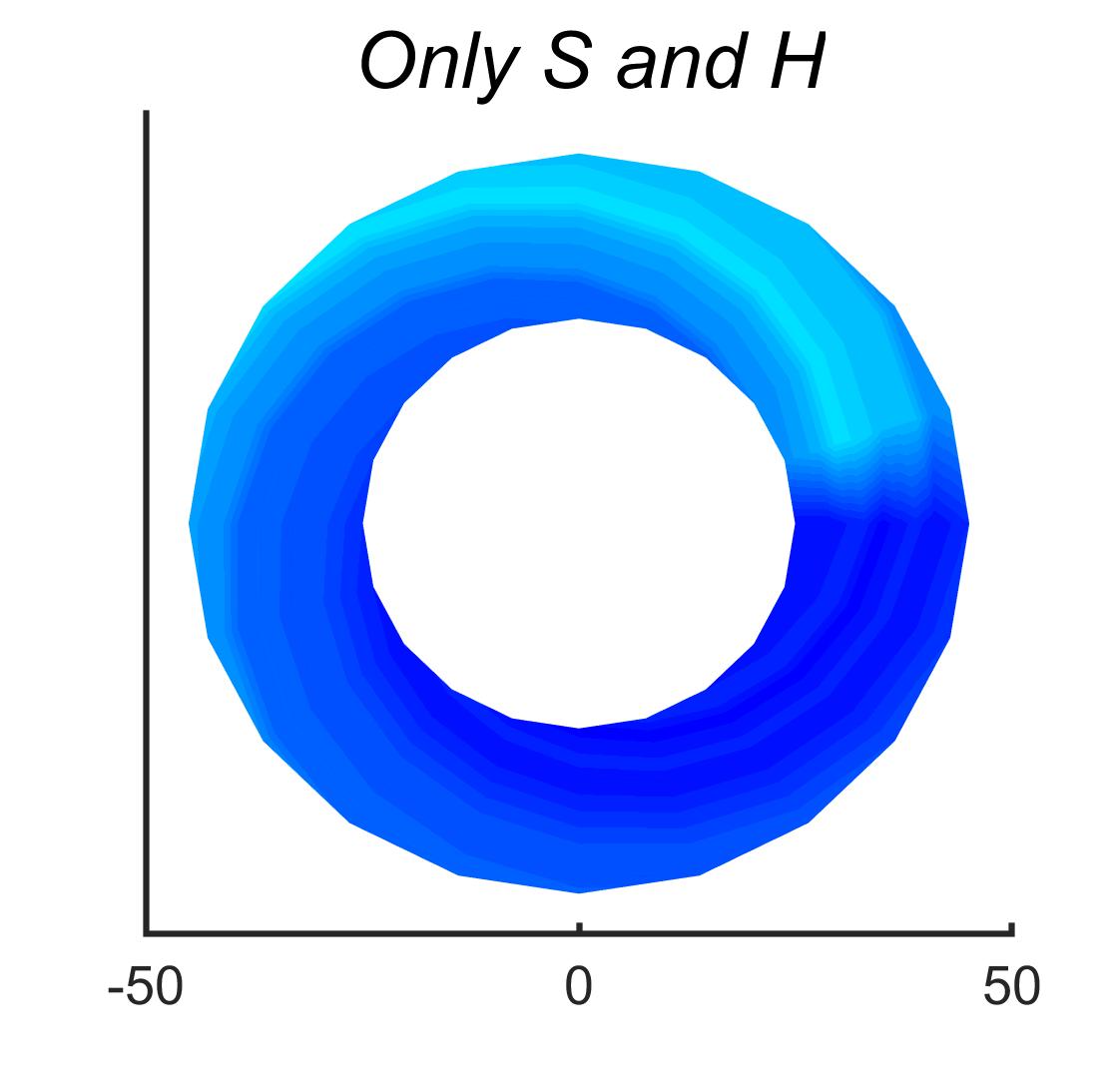}
		\end{subfigure}
		\begin{subfigure}{0.06\textwidth}
			\includegraphics[width=\textwidth]{other/colorbar.jpg}
		\end{subfigure}
		\caption{Accuracy of letter identification of the third randomly initialised small 5-layer convolutional network with single flankers ('SimpleNet 3'). Training and testing was done without acuity loss. Average accuracy without flankers was 97.25\%.}
		\label{smallrandominit3}
	\end{figure}

\begin{figure}
		\centering
		\begin{subfigure}{0.9\textwidth}
			\includegraphics[width=\textwidth]{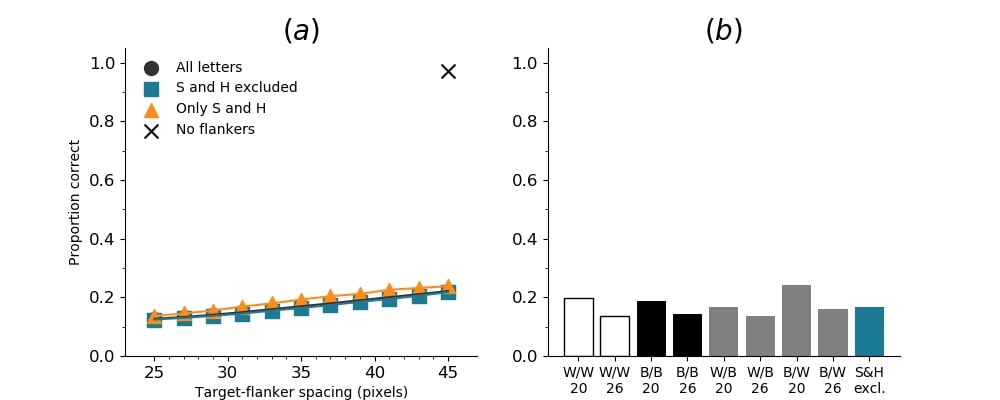}
		\end{subfigure}
		\begin{subfigure}{0.2562\textwidth}
			\includegraphics[width=\textwidth]{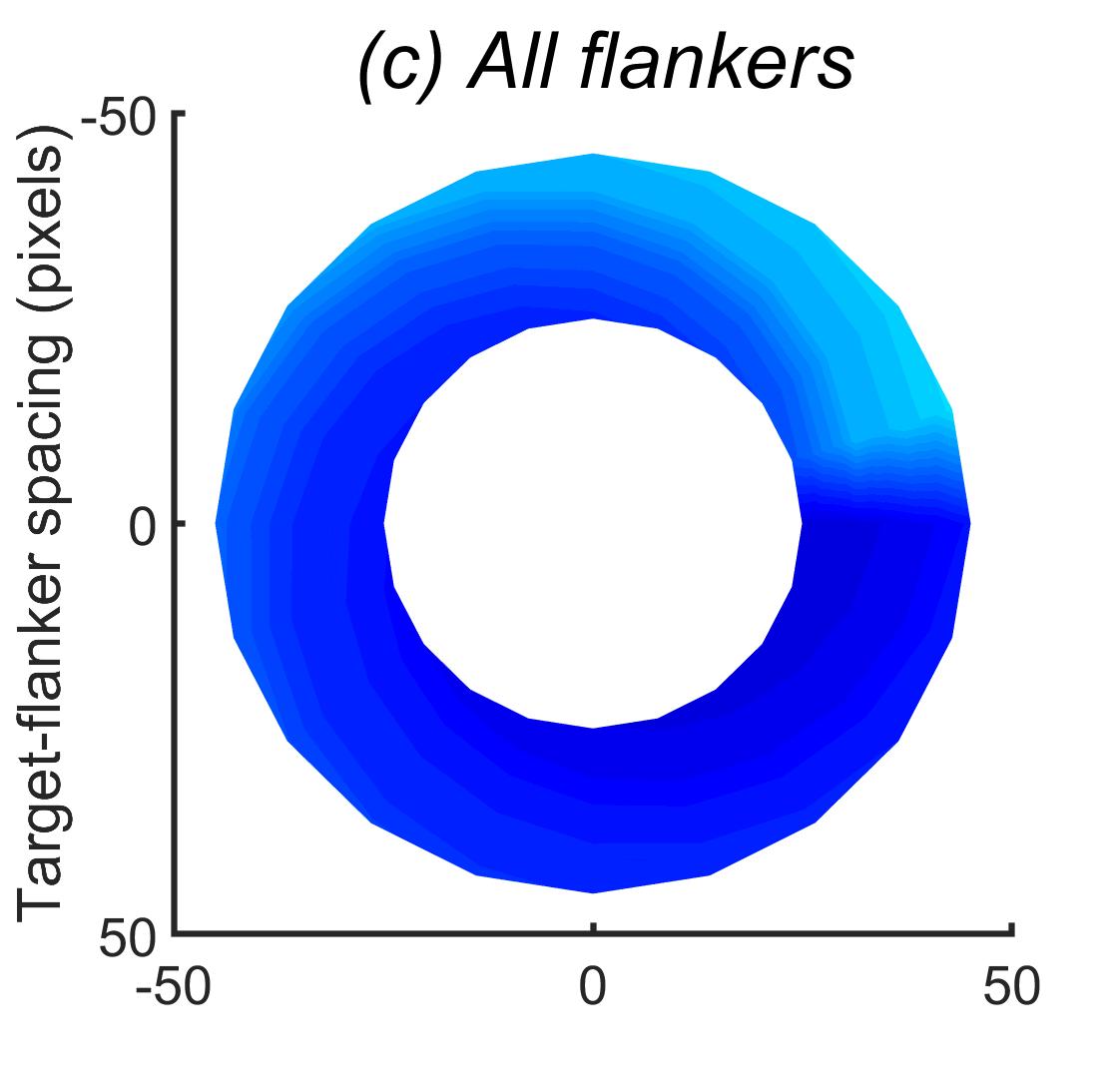}
		\end{subfigure}
		\begin{subfigure}{0.25\textwidth}
			\includegraphics[width=\textwidth]{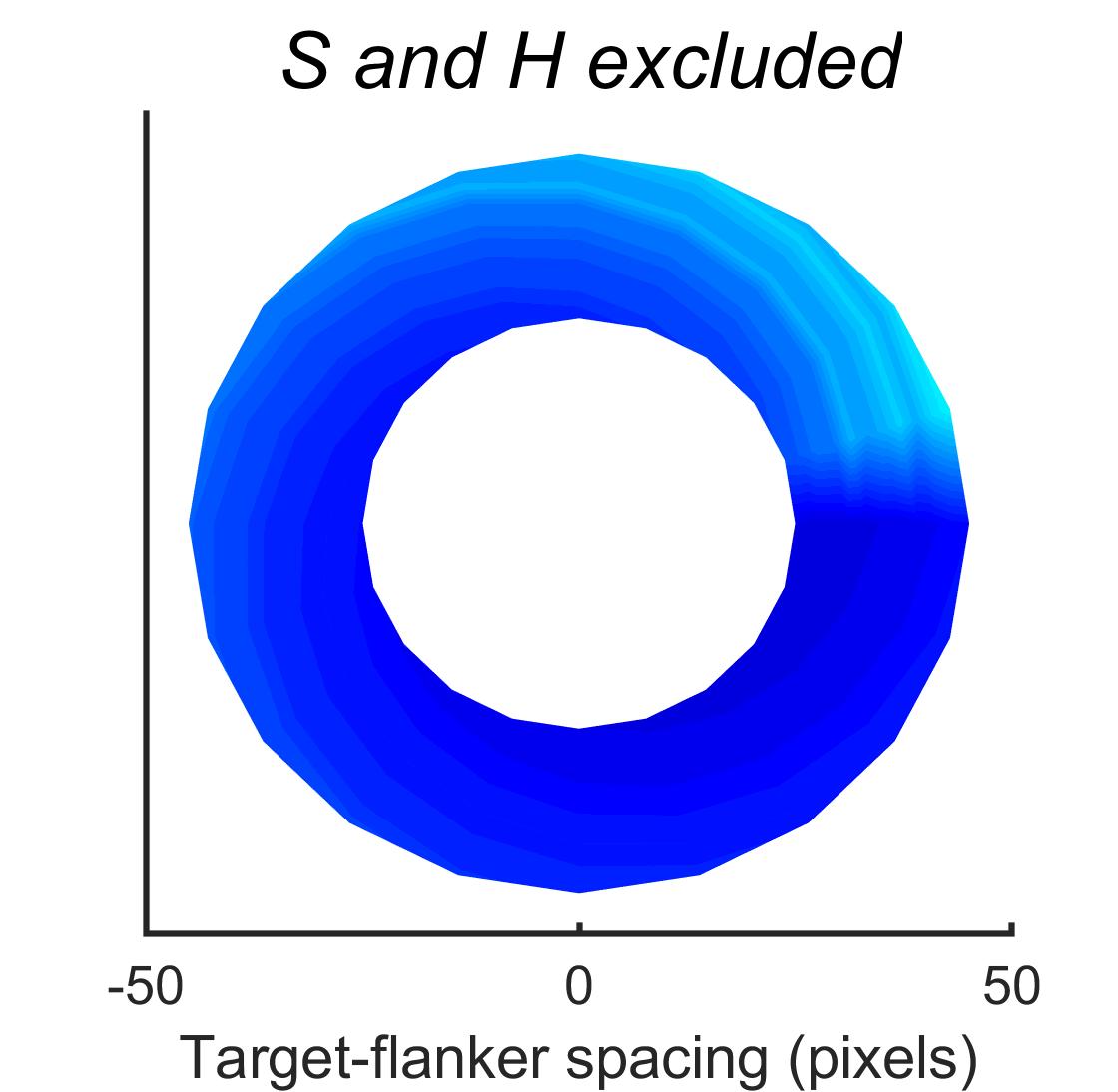}
		\end{subfigure}
		\begin{subfigure}{0.25\textwidth}
			\includegraphics[width=\textwidth]{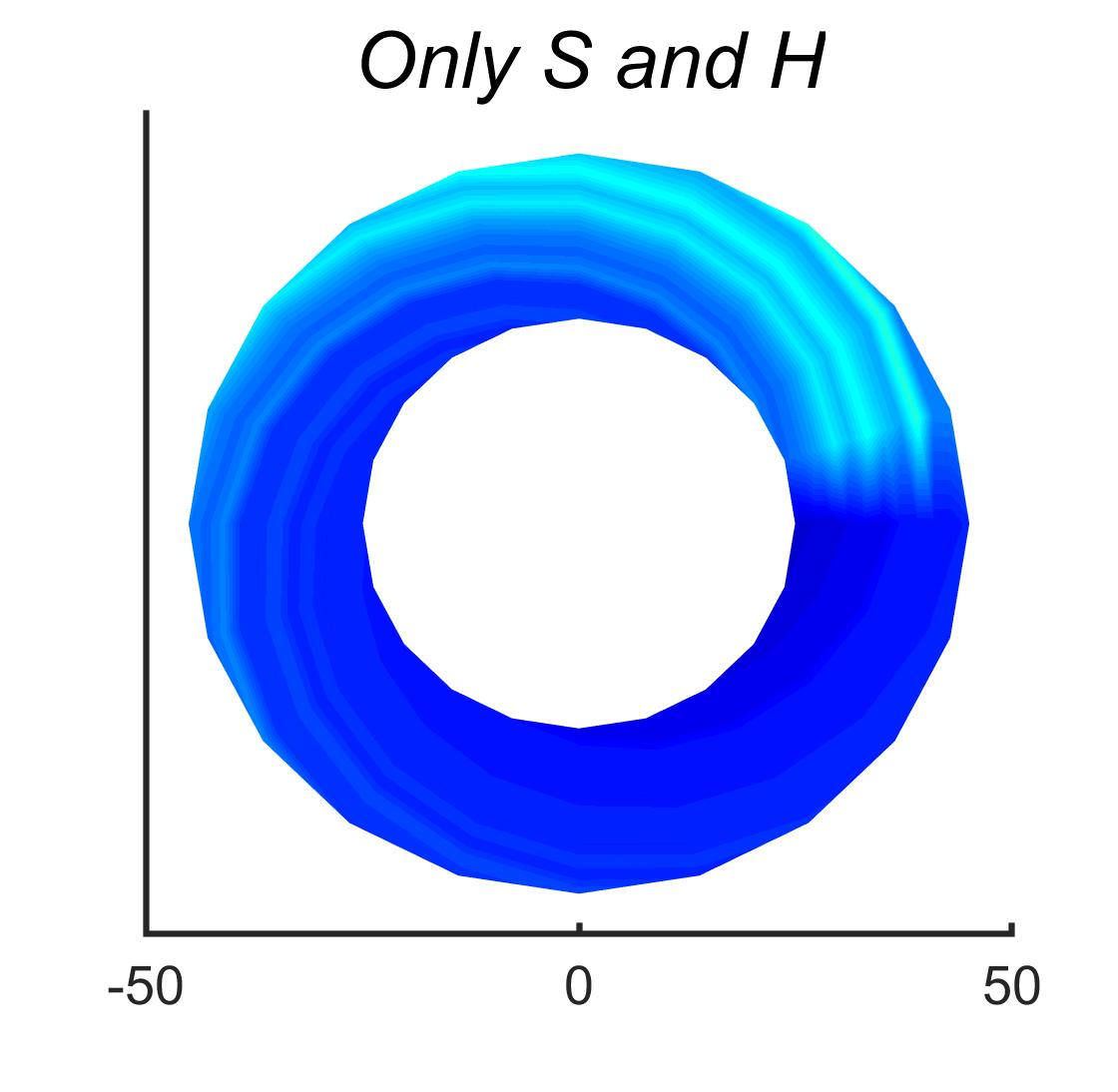}
		\end{subfigure}
		\begin{subfigure}{0.06\textwidth}
			\includegraphics[width=\textwidth]{other/colorbar.jpg}
		\end{subfigure}
		\caption{Accuracy of letter identification of the fifth randomly initialised small 5-layer convolutional network with single flankers ('SimpleNet 5'). Training and testing was done without acuity loss. Average accuracy without flankers was 97.22\%.}
		\label{smallrandominit5}
	\end{figure}

	Among humans, contrast polarity, and other cues of similarity, strongly modulates target identification. Objects similar to each other crowd more than dissimilar objects \citep{kooi_effect_1994, kennedy_chromatic_2010}. On the other hand, crowding is not sensitive to object size \citep{pelli_crowding_2004}. The strength and extent of crowding is comparable for objects of different sizes. Here, we examined if we observe the same pattern of results. In fact, we found no clear effects of size or contrast polarity. Performance was not higher for different polarity letters than it was for same polarity letters (for example, see Figure \ref{smallrandominit3} and Figure \ref{smallrandominit5}). However, they showed differences in performance for the two letter sizes. Most networks identified smaller letters than larger letters, however, the opposite pattern was noticeable in other networks. 
	
	Some networks, such as the DenseNet-121 with random initialisation (Figure \ref{drandominitsingle} were poor at detecting black targets, but were not modulated by target-flanker similarity. Other networks, such as the DenseNet-121 with random initialisation and with the targets and flankers on the right side (Figure \ref{drandominitsingleright2}) had clear difficulty distinguishing letters of a specific size and colour combination compared to the others. In general, we found no discernible patterns that applied to all models (even within a specific architecture).
	
	\subsection{Confusion between targets and flankers}
	Finally, we analysed the DenseNets' and VGG's reported output to examine whether targets were confused with flankers more often than they were confused with other letters. We found that for all single-flanker results there is little difference between the model reporting another target `at random' and the model reporting the flanker letter. On error trials, the flanker is reported 0.0125 percentage points more often than any other single letter, on average.

	This finding implies that in our experiments, the DCNNs were highly sensitive to the position of the target and that they were not prone to confuse the flanker as the target. This also rules out the hypothesis that flanker substitution contributes to crowding in DCNNs when DCNNs are trained in a simplistic manner, like it does in humans \citep{freeman_substitution_2012, hanus_quantifying_2013}. 
	
	\section{Discussion}
\begin{figure}
		\centering
		\begin{subfigure}{0.9\textwidth}
			\includegraphics[width=\textwidth]{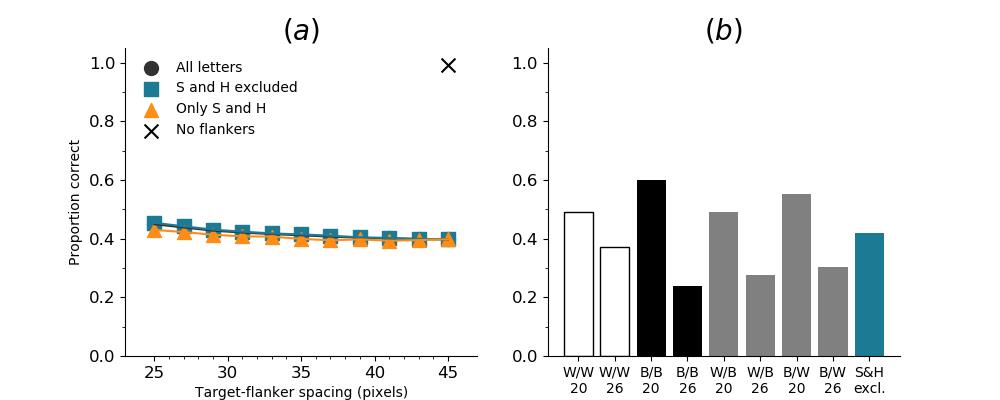}
		\end{subfigure}
		\begin{subfigure}{0.2562\textwidth}
			\includegraphics[width=\textwidth]{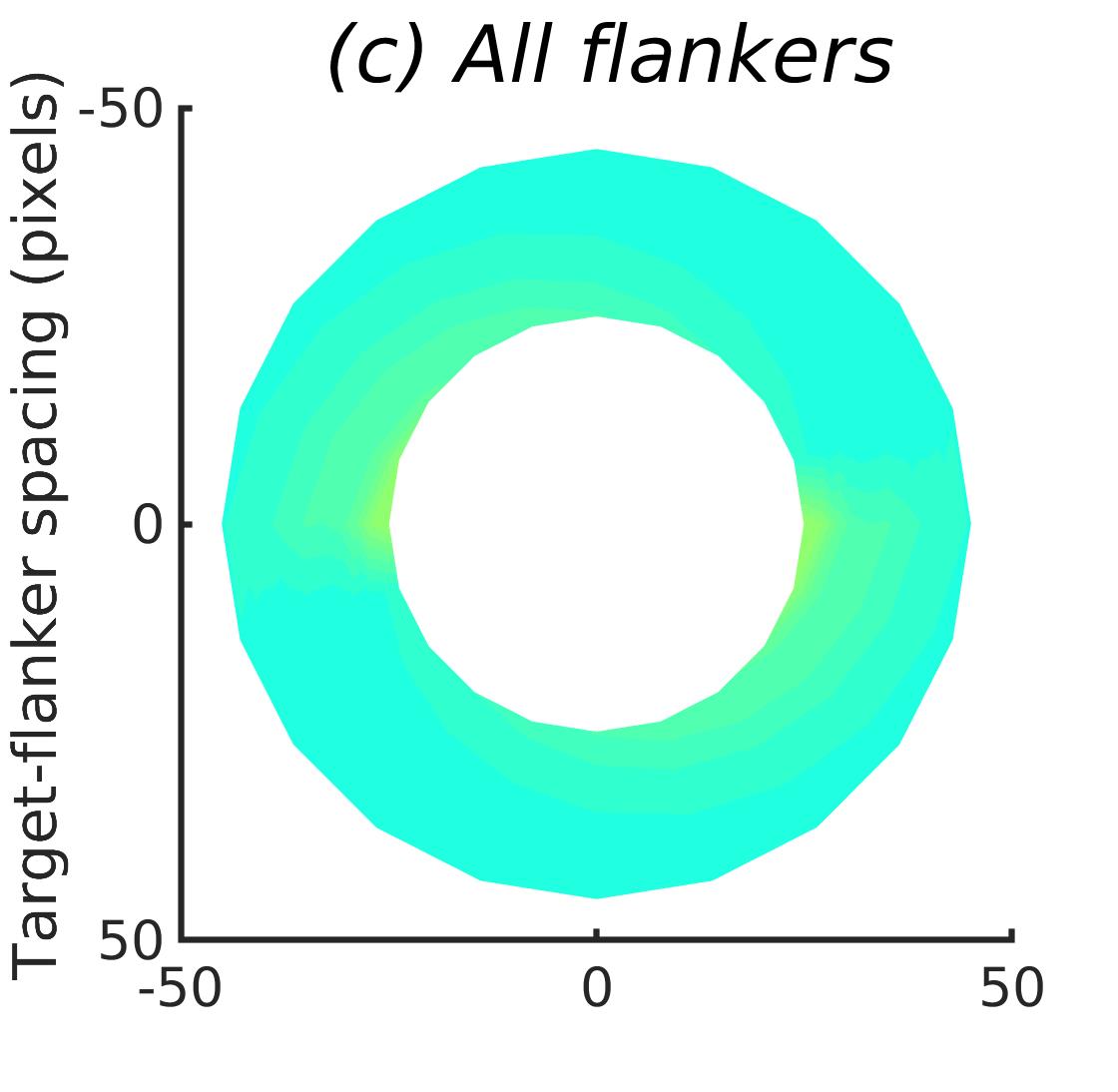}
		\end{subfigure}
		\begin{subfigure}{0.25\textwidth}
			\includegraphics[width=\textwidth]{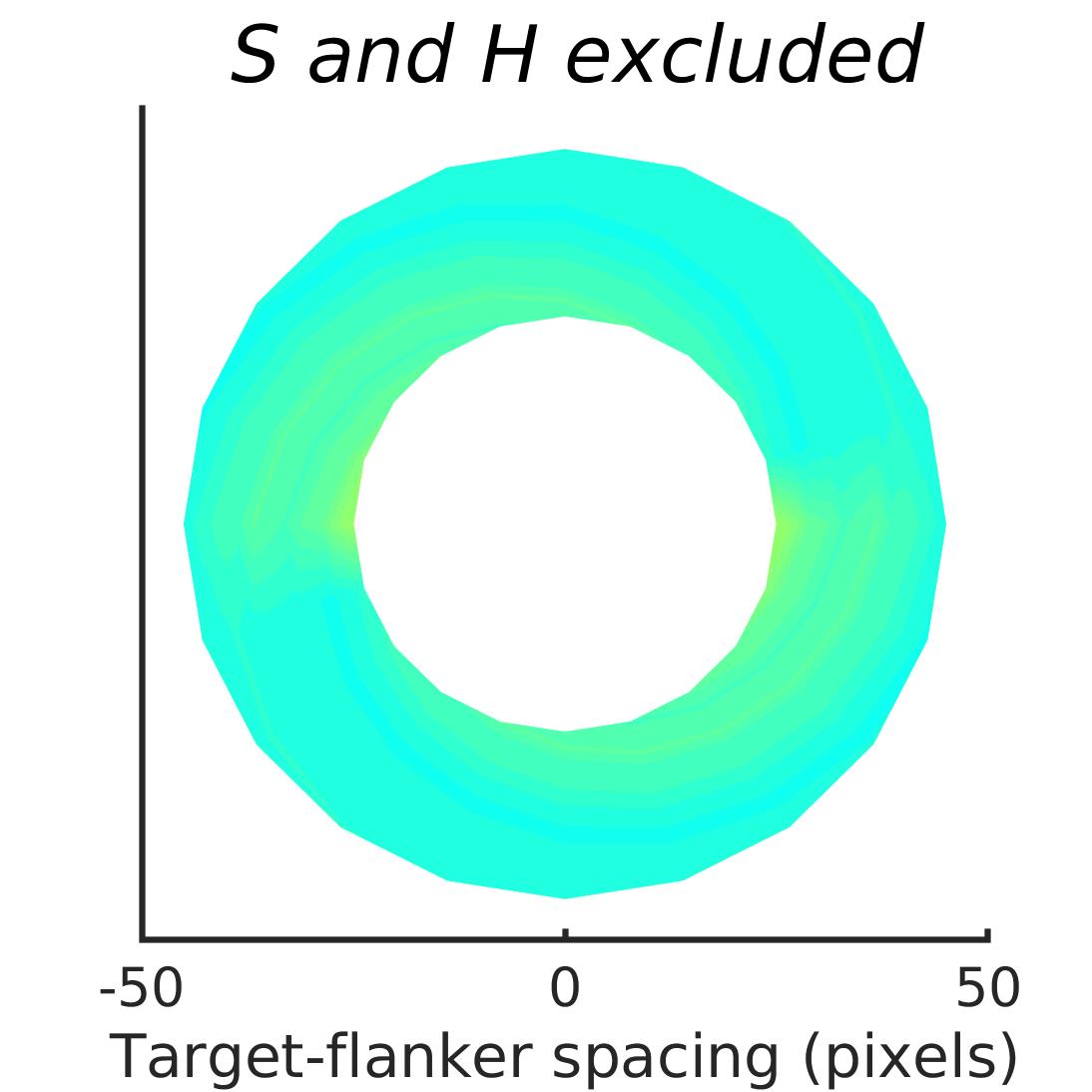}
		\end{subfigure}
		\begin{subfigure}{0.25\textwidth}
			\includegraphics[width=\textwidth]{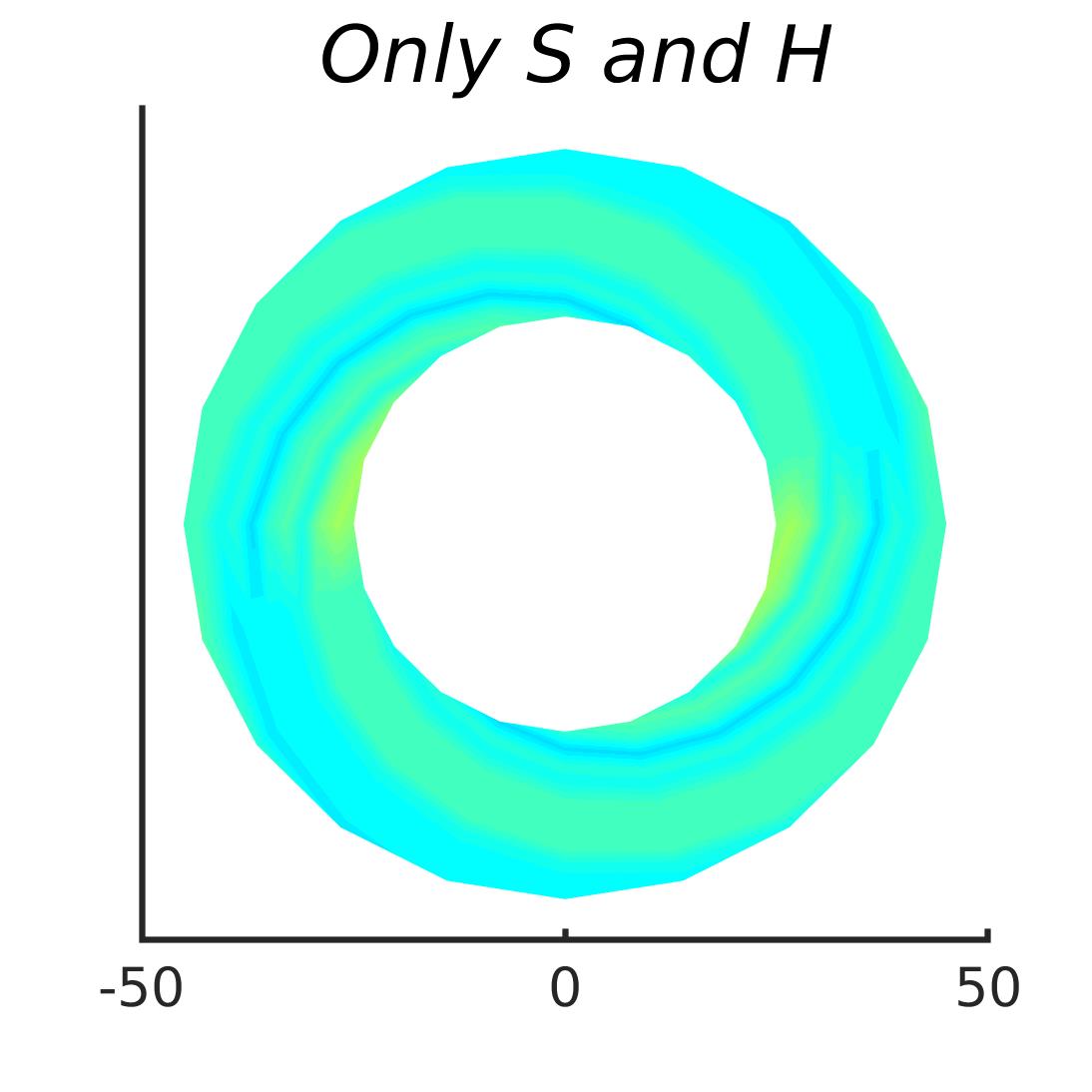}
		\end{subfigure}
		\begin{subfigure}{0.06\textwidth}
			\includegraphics[width=\textwidth]{other/colorbar.jpg}
		\end{subfigure}
		\caption{Accuracy of letter identification using the same weights as Figure \ref{dnoacuitylossimagenetsingle}, but with pair flankers. We suspect convergence issues with this test, resulting in unexpected test performance---while some experiments did not show a clear decrease in the degree of crowding with distance, this is one of the two models for which crowding increases with distance.}
		\label{dnoacuitylossimagenet}
	\end{figure}

	We investigated crowding in DCNNs and found that they follow a predictable pattern regardless of network topology, size or colour of flankers, or whether images have been reduced in acuity. Overall, we do find that flankers reduce target identification performance, demonstrating that all the networks we tested suffer from crowding. On the other hand, importantly, we found that object recognition in humans has distinctly different characteristics from those exhibited by DCNNs.
	The pattern of crowding found follows a combination of several factors:
	\begin{itemize}
		\item \textbf{Robustness to flanker interference:} We found that the SimpleNet and VGG-16 models were much more susceptible to flanker interference than the DenseNet-121. This suggests that some characteristics unique to the DenseNet (e.g., skip connections, batch normalization, or simply the prsence of more layers) causes the model to be more robust to clutter.
		\item \textbf{Distance of the flanker from the target:} In almost all experiments recognition performance for a target surrounded by known flankers strictly follows a positive relationship with distance between them. This suggests that crowding is, at least in part, caused by local pooling of information. This relationship is mild, however, and is in two models reversed (Figures \ref{drandominitsingleright2} and \ref{dnoacuitylossimagenet}).
		\item \textbf{Flanker substitution versus pooling:} Flankers near the image centre (in the lower visual field) cause more crowding than 'outer' ones, for a given spacing. These letters tend to be less subject to image degradation in our acuity loss manipulation. We suspect that this asymmetry in crowding is therefore due to local pooling, as suggested by \citet{volokitin_deep_2017}. In our experiments we also found that target-flanker confusion does not contribute much to crowding, further supporting the hypothesis that local pooling causes crowding in DCNNs under simplistic training regiments. This reason may partly explain why there is more crowding with more foveal flankers than peripheral flankers, unlike in humans \citep{petrov_asymmetries_2011, petrov_crowding_2007}. A caveat to keep in mind is that we also found that acuity loss does not drastically change the patterns of crowding, but instead its magnitude.
		\item \textbf{Amount of useful information in stimuli:} The bottom-corner position of the flanker towards the centre of the image caused most crowding in our experiments. In humans, the bottom-half of the visual field has higher resolution and lower crowding \citep{intriligator_spatial_2001}. The images of ruins and neighbourhoods we used in training and testing have a sizeable portion of their top-half contain “useless” information, possibly contributing to this effect. Additionally, when the backgrounds were vertically flipped, this bias towards the bottom-half of the image was neutralised. Further support for this argument is given by the fact that in our experiments, the ILSVRC-initialised models were subject to a higher degree of crowding. We suspect that this effect is primarily caused by the training data.
		\item \textbf{Unrecognised clutter:} When the networks are subject to flankers they do not recognise, these flankers cause effects that are unpredictable in terms of classification of the target in individual models. Often these stimuli cause a reduction in accuracy in positions and distances which do not follow a clear pattern. However, these effects may be mitigated by training and testing a model several times, and averaging results.
	\end{itemize}
	
	We also observed other dissimilarities in machine and human crowding. In many of our experiments, we found differences in the degree of crowding with differently-sized letters, violating the Bouma Law \citep{pelli_uncrowded_2008}. Additionally, black letters are not crowded more by other black letters than they are by white letters, and vice versa. In humans, this effect is clear \citep{kooi_effect_1994, kennedy_chromatic_2010}. Despite these differences, crowding in DCNNs and humans share some similarities. For example, the degree of crowding in both DCNNs and humans decreases with increased spacing between a target and its flankers \citep{bouma_interaction_1970, toet_two-dimensional_1992, pelli_crowding_2004}. The radial-tangential asymmetry also shares a resemblance with human crowding asymmetry, with radial flankers crowding the target more \citep{toet_two-dimensional_1992, petrov_asymmetries_2011}.
	
	We conclude that crowding is present in DCNNs regardless of whether a network is trained on unaltered images or acuity reduced input, and that its magnitude can be reduced by employing a more sophisticated architecture that does not rely only on convolutional, max pooling and densely connected layers. Based on the current evidence, we conjecture that local pooling is the primary source of crowding in DCNNs, and that the position in which crowding occurs is caused by the data the network has been subject to in training. As such, we suggest those who train networks to use data augmentation \citep{perez_effectiveness_2017} in order to minimise the effect of crowding.
	
	While DCNNs are loosely based on human models of object recognition, and have indeed been considered comparable, they exhibit patterns of behaviour that are substantially different from those in humans. At first glance, both demonstrate flanker induced interference. However, a closer look shows a myriad of differences. We suggest that these differences in behaviour of object recognition between humans and DCNNs are caused by one or several of many neural differences. For example, in the human visual cortex there are many different types of neurons which serve different purposes. The presented DCNNs also do not use recurrent connections---in the human visual cortex, there many recurrent connections, and these recurrent connections contribute enormously to visual processing \citep{bullier_role_2001}. The way in which the human visual system and DCNNs are built are fundamentally different, and our experiments show that they exhibit fundamentally different behaviour in object recognition tasks.
	
	\subsubsection*{Acknowledgements}
	We would like to acknowledge the use of a Tesla K40 GPU card that has been donated to Dr M. S. Baptista by Nvidia. We would also like to thank Dr Micha Elsner for helpful discussions.
	
	\bibliography{LonnqvistClarkeChakravarthi2019bib}

\begin{thebibliography}{}

\bibitem[Anstis, 1974]{anstis_letter:_1974}
Anstis, S.~M. (1974).
\newblock Letter: {A} chart demonstrating variations in acuity with retinal
  position.
\newblock {\em Vision Research}, 14(7), 589--592.

\bibitem[Berg et~al., 2007]{berg_generality_2007}
Berg, R. v.~d., Roerdink, J. B. T.~M., \& Cornelissen, F.~W. (2007).
\newblock On the generality of crowding: {Visual} crowding in size, saturation,
  and hue compared to orientation.
\newblock {\em Journal of Vision}, 7(2), 14--14.

\bibitem[Bonner \& Epstein, 2017]{bonner_coding_2017}
Bonner, M.~F. \& Epstein, R.~A. (2017).
\newblock Coding of navigational affordances in the human visual system.
\newblock {\em Proceedings of the National Academy of Sciences}, (pp.\
  201618228).

\bibitem[Bouma, 1970]{bouma_interaction_1970}
Bouma, H. (1970).
\newblock Interaction {Effects} in {Parafoveal} {Letter} {Recognition}.
\newblock {\em Nature}, 226(5241), 177--178.

\bibitem[Bullier et~al., 2001]{bullier_role_2001}
Bullier, J., Hupé, J.~M., James, A.~C., \& Girard, P. (2001).
\newblock The role of feedback connections in shaping the responses of visual
  cortical neurons.
\newblock {\em Progress in Brain Research}, 134, 193--204.

\bibitem[Cadieu et~al., 2014]{cadieu_deep_2014}
Cadieu, C.~F., Hong, H., Yamins, D. L.~K., Pinto, N., Ardila, D., Solomon,
  E.~A., Majaj, N.~J., \& DiCarlo, J.~J. (2014).
\newblock Deep {Neural} {Networks} {Rival} the {Representation} of {Primate}
  {IT} {Cortex} for {Core} {Visual} {Object} {Recognition}.
\newblock {\em PLOS Computational Biology}, 10(12), e1003963.

\bibitem[Chollet et~al., 2015]{chollet2015keras}
Chollet, F. et~al. (2015).
\newblock Keras.
\newblock \url{https://keras.io}.

\bibitem[Cichy et~al., 2017]{cichy_dynamics_2017}
Cichy, R.~M., Khosla, A., Pantazis, D., \& Oliva, A. (2017).
\newblock Dynamics of scene representations in the human brain revealed by
  magnetoencephalography and deep neural networks.
\newblock {\em NeuroImage}, 153, 346--358.

\bibitem[Cichy et~al., 2016]{cichy_comparison_2016}
Cichy, R.~M., Khosla, A., Pantazis, D., Torralba, A., \& Oliva, A. (2016).
\newblock Comparison of deep neural networks to spatio-temporal cortical
  dynamics of human visual object recognition reveals hierarchical
  correspondence.
\newblock {\em Scientific Reports}, 6(1), 1--13.

\bibitem[DiCarlo et~al., 2012]{dicarlo_how_2012}
DiCarlo, J.~J., Zoccolan, D., \& Rust, N.~C. (2012).
\newblock How does the brain solve visual object recognition?
\newblock {\em Neuron}, 73(3), 415--434.

\bibitem[Essen \& Maunsell, 1983]{essen_hierarchical_1983}
Essen, D. C.~V. \& Maunsell, J. H.~R. (1983).
\newblock Hierarchical organization and functional streams in the visual
  cortex.
\newblock {\em Trends in Neurosciences}, 6, 370--375.

\bibitem[Freeman et~al., 2012]{freeman_substitution_2012}
Freeman, J., Chakravarthi, R., \& Pelli, D.~G. (2012).
\newblock Substitution and pooling in crowding.
\newblock {\em Attention, Perception \& Psychophysics}, 74(2), 379--396.

\bibitem[Glorot \& Bengio, 2010]{glorot_understanding_2010}
Glorot, X. \& Bengio, Y. (2010).
\newblock Understanding the difﬁculty of training deep feedforward neural
  networks.
\newblock (pp.\~8).

\bibitem[Güçlü \& Gerven, 2015]{guclu_deep_2015}
Güçlü, U. \& Gerven, M. A. J.~v. (2015).
\newblock Deep {Neural} {Networks} {Reveal} a {Gradient} in the {Complexity} of
  {Neural} {Representations} across the {Ventral} {Stream}.
\newblock {\em Journal of Neuroscience}, 35(27), 10005--10014.

\bibitem[Hanus \& Vul, 2013]{hanus_quantifying_2013}
Hanus, D. \& Vul, E. (2013).
\newblock Quantifying error distributions in crowding.
\newblock {\em Journal of vision}, 13(4), 17.

\bibitem[He et~al., 2015a]{he_deep_2015}
He, K., Zhang, X., Ren, S., \& Sun, J. (2015a).
\newblock Deep {Residual} {Learning} for {Image} {Recognition}.
\newblock {\em arXiv:1512.03385 [cs]}.
\newblock arXiv: 1512.03385.

\bibitem[He et~al., 2015b]{he_resnet_2015}
He, K., Zhang, X., Ren, S., \& Sun, J. (2015b).
\newblock Deep residual learning for image recognition.
\newblock {\em CoRR}, abs/1512.03385.

\bibitem[Herzog et~al., 2015]{herzog_crowding_2015}
Herzog, M.~H., Sayim, B., Chicherov, V., \& Manassi, M. (2015).
\newblock Crowding, grouping, and object recognition: {A} matter of appearance.
\newblock {\em Journal of Vision}, 15(6).

\bibitem[Huang et~al., 2016]{huang_densely_2016}
Huang, G., Liu, Z., van~der Maaten, L., \& Weinberger, K.~Q. (2016).
\newblock Densely {Connected} {Convolutional} {Networks}.
\newblock {\em arXiv:1608.06993 [cs]}.
\newblock arXiv: 1608.06993.

\bibitem[Huckauf et~al., 1999]{huckauf_lateral_1999}
Huckauf, A., Heller, D., \& Nazir, T.~A. (1999).
\newblock Lateral masking: {Limitations} of the feature interaction account.
\newblock {\em Perception \& Psychophysics}, 61(1), 177--189.

\bibitem[Intriligator \& Cavanagh, 2001]{intriligator_spatial_2001}
Intriligator, J. \& Cavanagh, P. (2001).
\newblock The spatial resolution of visual attention.
\newblock {\em Cognitive Psychology}, 43(3), 171--216.

\bibitem[Ioffe \& Szegedy, 2015]{ioffe_batch_2015}
Ioffe, S. \& Szegedy, C. (2015).
\newblock Batch {Normalization}: {Accelerating} {Deep} {Network} {Training} by
  {Reducing} {Internal} {Covariate} {Shift}.

\bibitem[Kennedy \& Whitaker, 2010]{kennedy_chromatic_2010}
Kennedy, G.~J. \& Whitaker, D. (2010).
\newblock The chromatic selectivity of visual crowding.
\newblock {\em Journal of Vision}, 10(6), 15.

\bibitem[Khaligh-Razavi \& Kriegeskorte, 2014]{khaligh-razavi_deep_2014}
Khaligh-Razavi, S.-M. \& Kriegeskorte, N. (2014).
\newblock Deep {Supervised}, but {Not} {Unsupervised}, {Models} {May} {Explain}
  {IT} {Cortical} {Representation}.
\newblock {\em PLOS Computational Biology}, 10(11), e1003915.

\bibitem[Kheradpisheh et~al., 2016]{kheradpisheh_deep_2016}
Kheradpisheh, S.~R., Ghodrati, M., Ganjtabesh, M., \& Masquelier, T. (2016).
\newblock Deep {Networks} {Can} {Resemble} {Human} {Feed}-forward {Vision} in
  {Invariant} {Object} {Recognition}.
\newblock {\em Scientific Reports}, 6, 32672.

\bibitem[Kingma \& Ba, 2014]{kingma_adam:_2014}
Kingma, D.~P. \& Ba, J. (2014).
\newblock Adam: {A} {Method} for {Stochastic} {Optimization}.
\newblock {\em arXiv:1412.6980 [cs]}.
\newblock arXiv: 1412.6980.

\bibitem[Kooi et~al., 1994]{kooi_effect_1994}
Kooi, F.~L., Toet, A., Tripathy, S.~P., \& Levi, D.~M. (1994).
\newblock The effect of similarity and duration on spatial interaction in
  peripheral vision.
\newblock {\em Spatial Vision}, 8(2), 255--279.

\bibitem[Krizhevsky et~al., 2012]{Krizhevsky:2012:ICD:2999134.2999257}
Krizhevsky, A., Sutskever, I., \& Hinton, G.~E. (2012).
\newblock Imagenet classification with deep convolutional neural networks.
\newblock In {\em Proceedings of the 25th International Conference on Neural
  Information Processing Systems - Volume 1}, NIPS'12  (pp.\ 1097--1105).  USA:
  Curran Associates Inc.

\bibitem[Levi, 2008]{levi_crowding--essential_2008}
Levi, D.~M. (2008).
\newblock Crowding--an essential bottleneck for object recognition: a
  mini-review.
\newblock {\em Vision Research}, 48(5), 635--654.

\bibitem[Livne \& Sagi, 2007]{livne_configuration_2007}
Livne, T. \& Sagi, D. (2007).
\newblock Configuration influence on crowding.
\newblock {\em Journal of Vision}, 7(2), 4.1--12.

\bibitem[Manassi \& Whitney, 2018]{manassi_multi-level_2018}
Manassi, M. \& Whitney, D. (2018).
\newblock Multi-level {Crowding} and the {Paradox} of {Object} {Recognition} in
  {Clutter}.
\newblock {\em Current biology: CB}, 28(3), R127--R133.

\bibitem[Mnih et~al., 2014]{mnih_recurrent_2014}
Mnih, V., Heess, N., \& Graves, A. (2014).
\newblock Recurrent {Models} of {Visual} {Attention}.
\newblock (pp.\~9).

\bibitem[NKriegeskorte, 2017]{nkriegeskorte_coarser_2017}
NKriegeskorte, A. (2017).
\newblock Do coarser spatial patterns represent coarser categories in visual
  cortex?

\bibitem[Pelli et~al., 2004]{pelli_crowding_2004}
Pelli, D.~G., Palomares, M., \& Majaj, N.~J. (2004).
\newblock Crowding is unlike ordinary masking: {Distinguishing} feature
  integration from detection.
\newblock {\em Journal of Vision}, 4(12), 12--12.

\bibitem[Pelli \& Tillman, 2008]{pelli_uncrowded_2008}
Pelli, D.~G. \& Tillman, K.~A. (2008).
\newblock The uncrowded window of object recognition.
\newblock {\em Nature neuroscience}, 11(10), 1129--1135.

\bibitem[Perez \& Wang, 2017]{perez_effectiveness_2017}
Perez, L. \& Wang, J. (2017).
\newblock The {Effectiveness} of {Data} {Augmentation} in {Image}
  {Classification} using {Deep} {Learning}.
\newblock {\em arXiv:1712.04621 [cs]}.
\newblock arXiv: 1712.04621.

\bibitem[Petrov \& Meleshkevich, 2011]{petrov_asymmetries_2011}
Petrov, Y. \& Meleshkevich, O. (2011).
\newblock Asymmetries and idiosyncratic hot spots in crowding.
\newblock {\em Vision Research}, 51(10), 1117--1123.

\bibitem[Petrov et~al., 2007]{petrov_crowding_2007}
Petrov, Y., Popple, A.~V., \& McKee, S.~P. (2007).
\newblock Crowding and surround suppression: {Not} to be confused.
\newblock {\em Journal of vision}, 7(2), 12.1--12.9.

\bibitem[Schrimpf et~al., 2018]{SchrimpfKubilius2018BrainScore}
Schrimpf, M., Kubilius, J., Hong, H., Majaj, N.~J., Rajalingham, R., Issa,
  E.~B., Kar, K., Bashivan, P., Prescott-Roy, J., Schmidt, K., Yamins, D.
  L.~K., \& DiCarlo, J.~J. (2018).
\newblock Brain-score: Which artificial neural network for object recognition
  is most brain-like?
\newblock {\em bioRxiv preprint}.

\bibitem[Simonyan \& Zisserman, 2014]{simonyan_very_2014}
Simonyan, K. \& Zisserman, A. (2014).
\newblock Very {Deep} {Convolutional} {Networks} for {Large}-{Scale} {Image}
  {Recognition}.

\bibitem[Song et~al., 2014]{song_double_2014}
Song, S., Levi, D.~M., \& Pelli, D.~G. (2014).
\newblock A double dissociation of the acuity and crowding limits to letter
  identification, and the promise of improved visual screening.
\newblock {\em Journal of Vision}, 14(5), 3.

\bibitem[Strasburger \& Malania, 2013]{strasburger_source_2013}
Strasburger, H. \& Malania, M. (2013).
\newblock Source confusion is a major cause of crowding.
\newblock {\em Journal of Vision}, 13(1), 24--24.

\bibitem[Szegedy et~al., 2014]{szegedy_going_2014}
Szegedy, C., Liu, W., Jia, Y., Sermanet, P., Reed, S., Anguelov, D., Erhan, D.,
  Vanhoucke, V., \& Rabinovich, A. (2014).
\newblock Going {Deeper} with {Convolutions}.
\newblock {\em arXiv:1409.4842 [cs]}.
\newblock arXiv: 1409.4842.

\bibitem[Toet \& Levi, 1992]{toet_two-dimensional_1992}
Toet, A. \& Levi, D.~M. (1992).
\newblock The two-dimensional shape of spatial interaction zones in the
  parafovea.
\newblock {\em Vision Research}, 32(7), 1349--1357.

\bibitem[Ullman, 2007]{ullman_object_2007}
Ullman, S. (2007).
\newblock Object recognition and segmentation by a fragment-based hierarchy.
\newblock {\em Trends in Cognitive Sciences}, 11(2), 58--64.

\bibitem[Volokitin et~al., 2017]{volokitin_deep_2017}
Volokitin, A., Roig, G., \& Poggio, T.~A. (2017).
\newblock Do {Deep} {Neural} {Networks} {Suffer} from {Crowding}?
\newblock (pp.\~11).

\bibitem[Wallace \& Tjan, 2011]{wallace_object_2011}
Wallace, J.~M. \& Tjan, B.~S. (2011).
\newblock Object crowding.
\newblock {\em Journal of Vision}, 11(6).

\bibitem[Xu et~al., 2015]{xu_empirical_2015}
Xu, B., Wang, N., Chen, T., \& Li, M. (2015).
\newblock Empirical {Evaluation} of {Rectified} {Activations} in
  {Convolutional} {Network}.
\newblock {\em arXiv:1505.00853 [cs, stat]}.
\newblock arXiv: 1505.00853.

\bibitem[Yamins \& DiCarlo, 2016]{yamins_using_2016}
Yamins, D. L.~K. \& DiCarlo, J.~J. (2016).
\newblock Using goal-driven deep learning models to understand sensory cortex.
\newblock {\em Nature Neuroscience}, 19(3), 356--365.

\bibitem[Zeiler \& Fergus, 2013]{zeiler_visualizing_2013}
Zeiler, M.~D. \& Fergus, R. (2013).
\newblock Visualizing and {Understanding} {Convolutional} {Networks}.
\newblock {\em arXiv:1311.2901 [cs]}.
\newblock arXiv: 1311.2901.

\bibitem[Zhang et~al., 2017]{zhang_interpretable_2017}
Zhang, Q., Wu, Y.~N., \& Zhu, S.-C. (2017).
\newblock Interpretable {Convolutional} {Neural} {Networks}.
\newblock {\em arXiv:1710.00935 [cs]}.
\newblock arXiv: 1710.00935.

\end{thebibliography}
	
	\newpage
	\section*{Supplementary Materials}
	\subsection*{Supplementary figures}
	\beginsupplementaryfigures
	
	\begin{figure}[h!]
		\centering
		\begin{subfigure}{0.9\textwidth}
			\includegraphics[width=\textwidth]{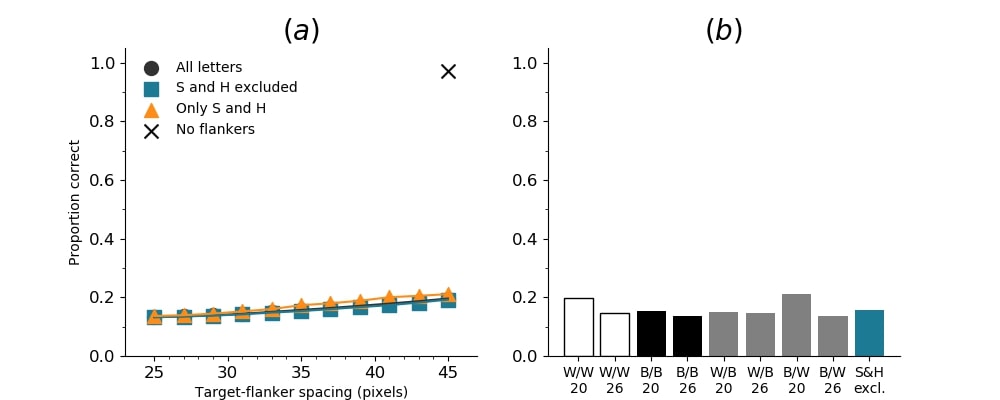}
		\end{subfigure}
		\begin{subfigure}{0.2562\textwidth}
			\includegraphics[width=\textwidth]{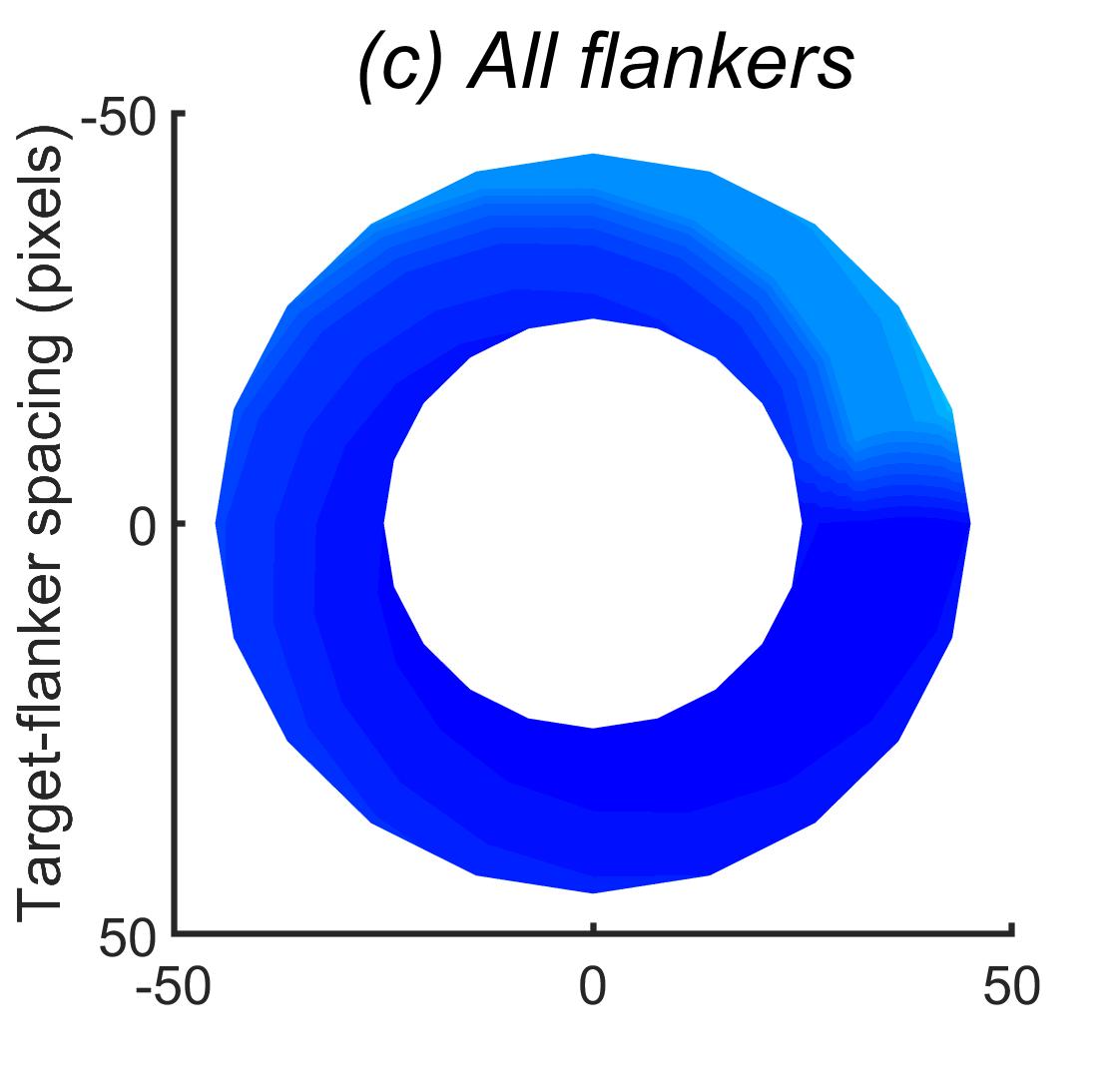}
		\end{subfigure}
		\begin{subfigure}{0.25\textwidth}
			\includegraphics[width=\textwidth]{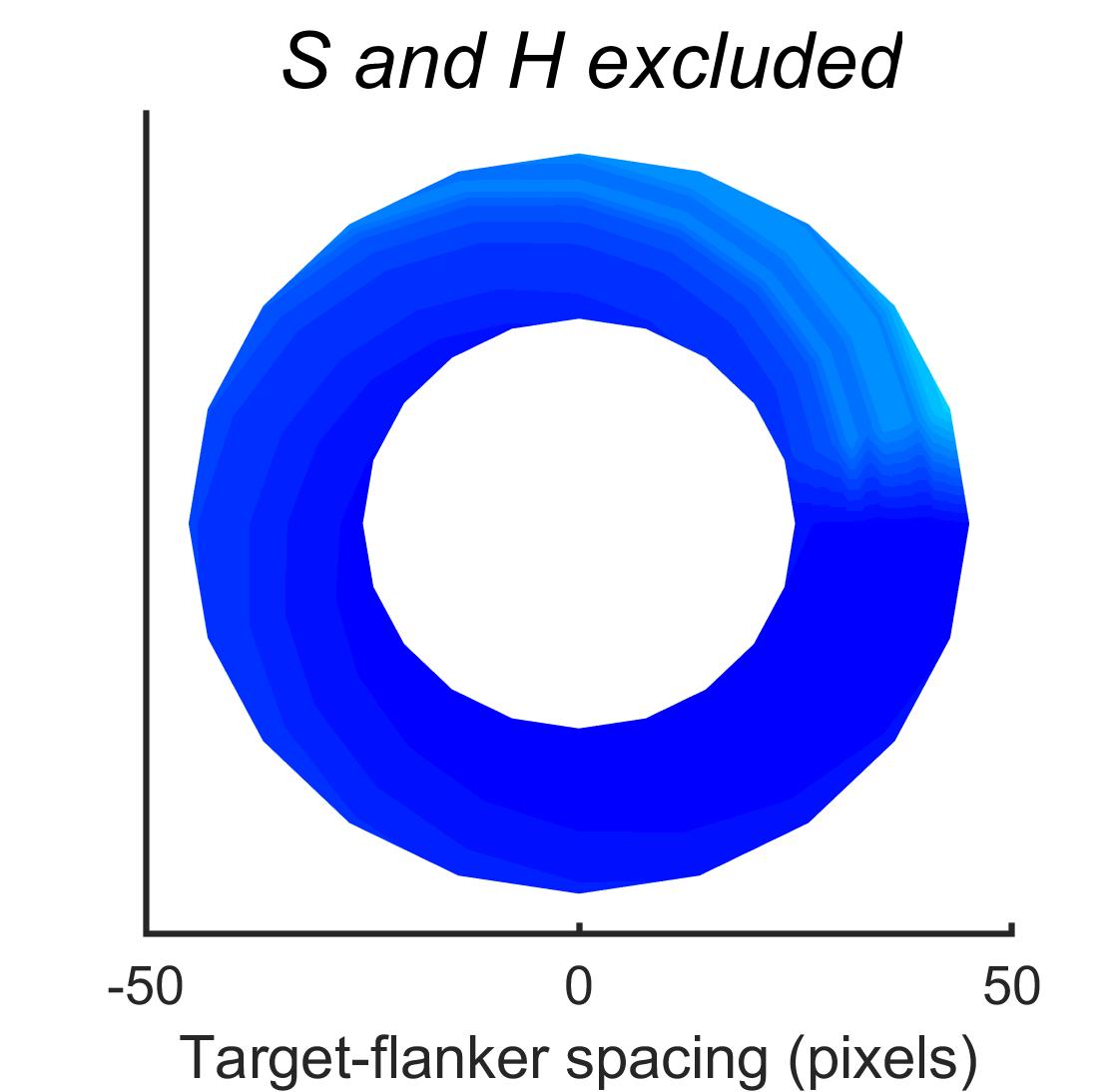}
		\end{subfigure}
		\begin{subfigure}{0.25\textwidth}
			\includegraphics[width=\textwidth]{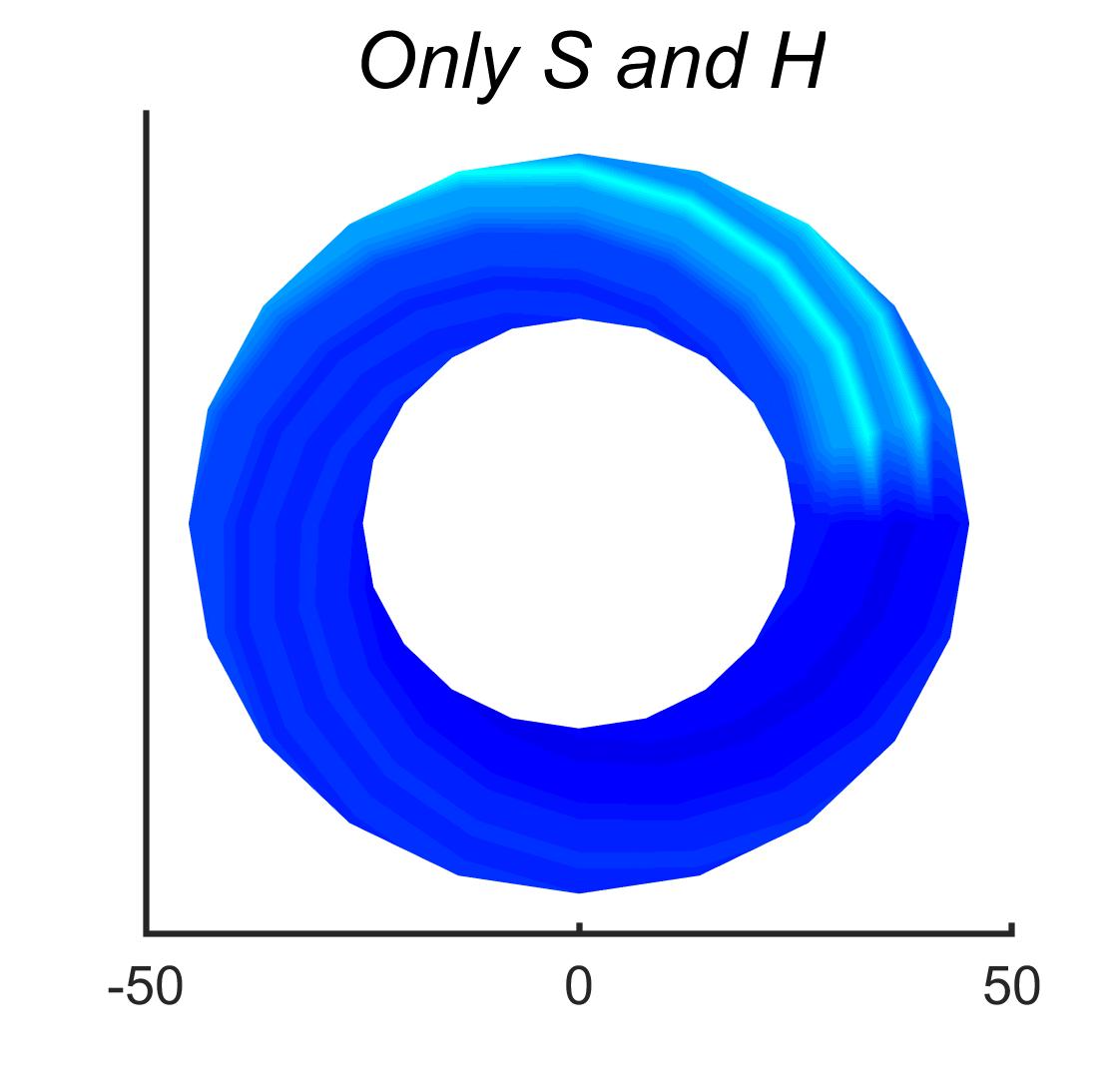}
		\end{subfigure}
		\begin{subfigure}{0.06\textwidth}
			\includegraphics[width=\textwidth]{other/colorbar.jpg}
		\end{subfigure}
		\caption{Accuracy of letter identification of the first randomly initialised small 5-layer convolutional network with single flankers ('SimpleNet 1'). Training and testing was done without acuity loss. Average accuracy without flankers was 97.16\%.}
		\label{smallrandominit1}
	\end{figure}
	
	\begin{figure}
		\centering
		\begin{subfigure}{0.9\textwidth}
			\includegraphics[width=\textwidth]{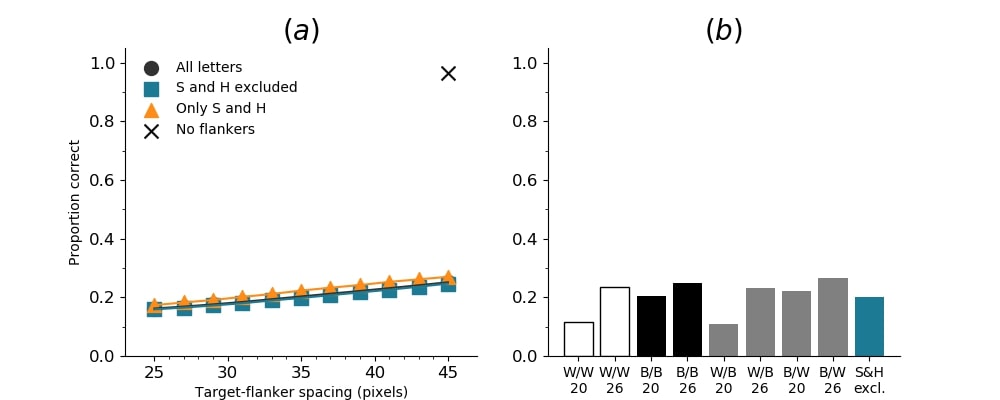}
		\end{subfigure}
		\begin{subfigure}{0.2562\textwidth}
			\includegraphics[width=\textwidth]{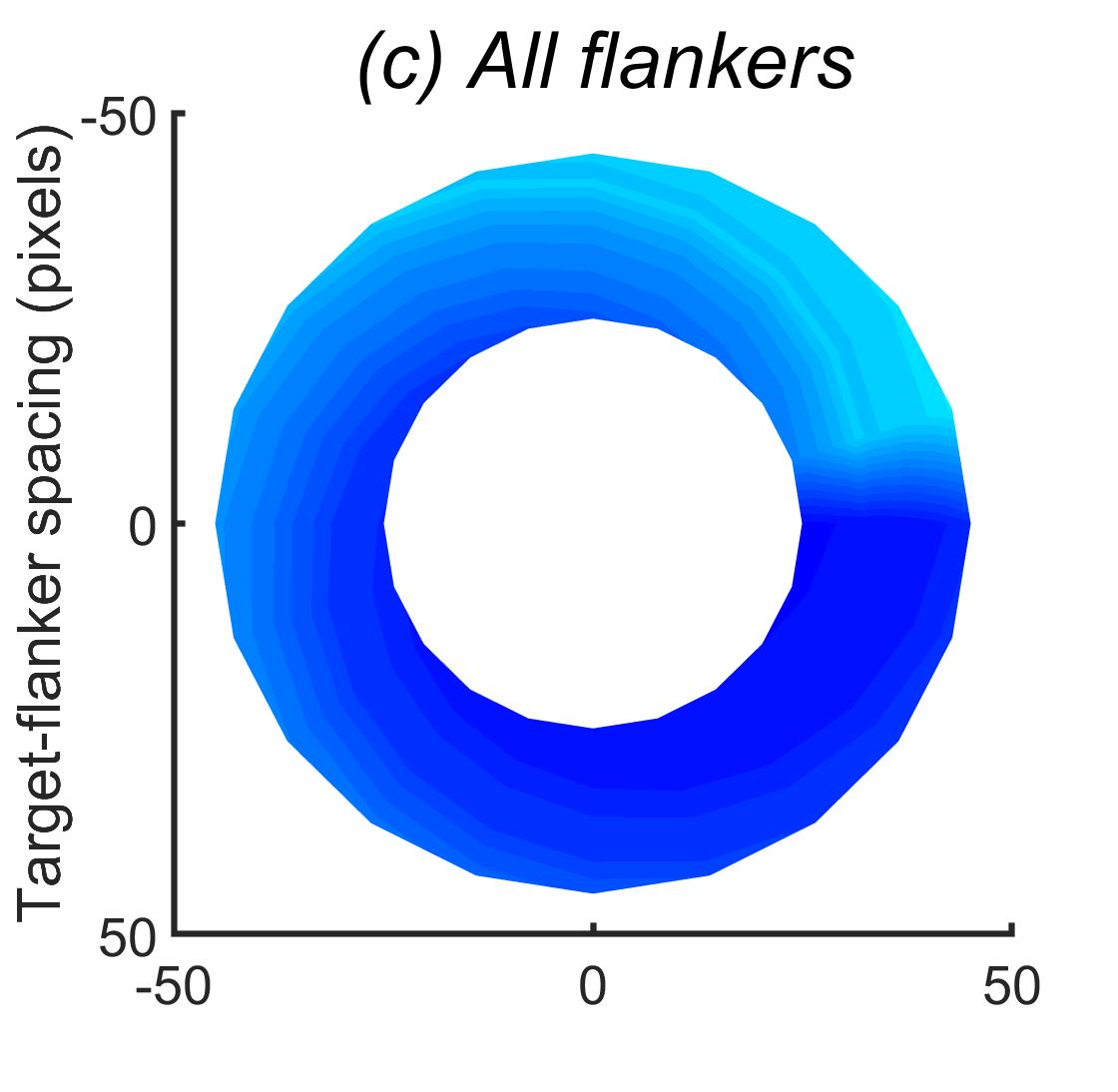}
		\end{subfigure}
		\begin{subfigure}{0.25\textwidth}
			\includegraphics[width=\textwidth]{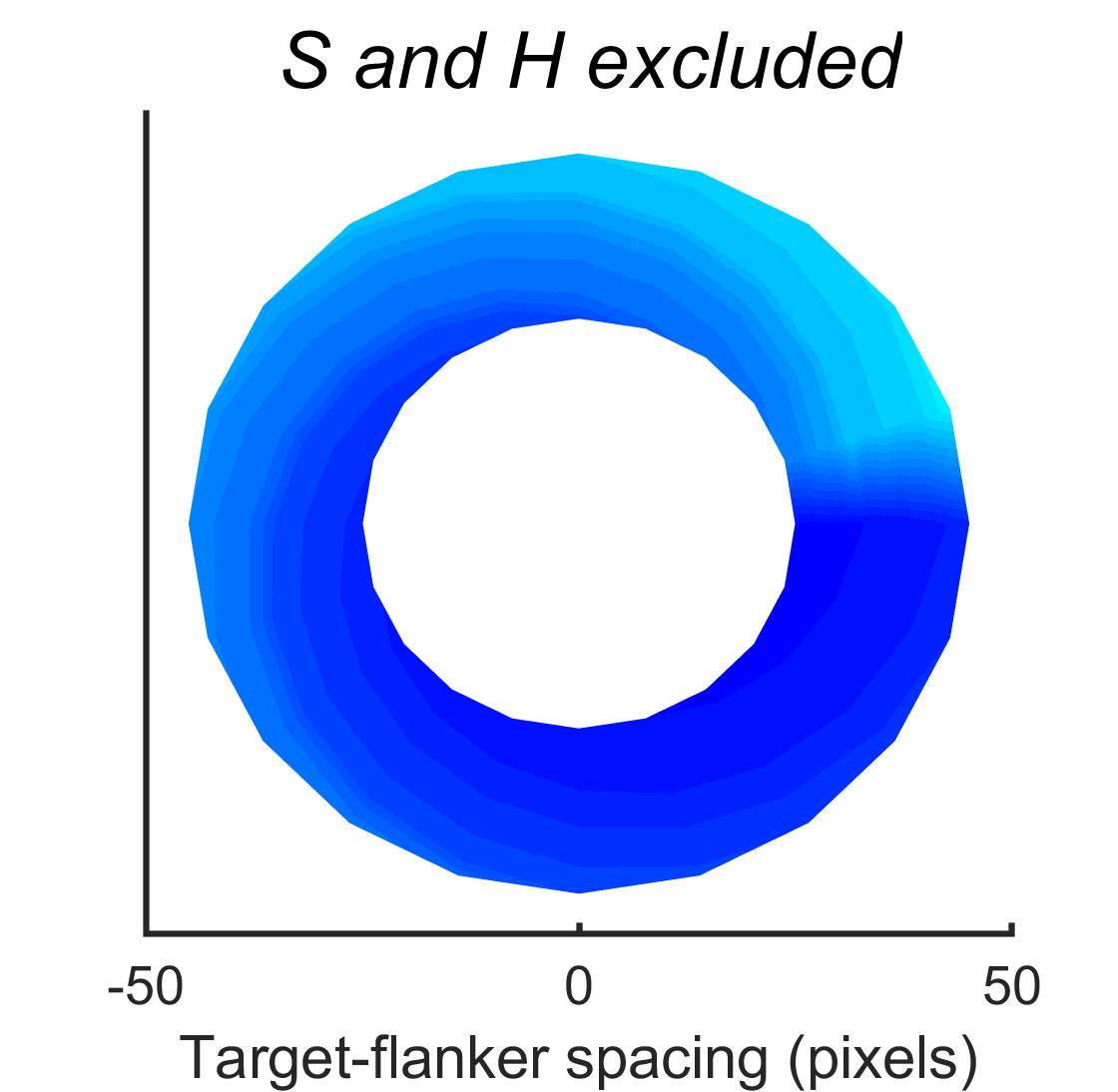}
		\end{subfigure}
		\begin{subfigure}{0.25\textwidth}
			\includegraphics[width=\textwidth]{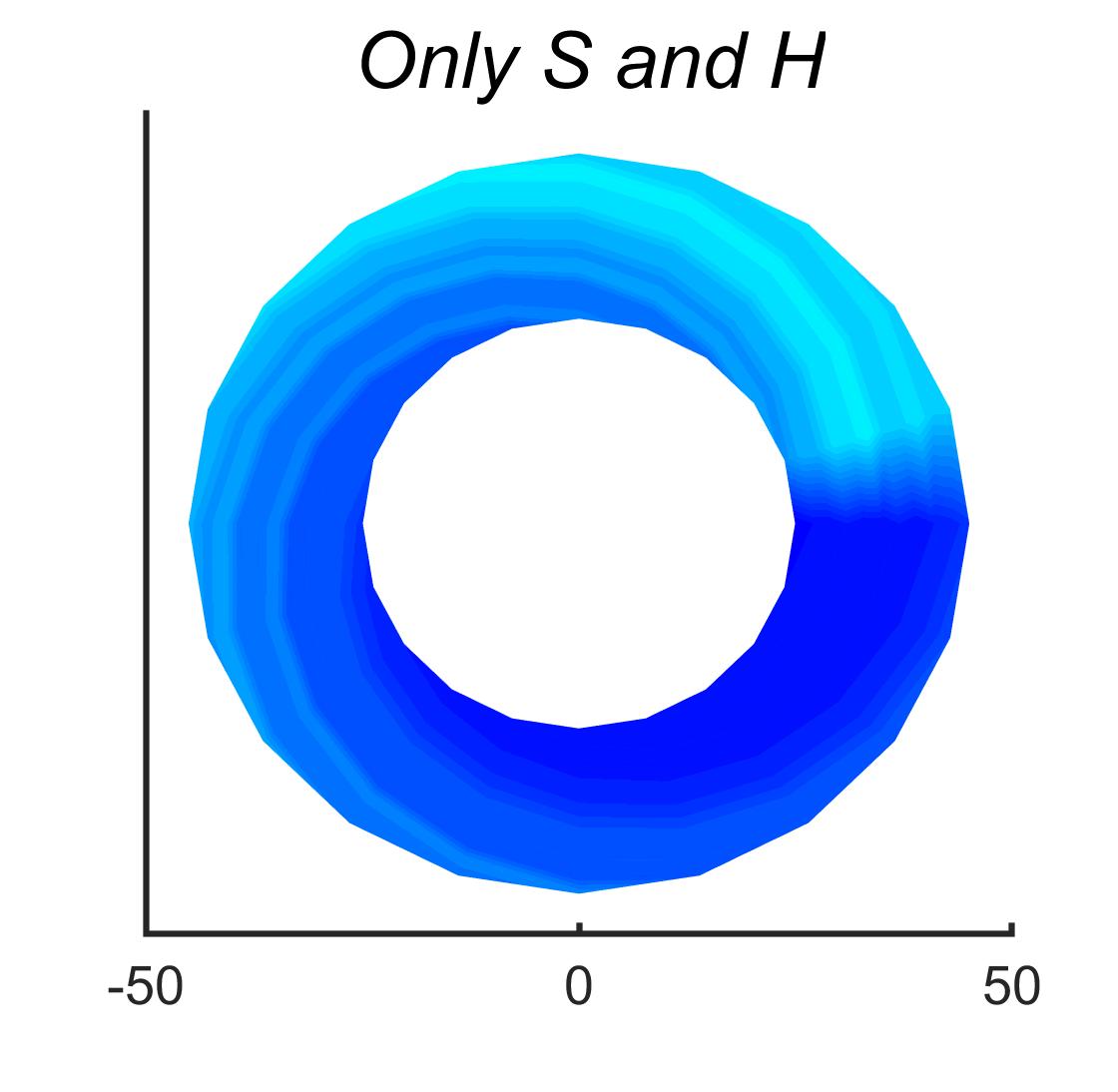}
		\end{subfigure}
		\begin{subfigure}{0.06\textwidth}
			\includegraphics[width=\textwidth]{other/colorbar.jpg}
		\end{subfigure}
		\caption{Accuracy of letter identification of the second randomly initialised small 5-layer convolutional network with single flankers ('SimpleNet 2'). Training and testing was done without acuity loss. Average accuracy without flankers was 96.62\%.}
		\label{smallrandominit2}
	\end{figure}
	
	\begin{figure}
		\centering
		\begin{subfigure}{0.9\textwidth}
			\includegraphics[width=\textwidth]{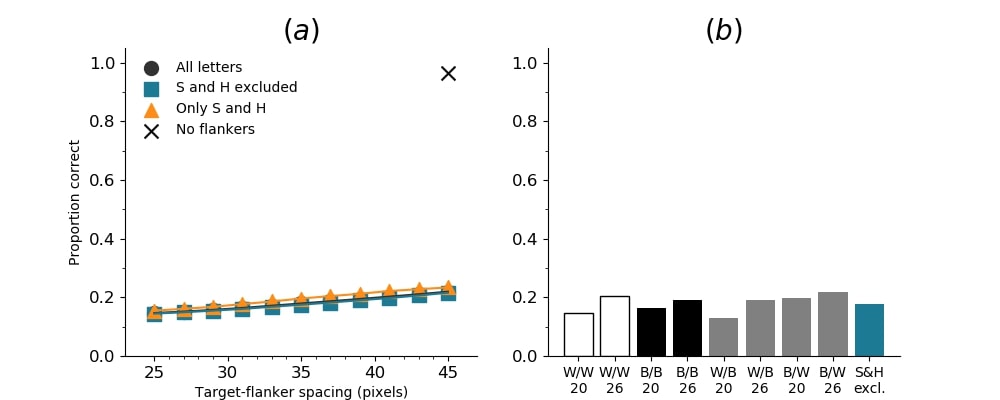}
		\end{subfigure}
		\begin{subfigure}{0.2562\textwidth}
			\includegraphics[width=\textwidth]{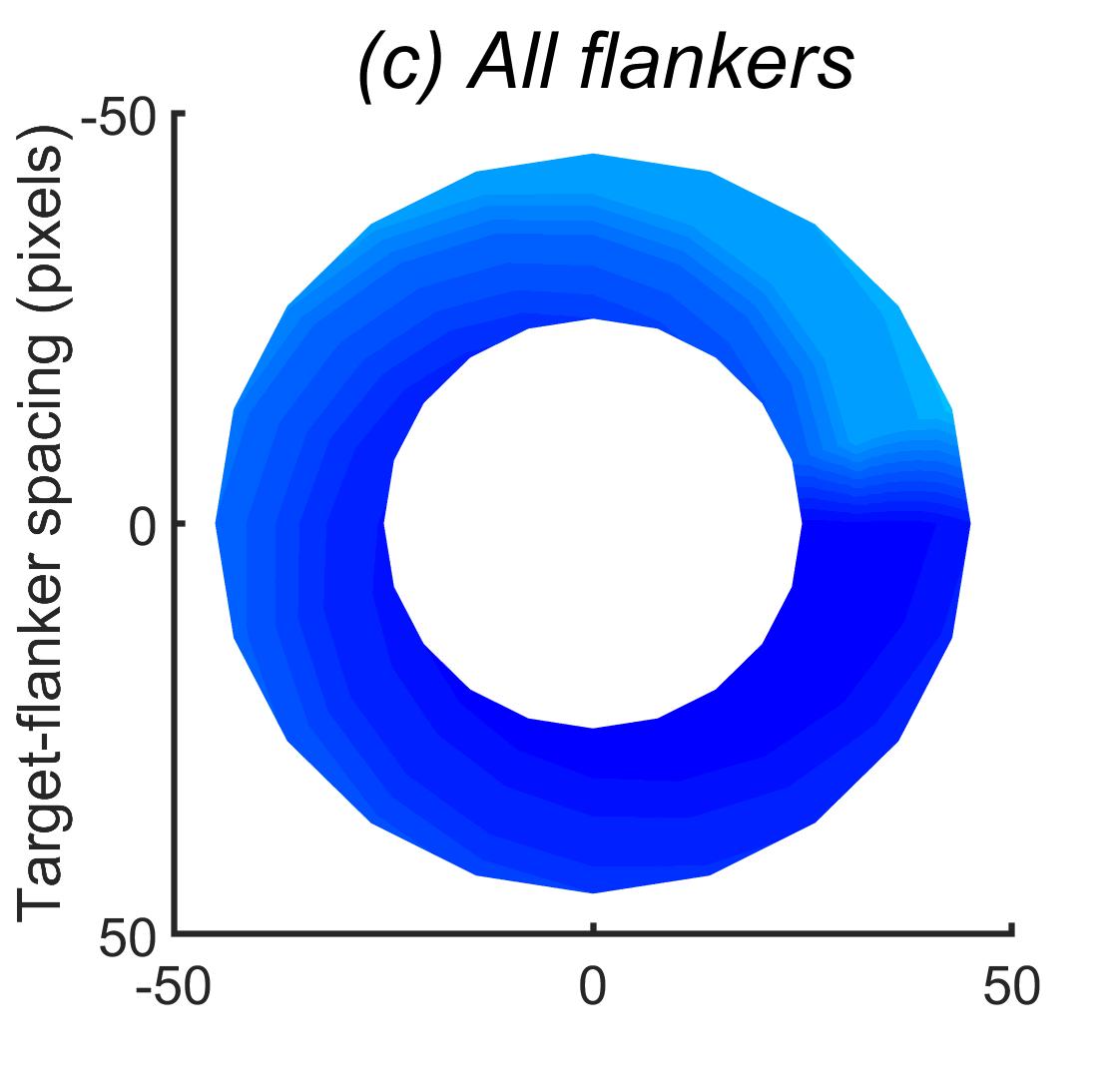}
		\end{subfigure}
		\begin{subfigure}{0.25\textwidth}
			\includegraphics[width=\textwidth]{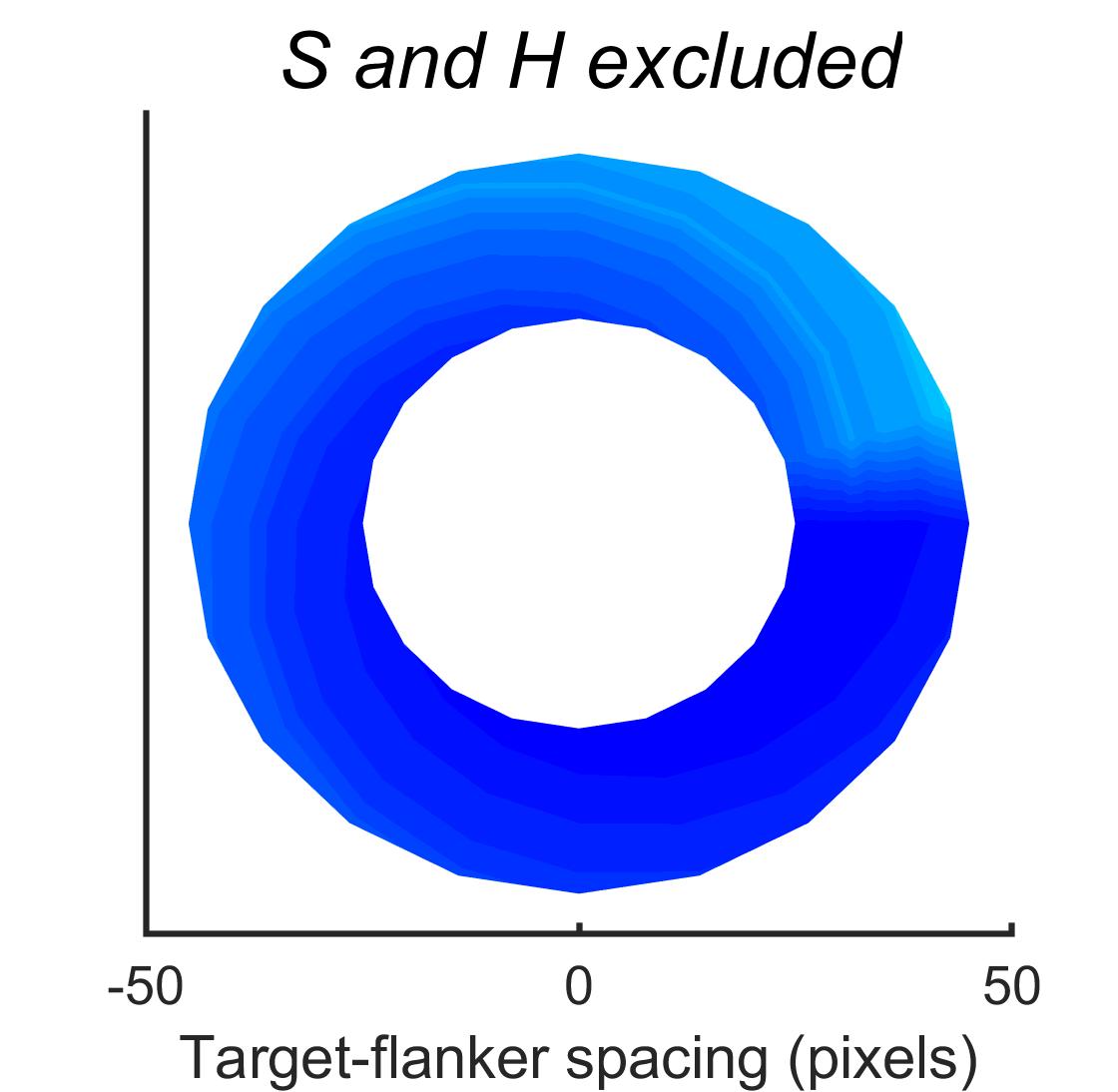}
		\end{subfigure}
		\begin{subfigure}{0.25\textwidth}
			\includegraphics[width=\textwidth]{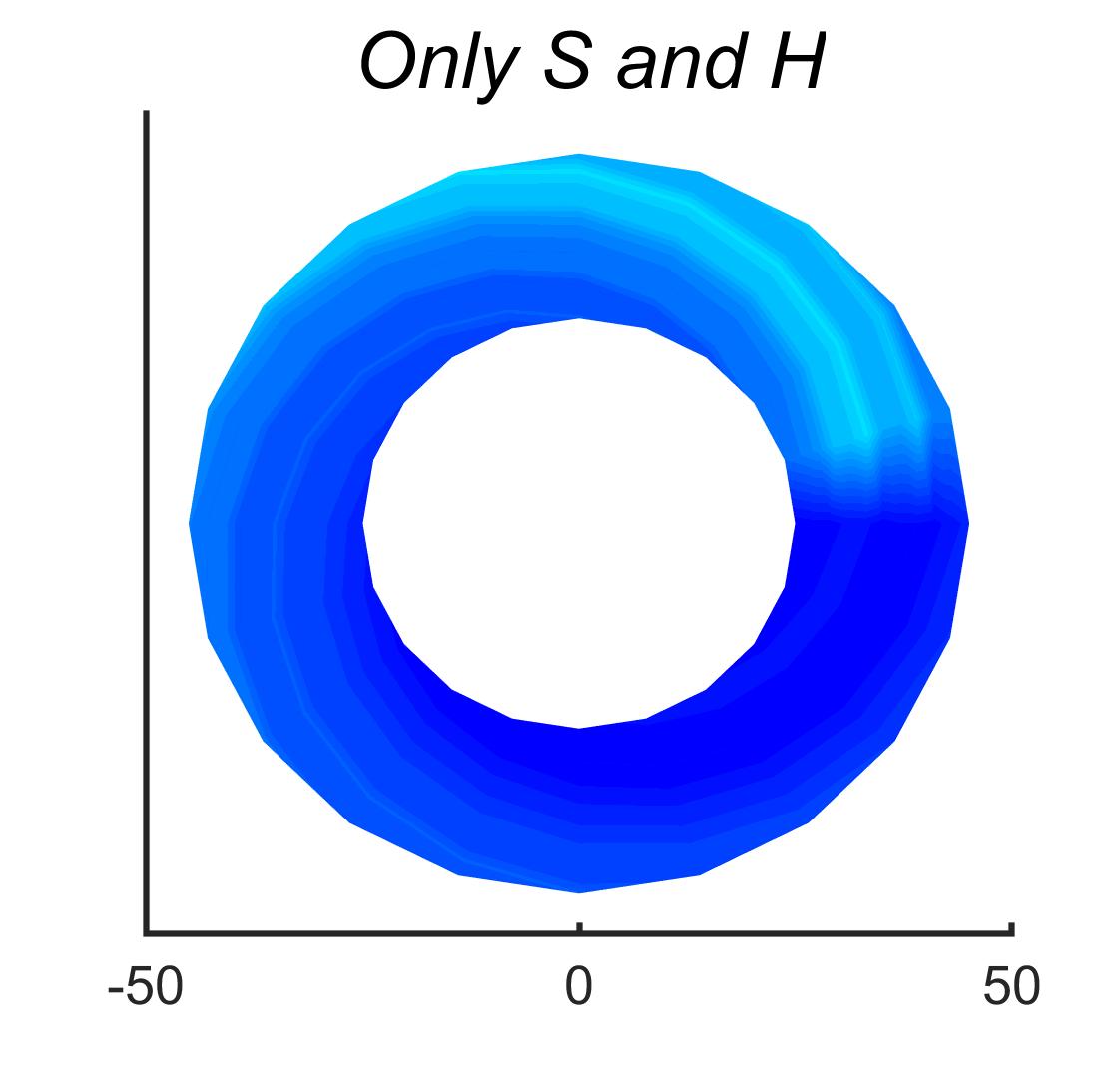}
		\end{subfigure}
		\begin{subfigure}{0.06\textwidth}
			\includegraphics[width=\textwidth]{other/colorbar.jpg}
		\end{subfigure}
		\caption{Accuracy of letter identification of the fourth randomly initialised small 5-layer convolutional network with single flankers ('SimpleNet 4'). Training and testing was done without acuity loss. Average accuracy without flankers was 96.59\%.}
		\label{smallrandominit4}
	\end{figure}
	
	\begin{figure}
		\centering
		\begin{subfigure}{0.9\textwidth}
			\includegraphics[width=\textwidth]{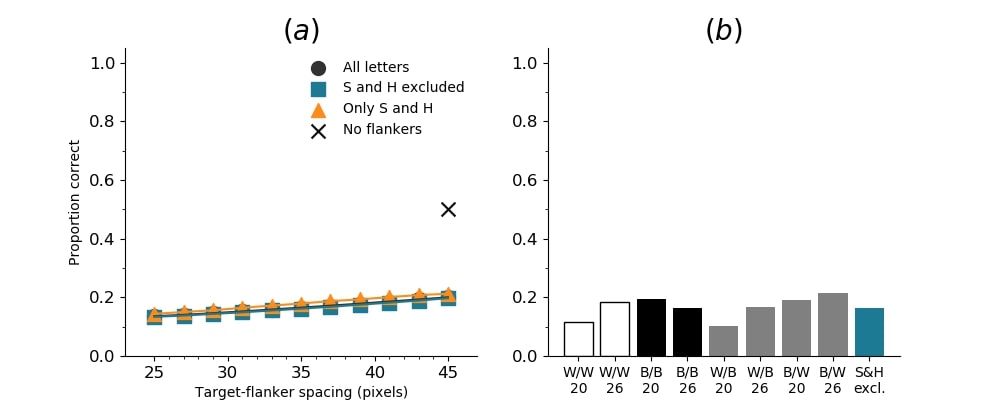}
		\end{subfigure}
		\begin{subfigure}{0.2562\textwidth}
			\includegraphics[width=\textwidth]{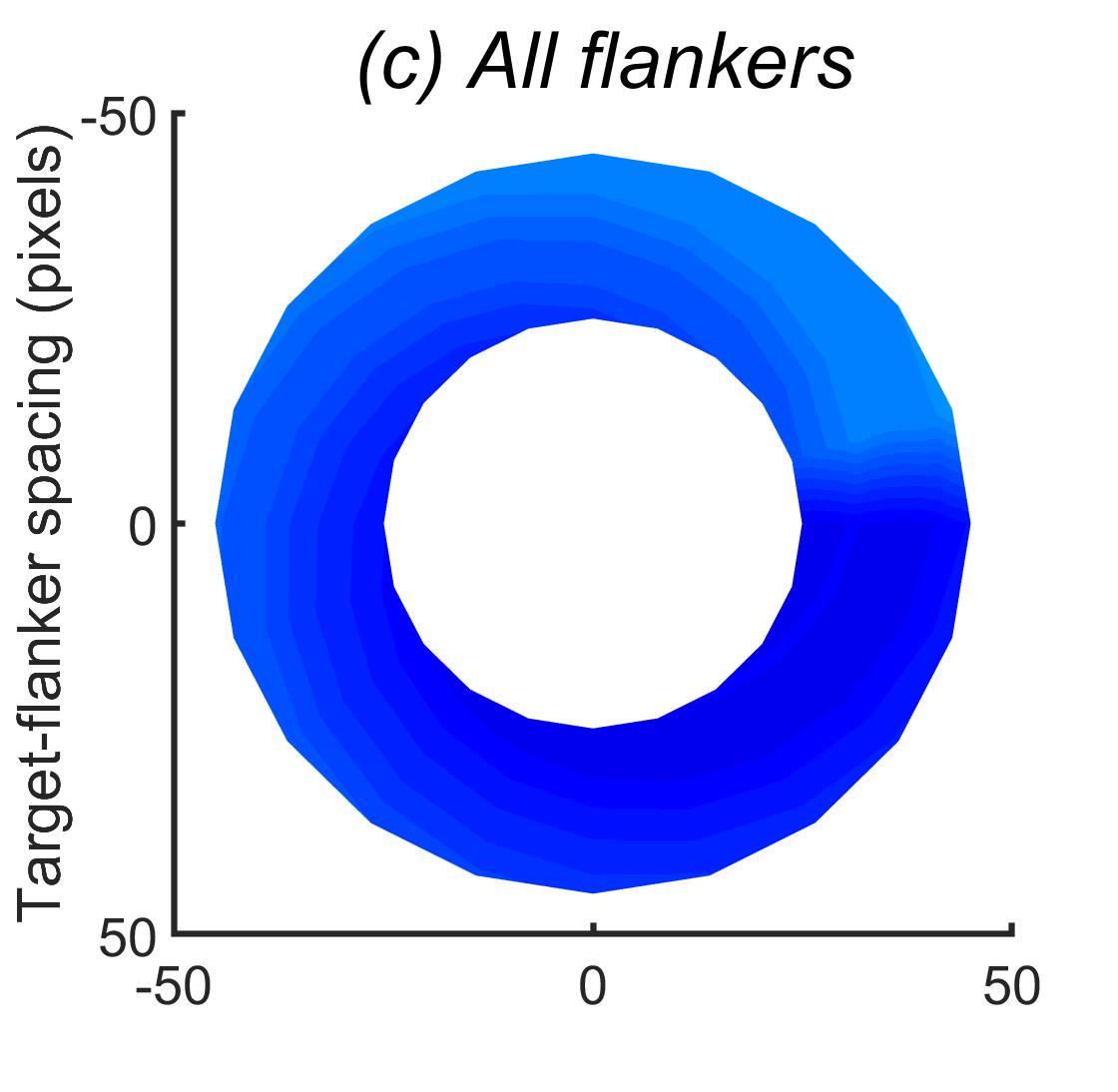}
		\end{subfigure}
		\begin{subfigure}{0.25\textwidth}
			\includegraphics[width=\textwidth]{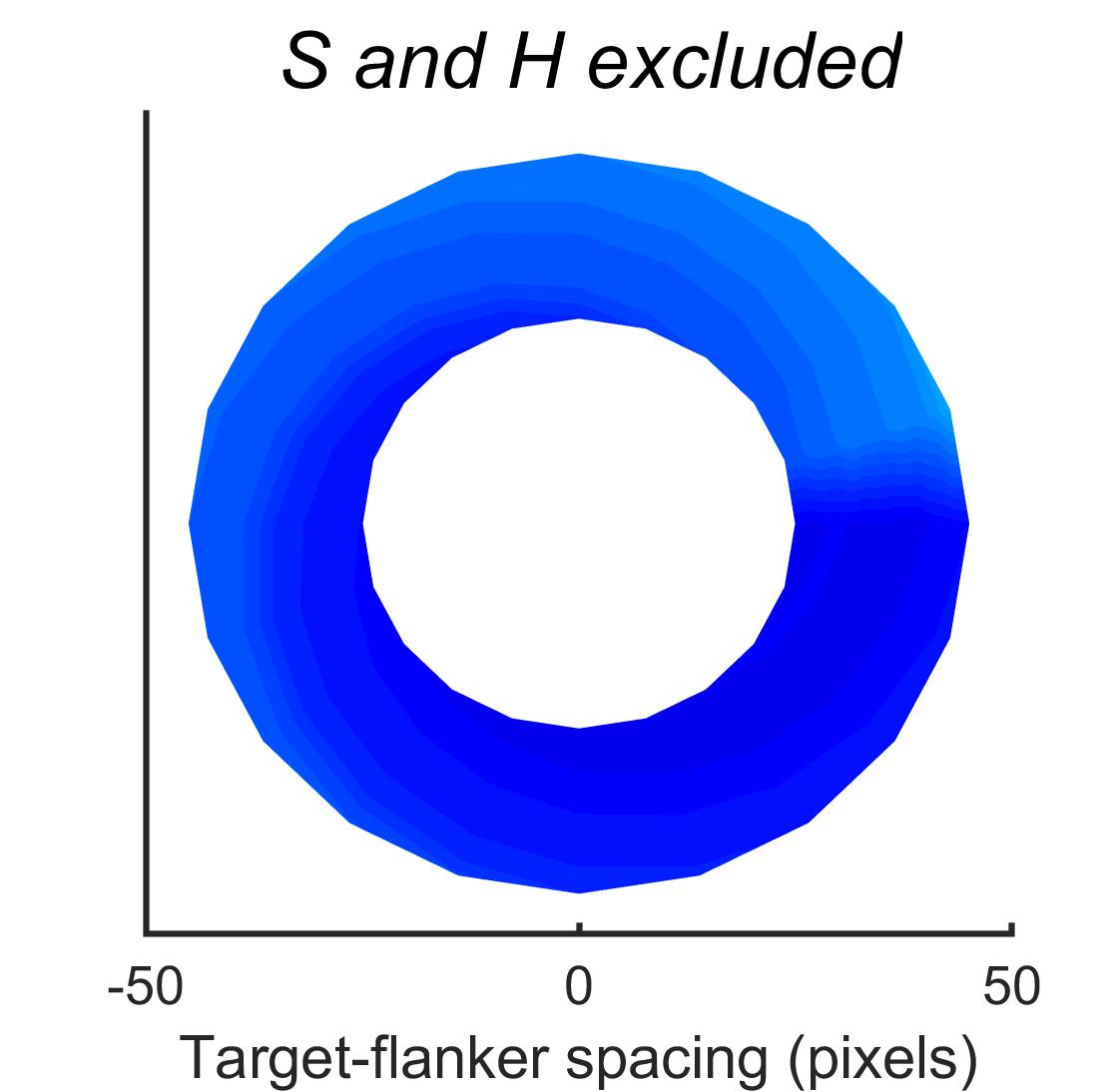}
		\end{subfigure}
		\begin{subfigure}{0.25\textwidth}
			\includegraphics[width=\textwidth]{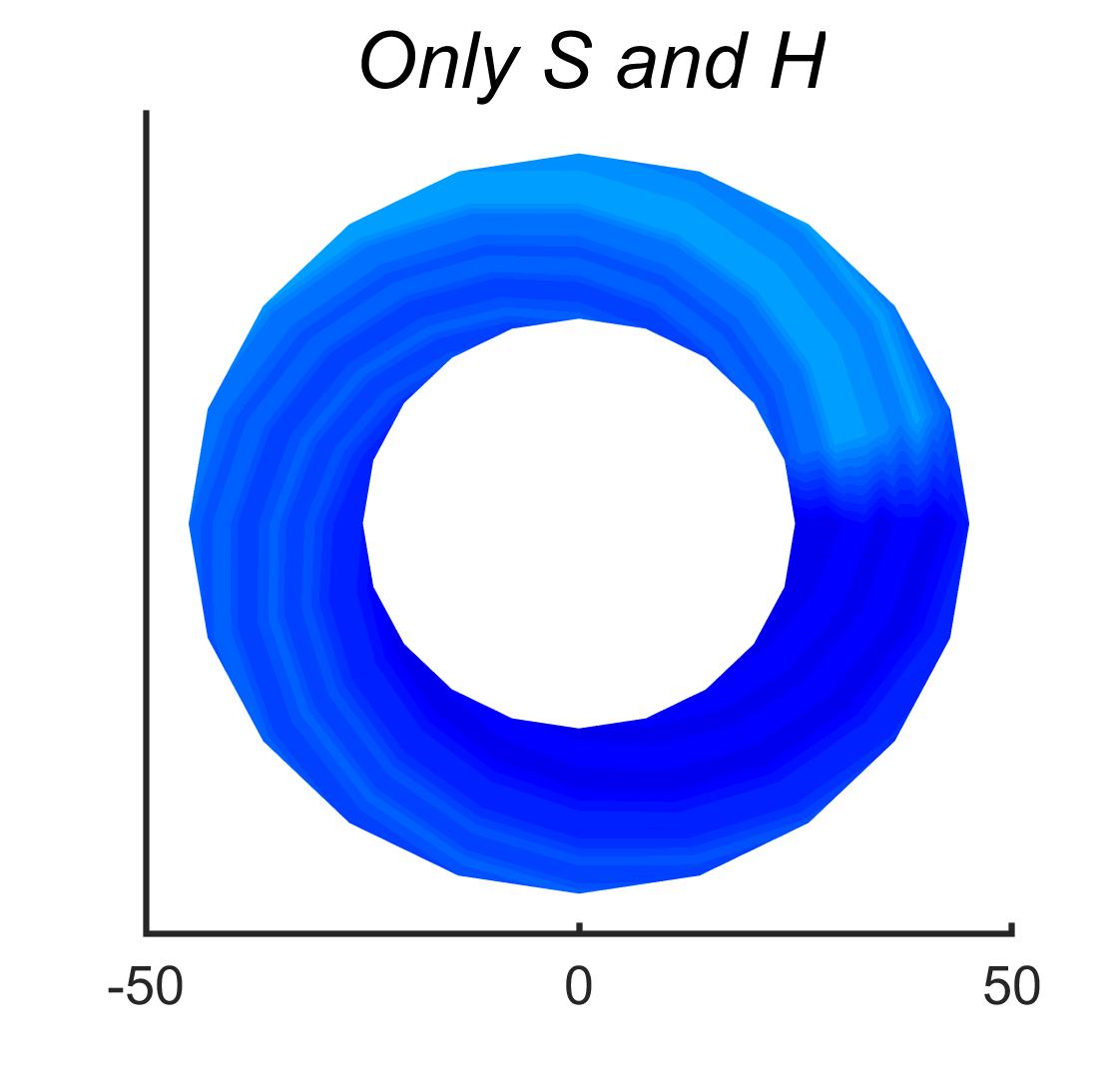}
		\end{subfigure}
		\begin{subfigure}{0.06\textwidth}
			\includegraphics[width=\textwidth]{other/colorbar.jpg}
		\end{subfigure}
		\caption{Accuracy of letter identification of the randomly initialised small 5-layer convolutional network with single flankers ('SimpleNet'). A total of five independent training sessions and test sessions are combined in this figure. Training of all five models in this figure was done without acuity loss, and testing was done with acuity loss. Average accuracy without flankers was 50.02\%.}
		\label{smallrandominitacuitylossconcat}
	\end{figure}
	
	\begin{figure}
		\centering
		\begin{subfigure}{0.9\textwidth}
			\includegraphics[width=\textwidth]{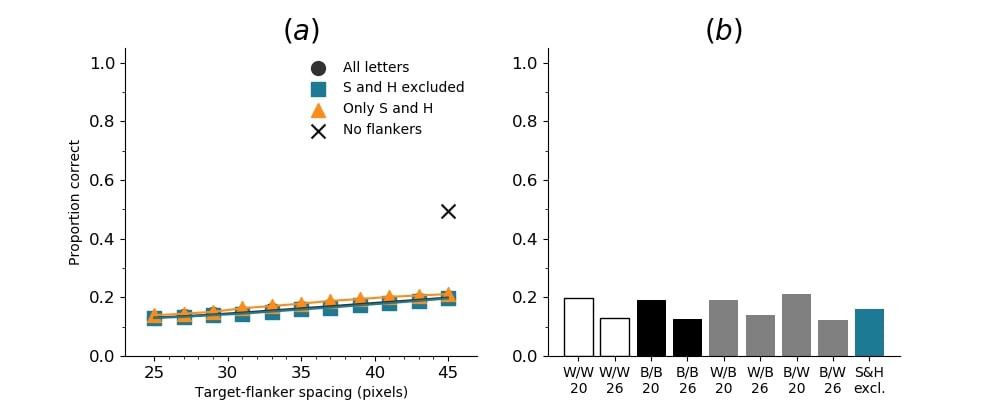}
		\end{subfigure}
		\begin{subfigure}{0.2562\textwidth}
			\includegraphics[width=\textwidth]{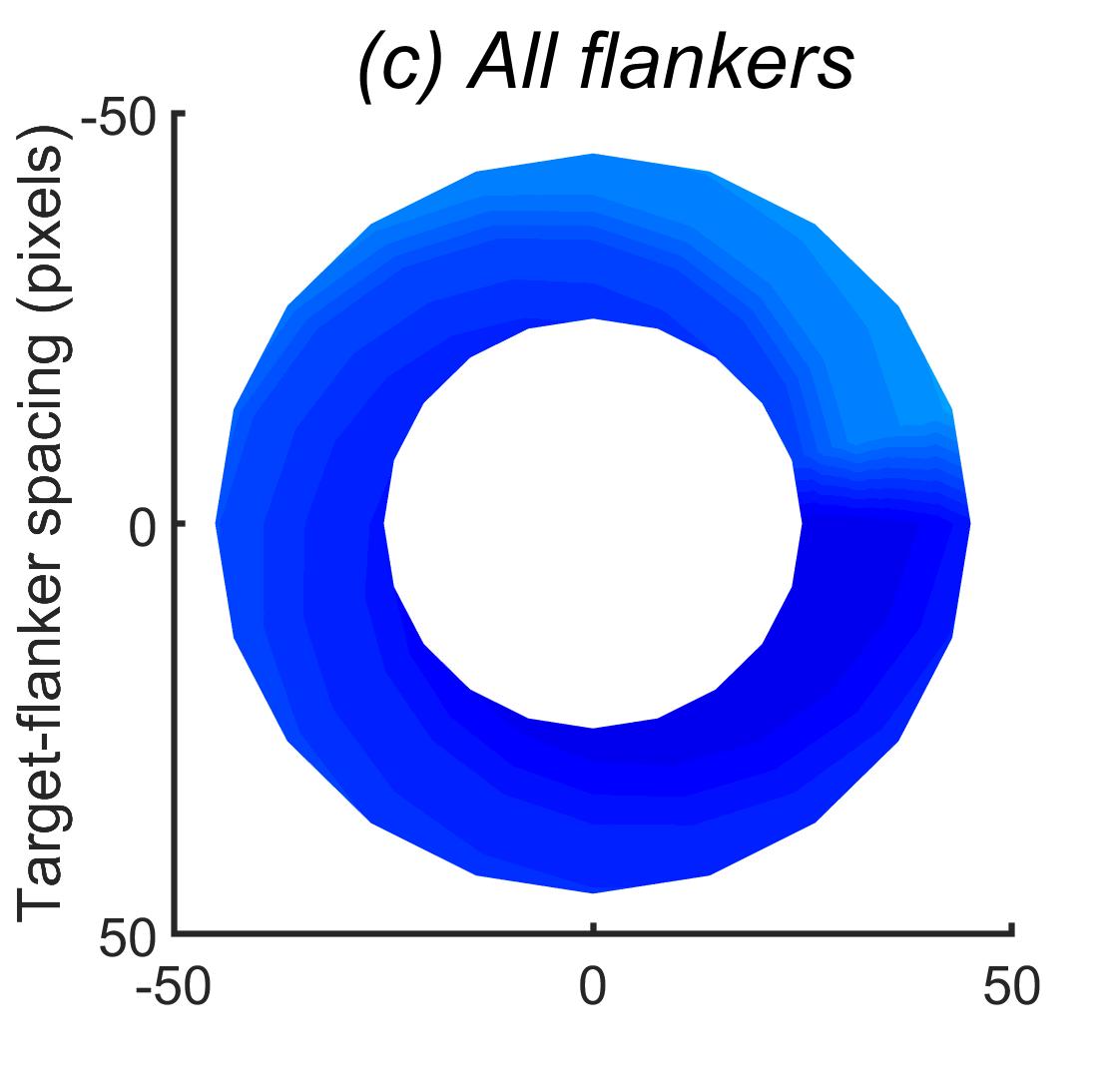}
		\end{subfigure}
		\begin{subfigure}{0.25\textwidth}
			\includegraphics[width=\textwidth]{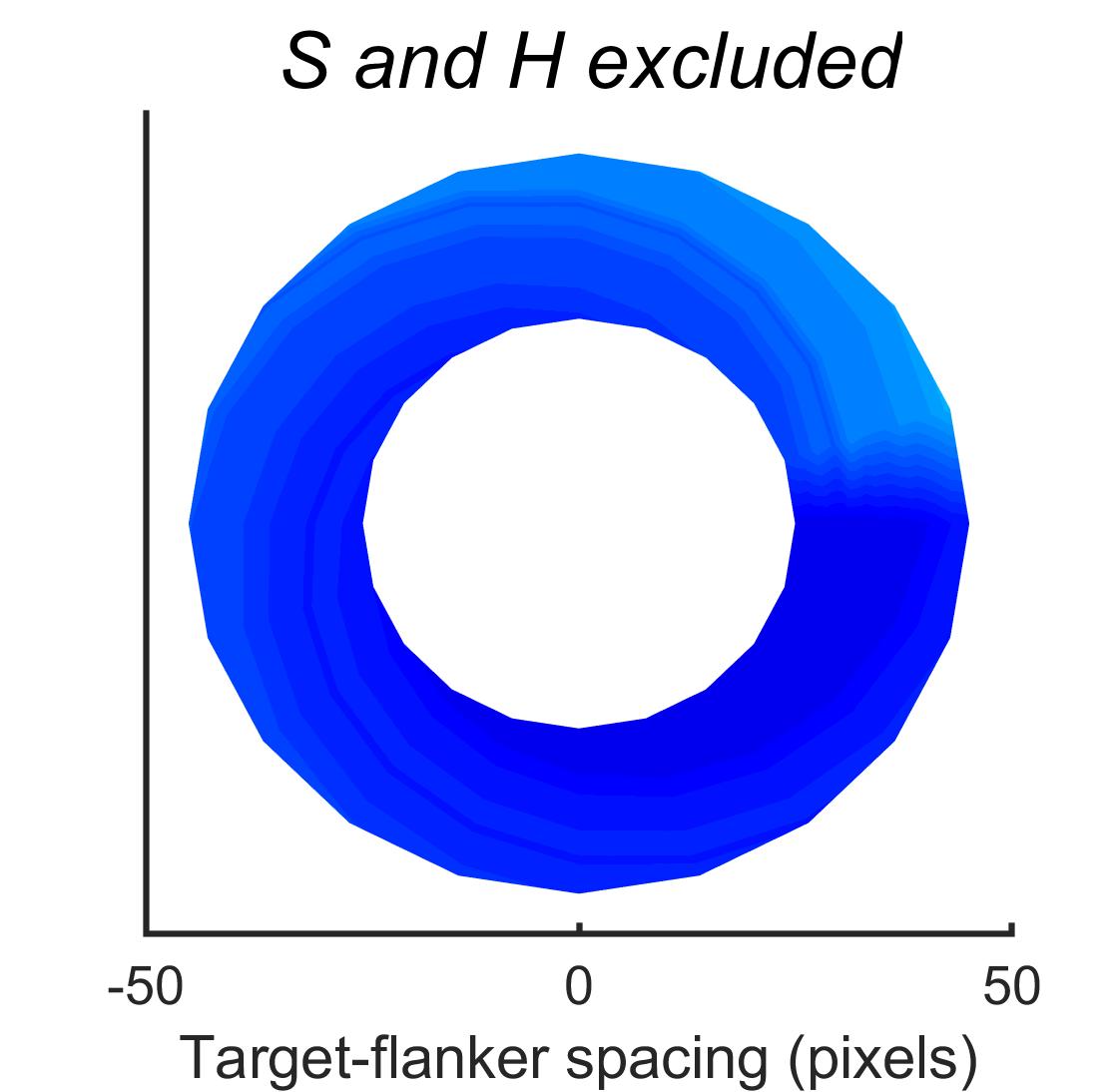}
		\end{subfigure}
		\begin{subfigure}{0.25\textwidth}
			\includegraphics[width=\textwidth]{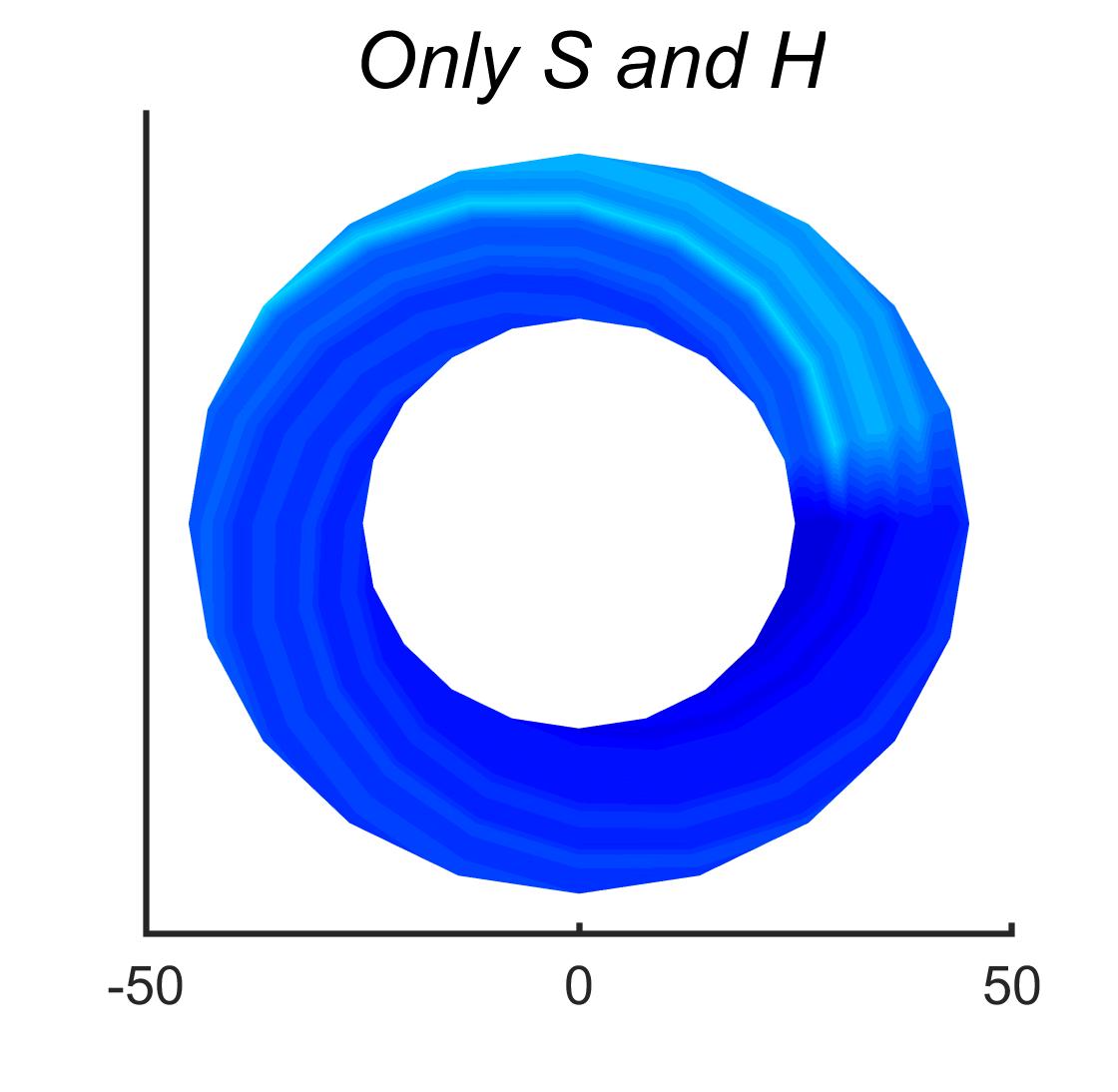}
		\end{subfigure}
		\begin{subfigure}{0.06\textwidth}
			\includegraphics[width=\textwidth]{other/colorbar.jpg}
		\end{subfigure}
		\caption{Accuracy of letter identification of the first randomly initialised small 5-layer convolutional network with single flankers ('SimpleNet 1'). Training was done without acuity loss, and testing with acuity loss. Average accuracy for acuity-reduced data without flankers was 49.33\%.}
		\label{smallrandominitacuityloss1}
	\end{figure}
	
	\begin{figure}
		\centering
		\begin{subfigure}{0.9\textwidth}
			\includegraphics[width=\textwidth]{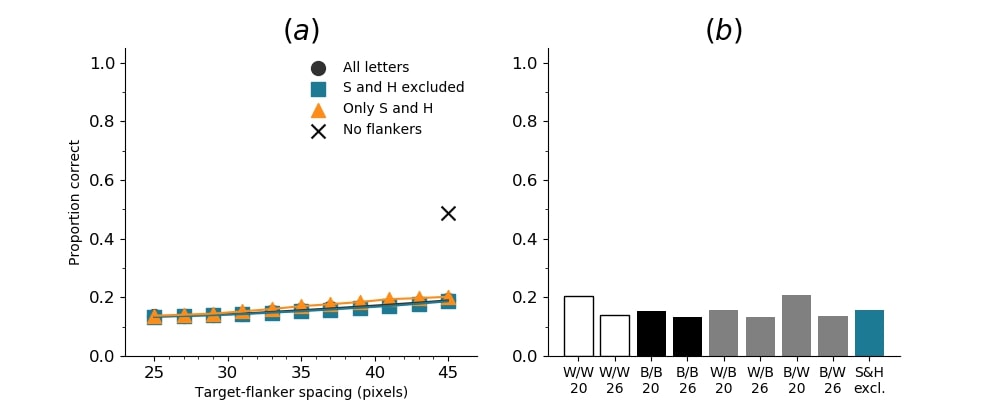}
		\end{subfigure}
		\begin{subfigure}{0.2562\textwidth}
			\includegraphics[width=\textwidth]{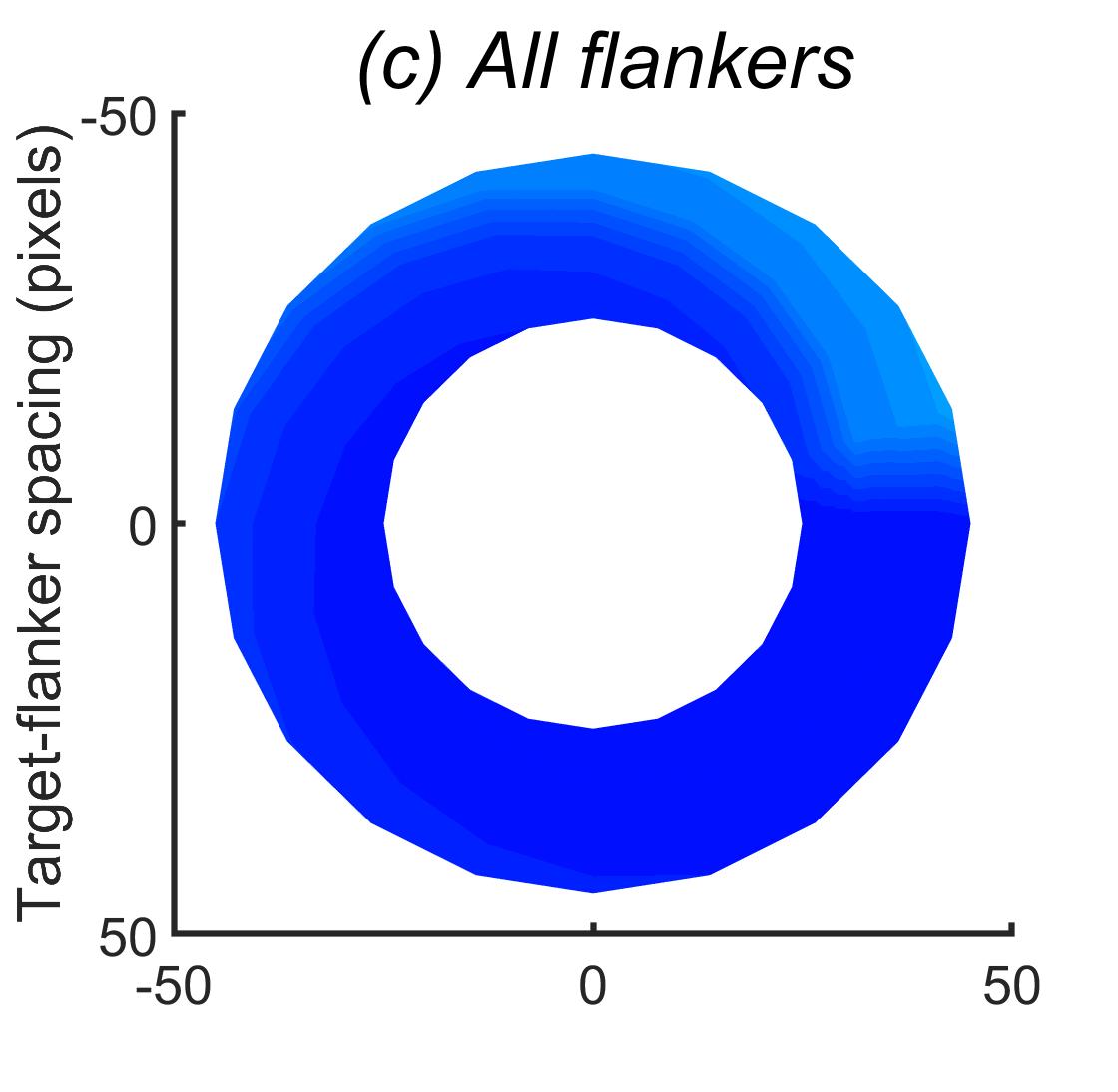}
		\end{subfigure}
		\begin{subfigure}{0.25\textwidth}
			\includegraphics[width=\textwidth]{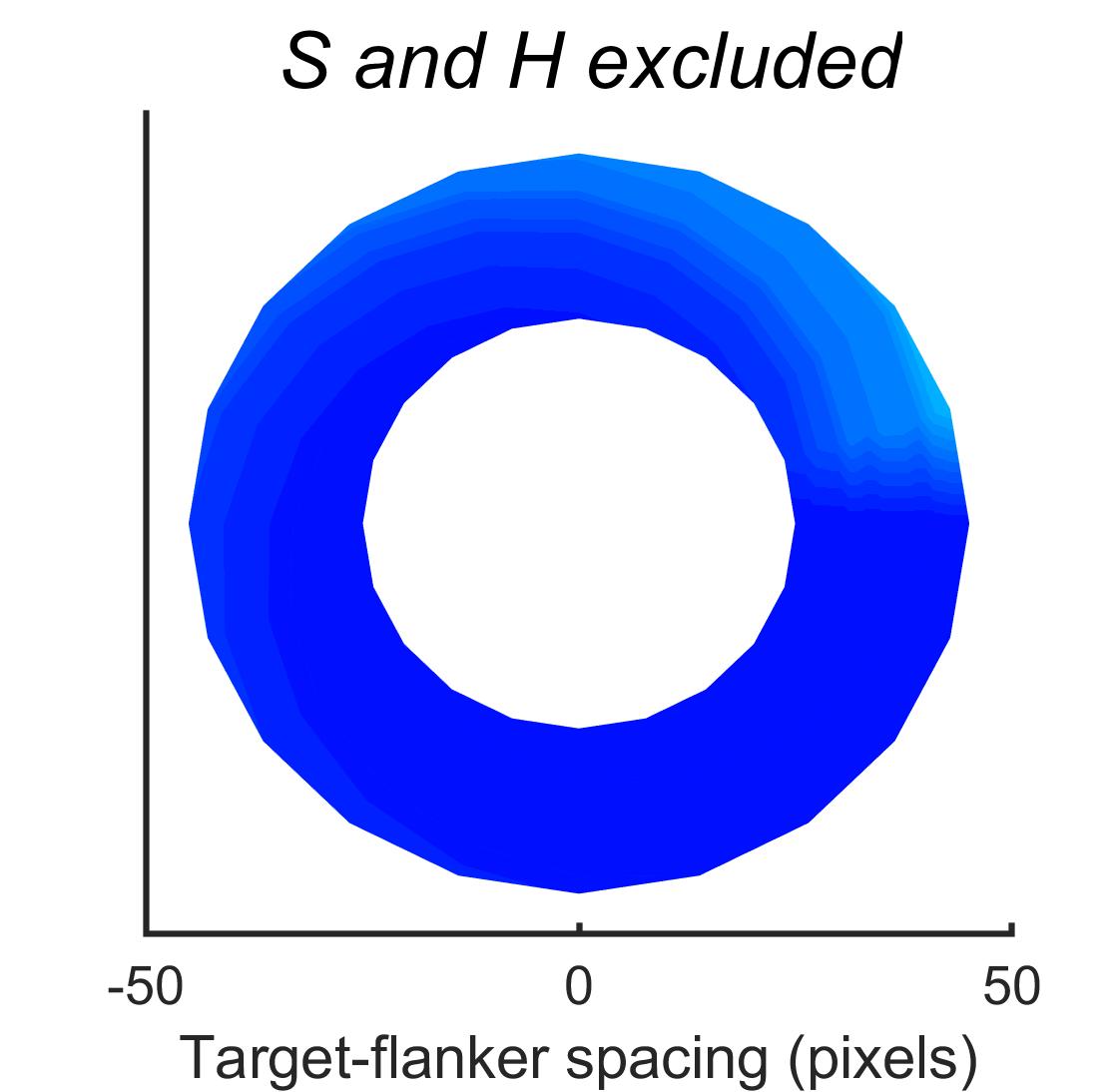}
		\end{subfigure}
		\begin{subfigure}{0.25\textwidth}
			\includegraphics[width=\textwidth]{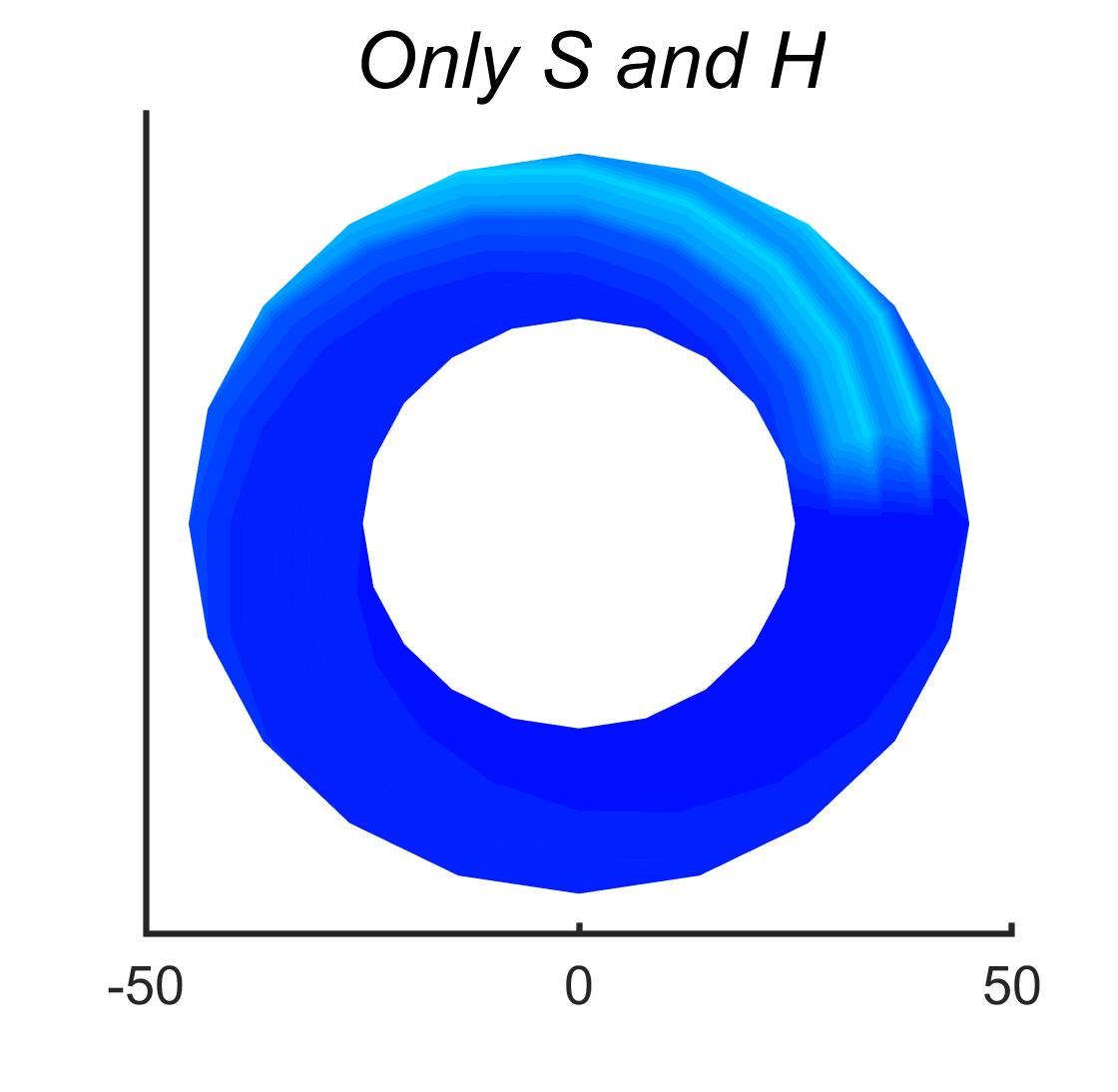}
		\end{subfigure}
		\begin{subfigure}{0.06\textwidth}
			\includegraphics[width=\textwidth]{other/colorbar.jpg}
		\end{subfigure}
		\caption{Accuracy of letter identification of the second randomly initialised small 5-layer convolutional network with single flankers ('SimpleNet 2'). Training was done without acuity loss, and testing with acuity loss. Average accuracy for acuity-reduced data without flankers was 48.91\%.}
		\label{smallrandominitacuityloss2}
	\end{figure}
	
	\begin{figure}
		\centering
		\begin{subfigure}{0.9\textwidth}
			\includegraphics[width=\textwidth]{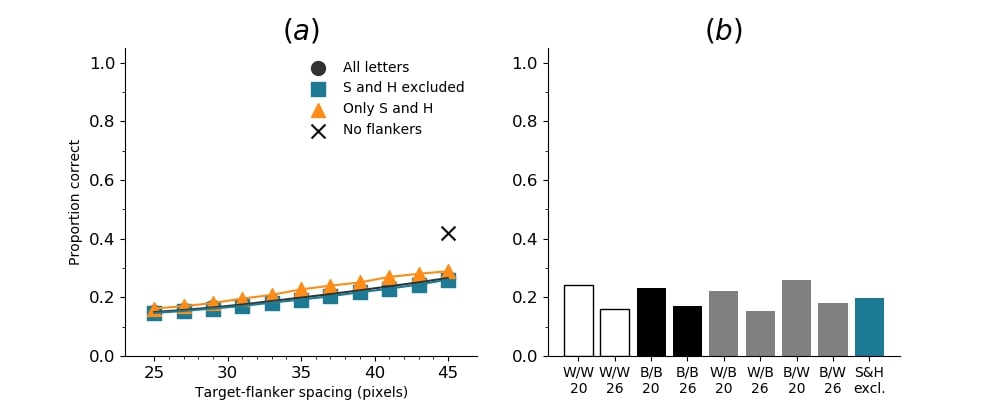}
		\end{subfigure}
		\begin{subfigure}{0.2562\textwidth}
			\includegraphics[width=\textwidth]{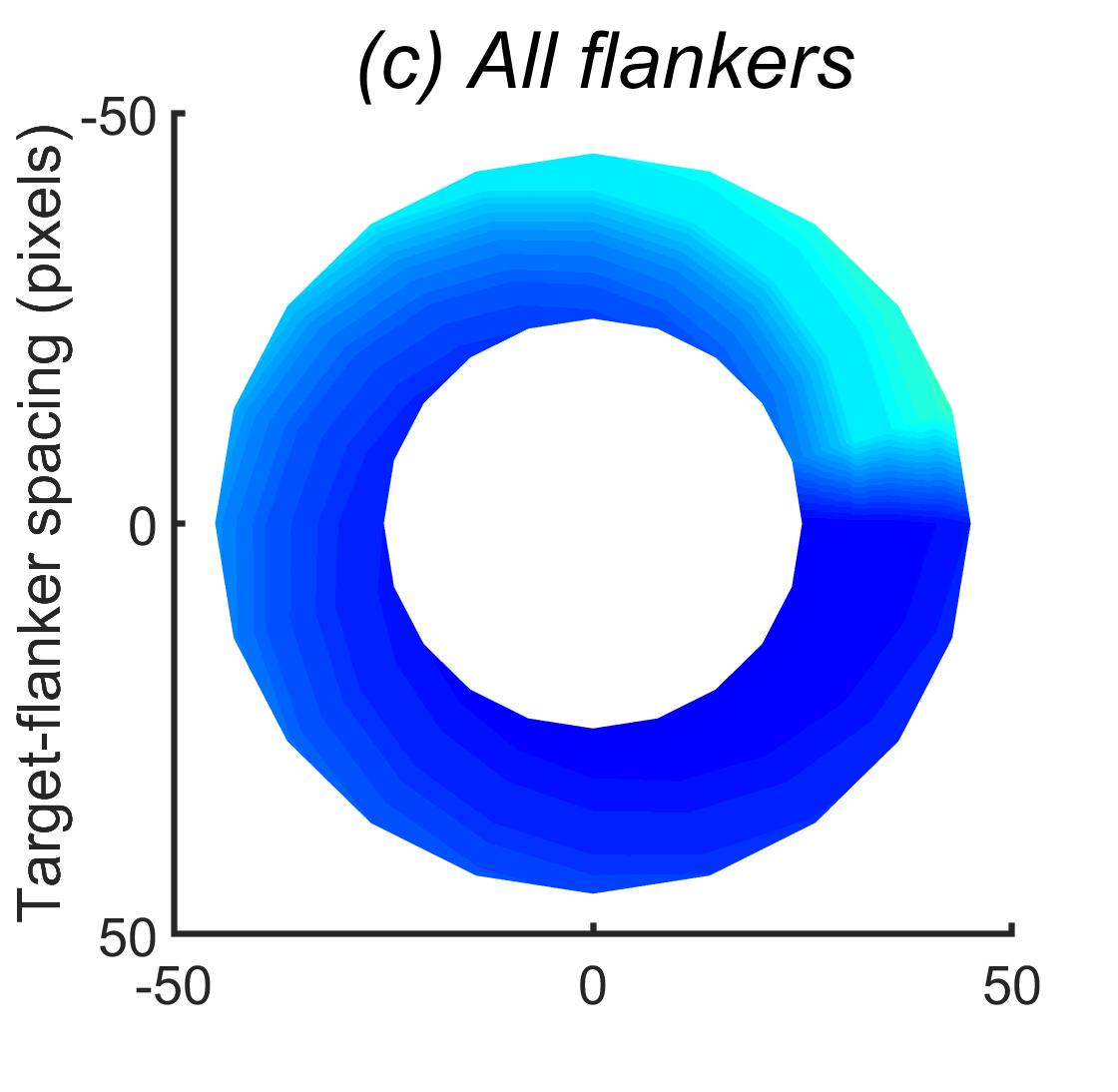}
		\end{subfigure}
		\begin{subfigure}{0.25\textwidth}
			\includegraphics[width=\textwidth]{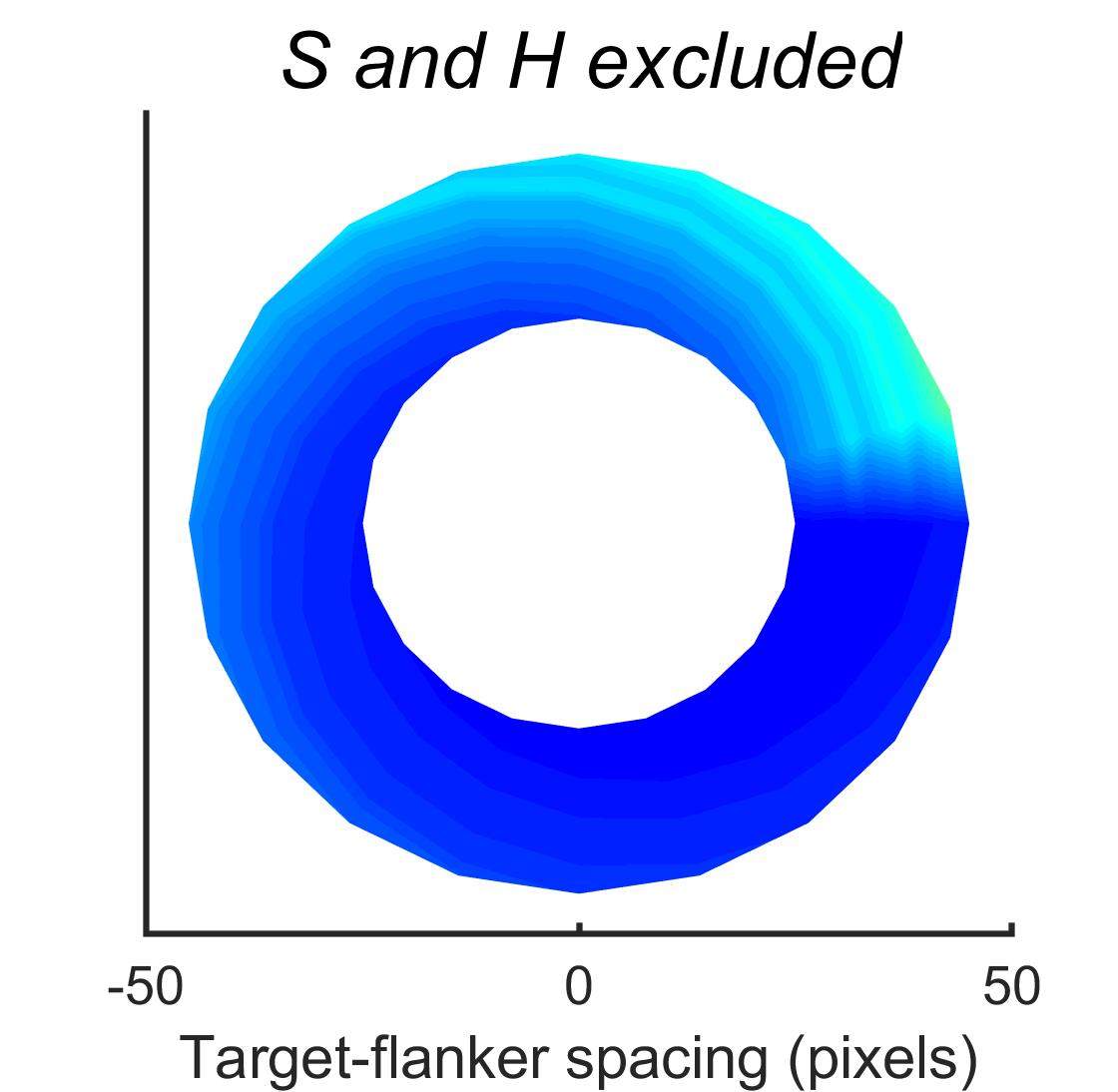}
		\end{subfigure}
		\begin{subfigure}{0.25\textwidth}
			\includegraphics[width=\textwidth]{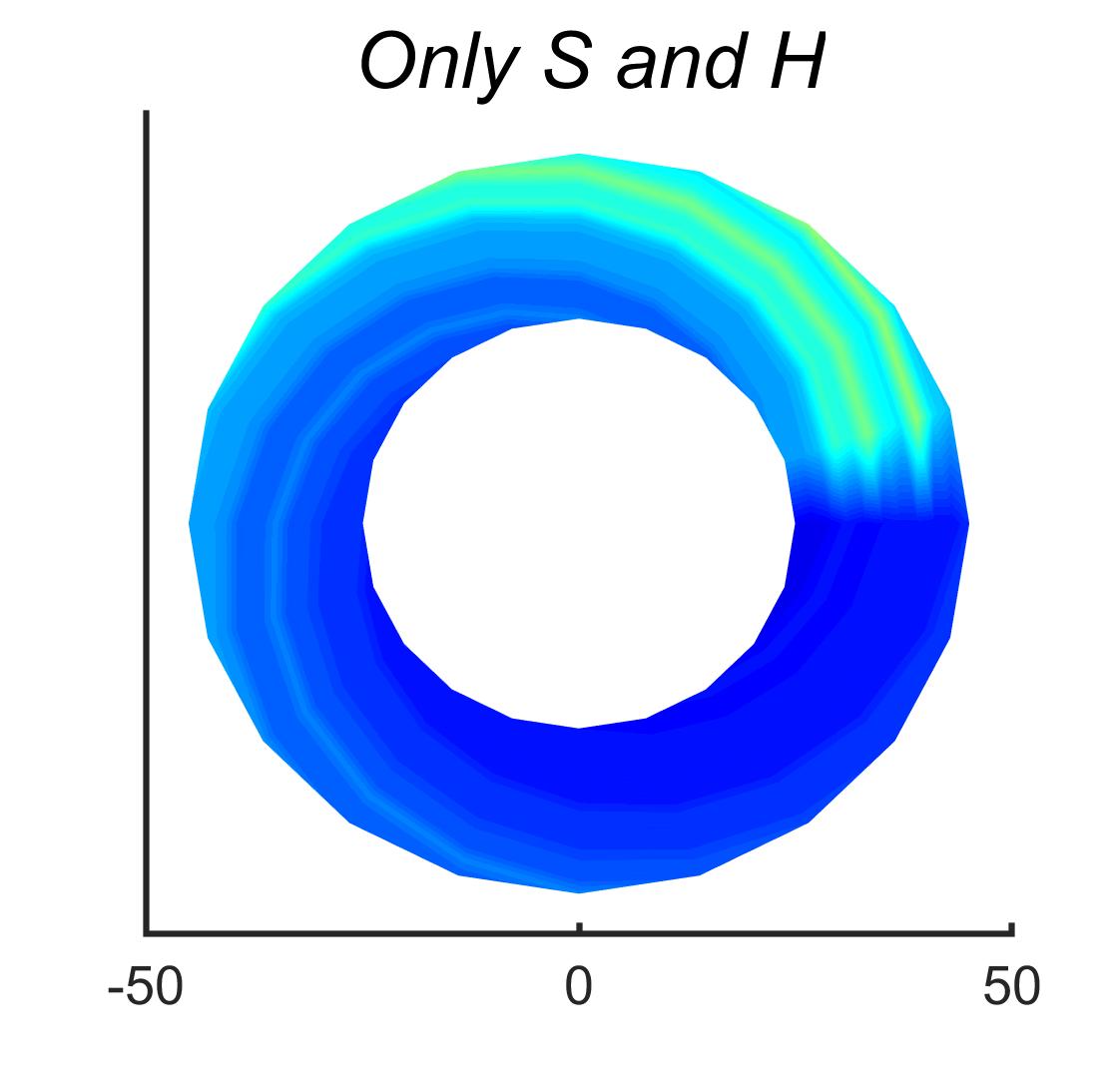}
		\end{subfigure}
		\begin{subfigure}{0.06\textwidth}
			\includegraphics[width=\textwidth]{other/colorbar.jpg}
		\end{subfigure}
		\caption{Accuracy of letter identification of the third randomly initialised small 5-layer convolutional network with single flankers ('SimpleNet 3'). Training was done without acuity loss, and testing with acuity loss. Average accuracy for acuity-reduced data without flankers was 42.06\%.}
		\label{smallrandominitacuityloss3}
	\end{figure}
	
	\begin{figure}
		\centering
		\begin{subfigure}{0.9\textwidth}
			\includegraphics[width=\textwidth]{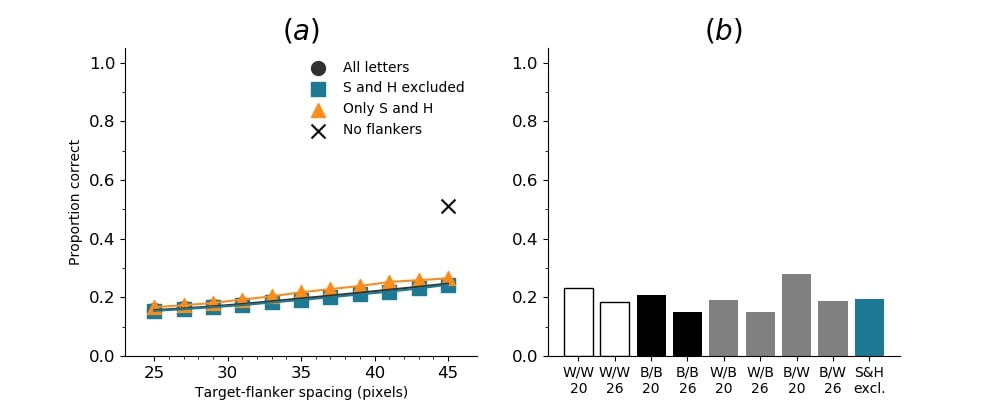}
		\end{subfigure}
		\begin{subfigure}{0.2562\textwidth}
			\includegraphics[width=\textwidth]{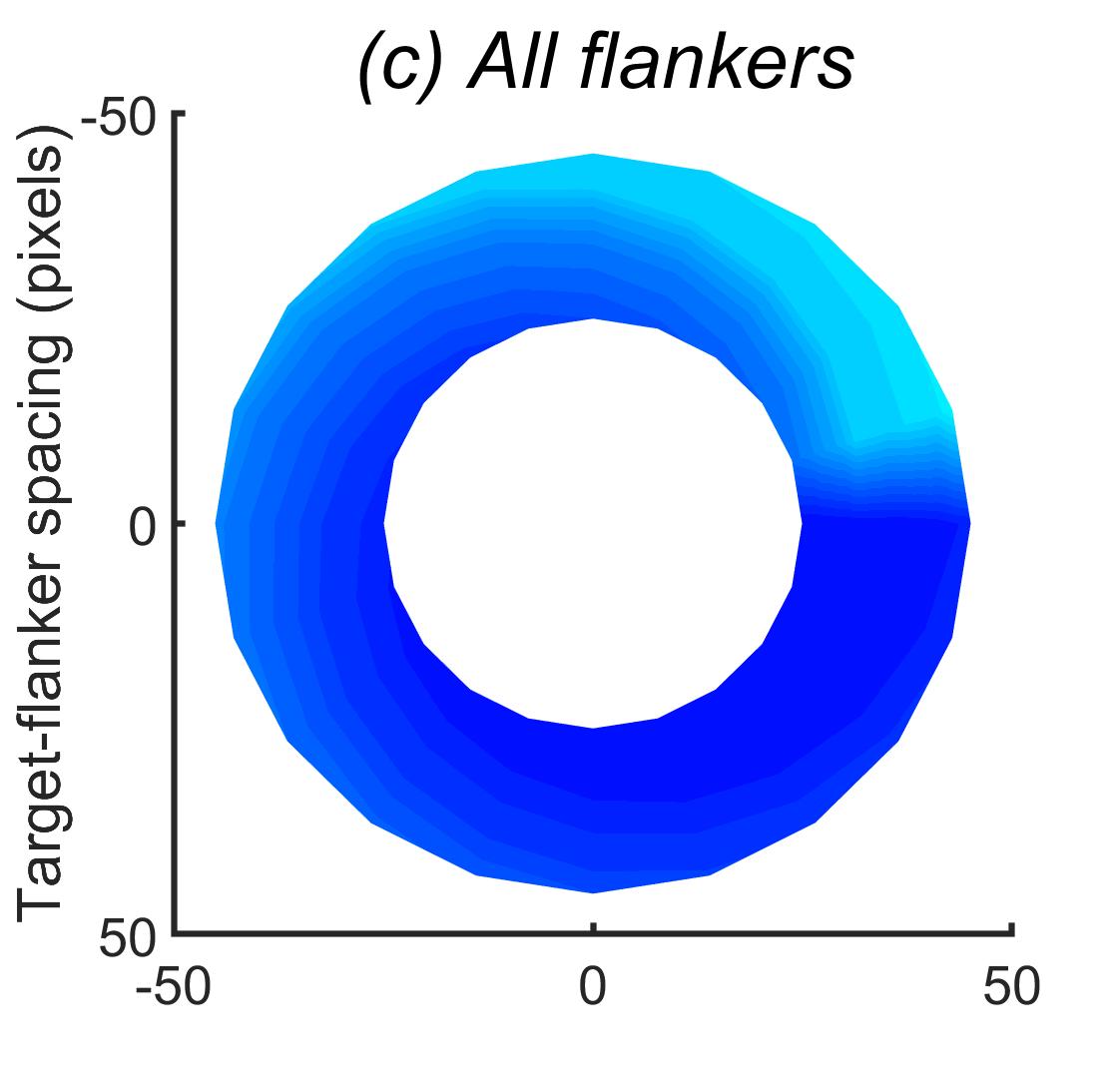}
		\end{subfigure}
		\begin{subfigure}{0.25\textwidth}
			\includegraphics[width=\textwidth]{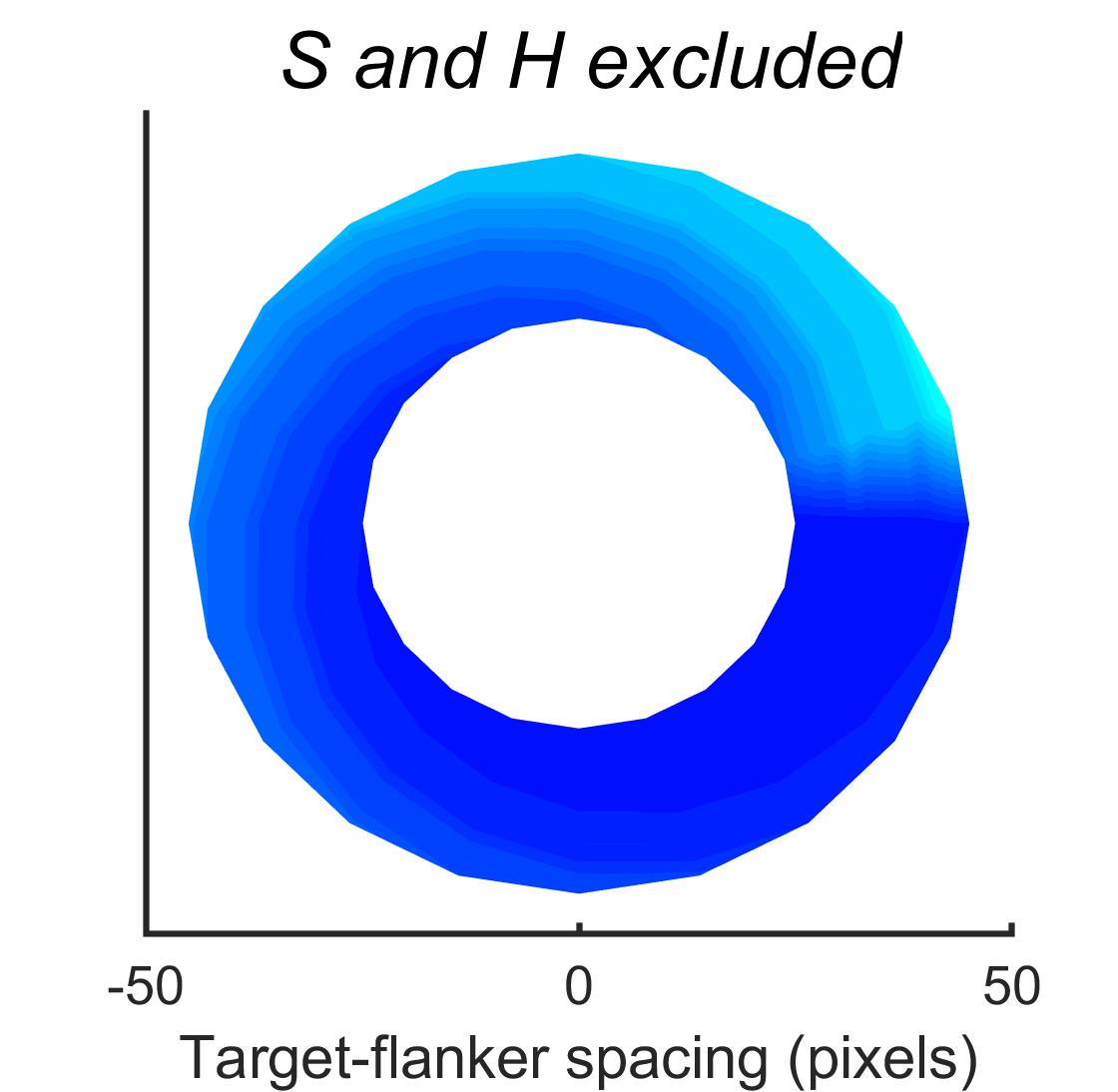}
		\end{subfigure}
		\begin{subfigure}{0.25\textwidth}
			\includegraphics[width=\textwidth]{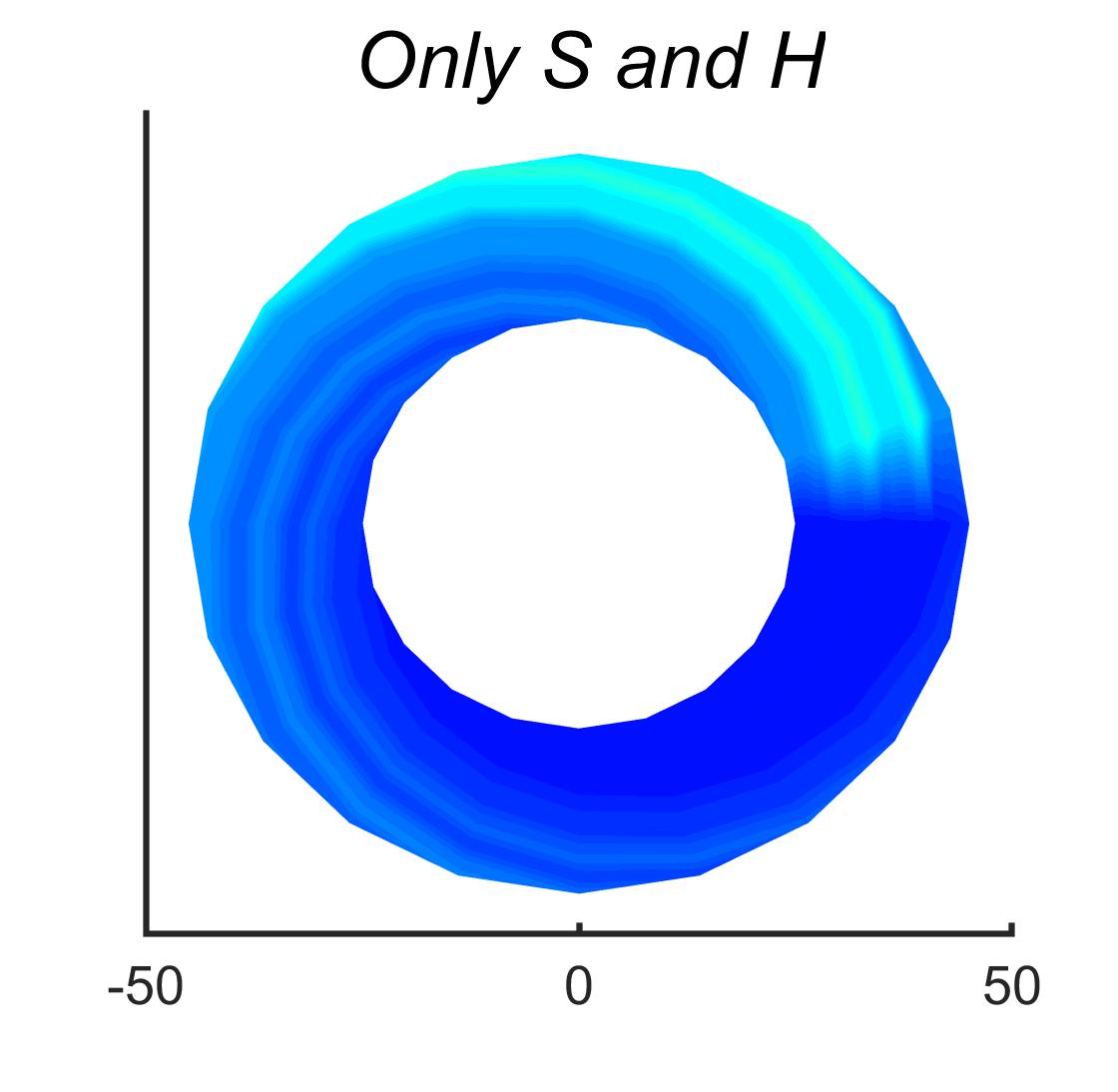}
		\end{subfigure}
		\begin{subfigure}{0.06\textwidth}
			\includegraphics[width=\textwidth]{other/colorbar.jpg}
		\end{subfigure}
		\caption{Accuracy of letter identification of the fourth randomly initialised small 5-layer convolutional network with single flankers ('SimpleNet 4'). Training was done without acuity loss, and testing with acuity loss. Average accuracy for acuity-reduced data without flankers was 51.21\%.}
		\label{smallrandominitacuityloss4}
	\end{figure}
	
	\begin{figure}
		\centering
		\begin{subfigure}{0.9\textwidth}
			\includegraphics[width=\textwidth]{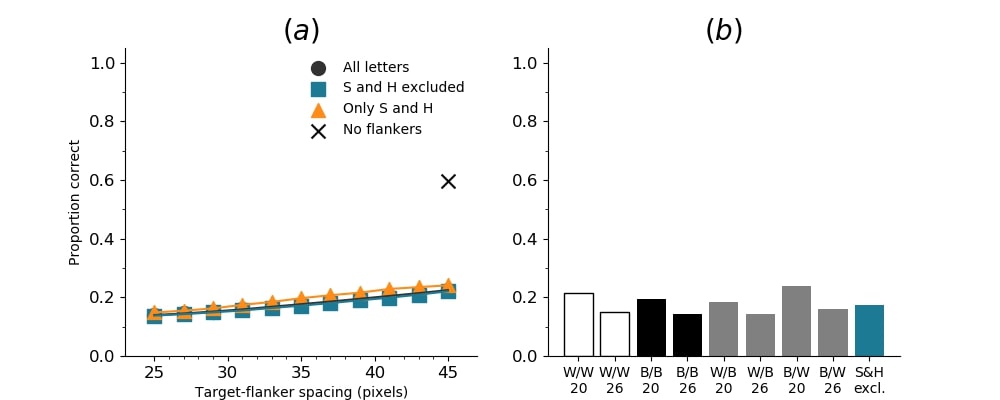}
		\end{subfigure}
		\begin{subfigure}{0.2562\textwidth}
			\includegraphics[width=\textwidth]{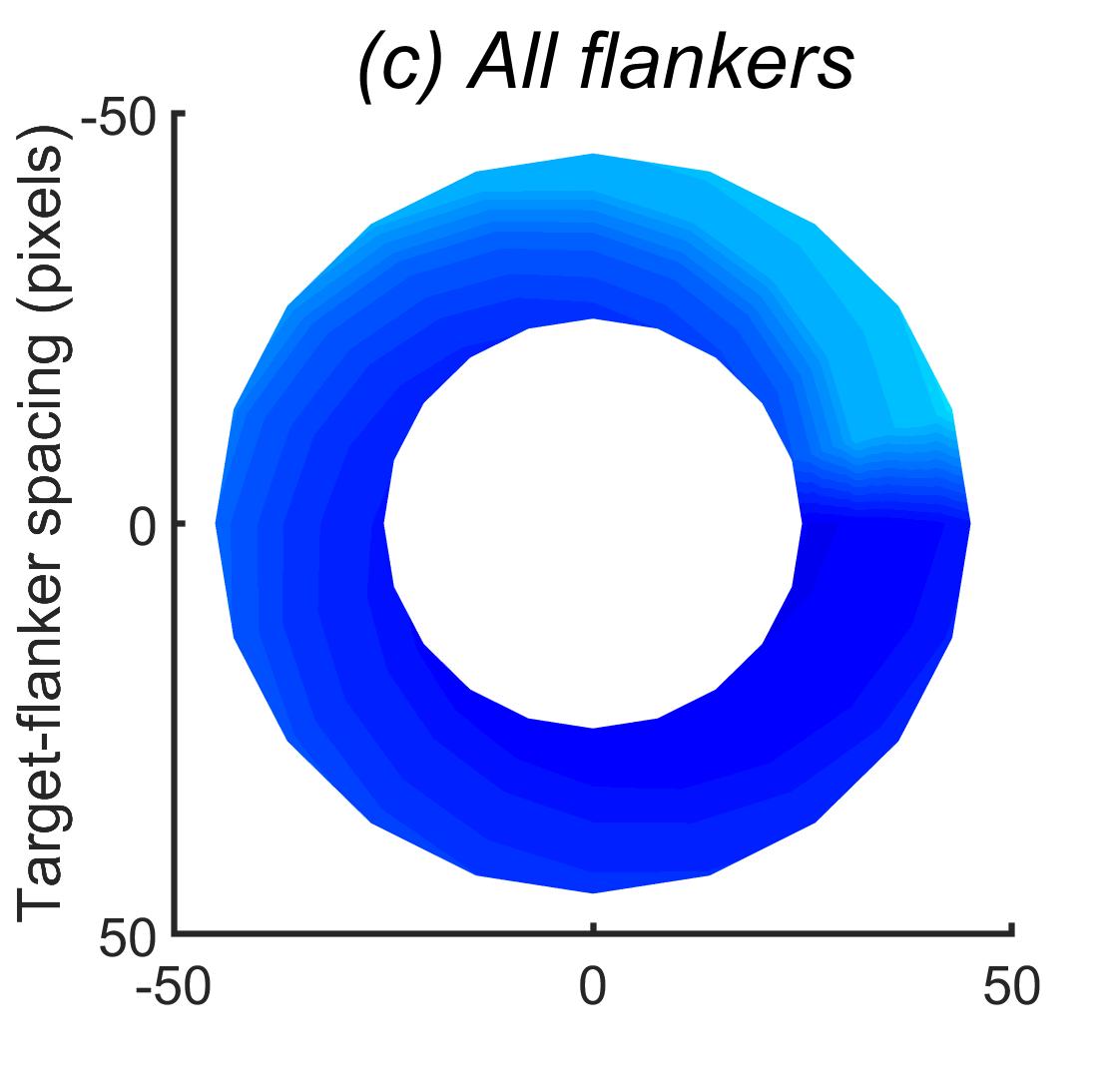}
		\end{subfigure}
		\begin{subfigure}{0.25\textwidth}
			\includegraphics[width=\textwidth]{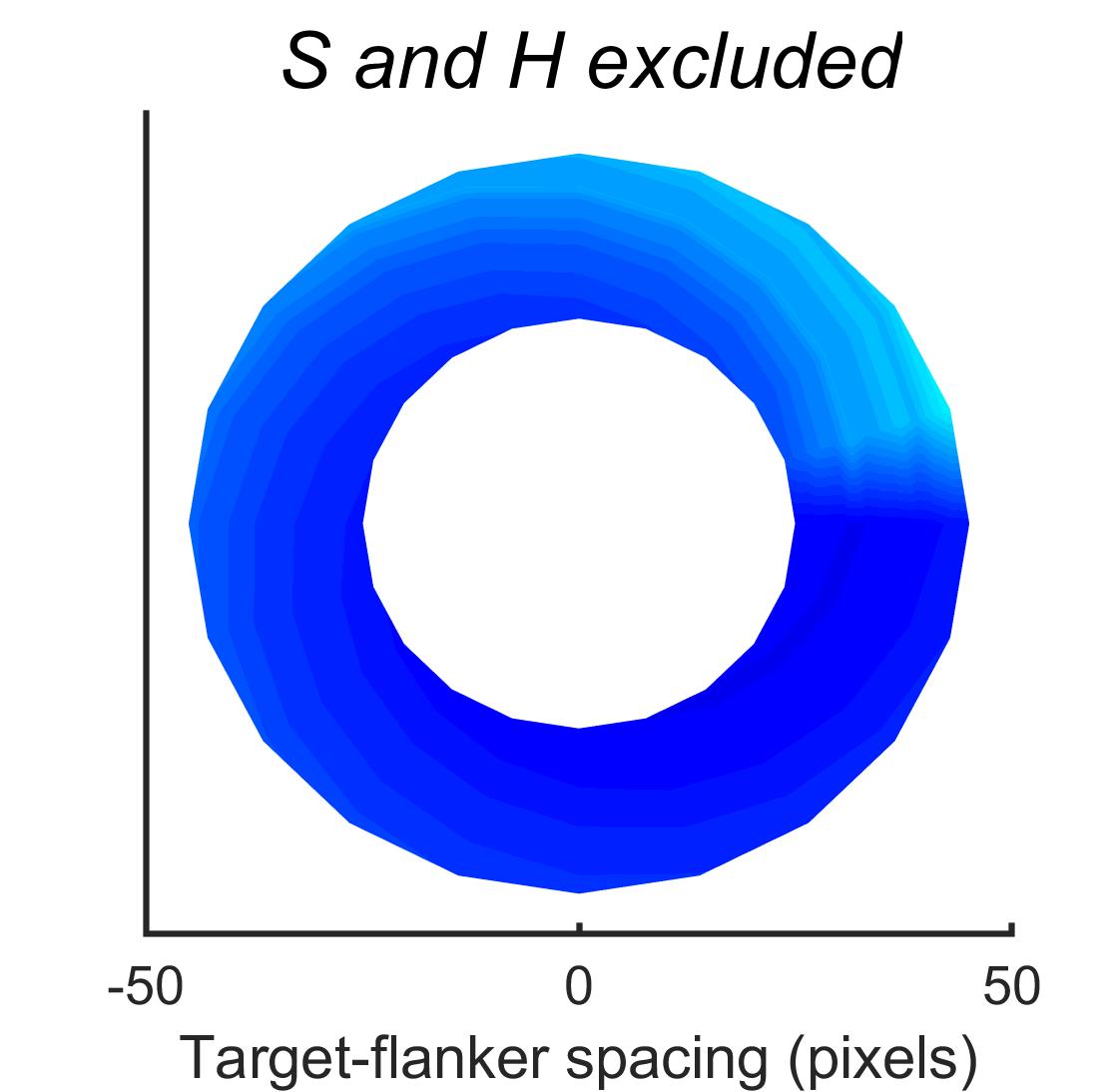}
		\end{subfigure}
		\begin{subfigure}{0.25\textwidth}
			\includegraphics[width=\textwidth]{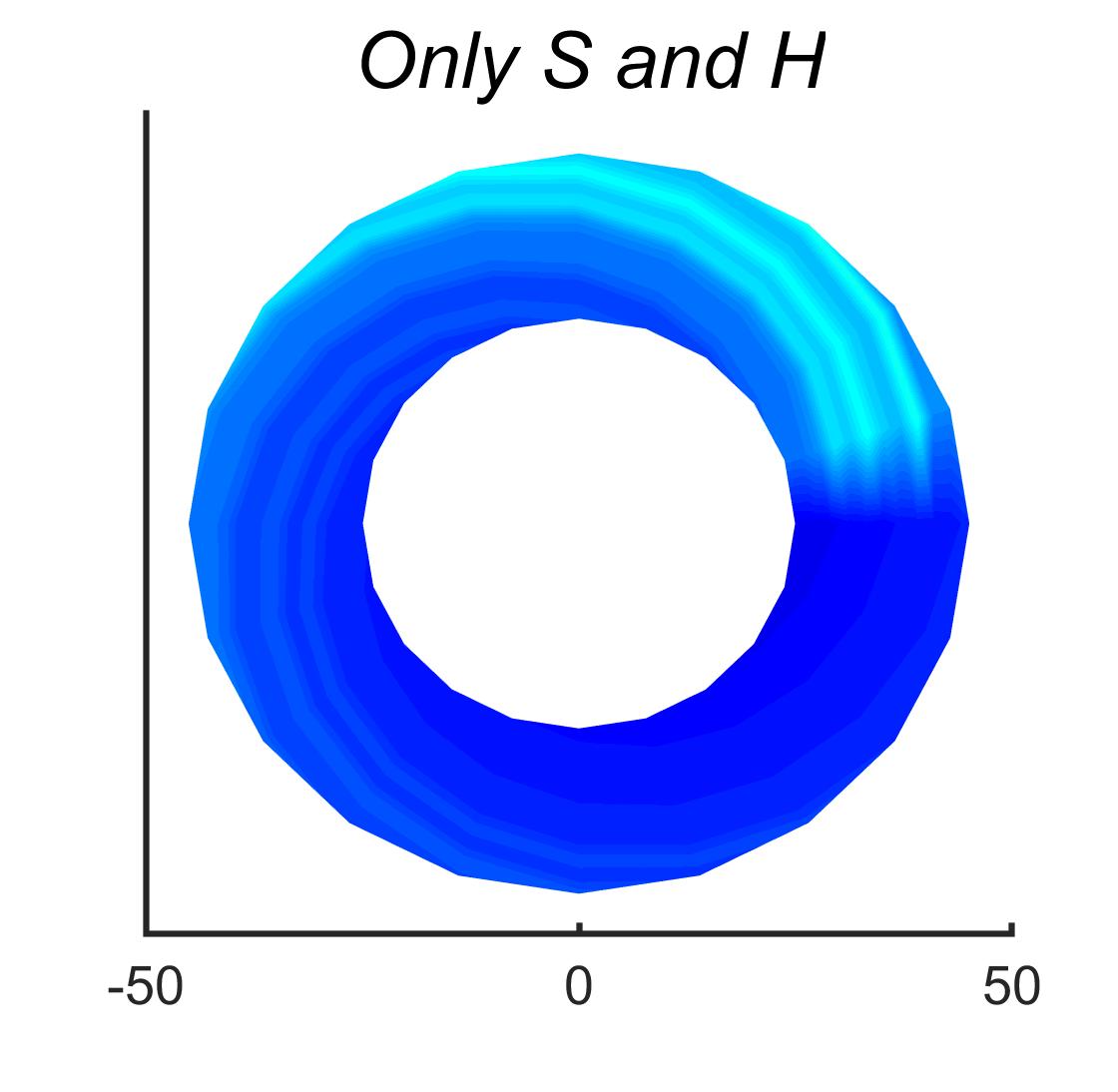}
		\end{subfigure}
		\begin{subfigure}{0.06\textwidth}
			\includegraphics[width=\textwidth]{other/colorbar.jpg}
		\end{subfigure}
		\caption{Accuracy of letter identification of the fifth randomly initialised small 5-layer convolutional network with single flankers ('SimpleNet 5'). Training was done without acuity loss, and testing with acuity loss. Average accuracy for acuity-reduced data without flankers was 59.51\%.}
		\label{smallrandominitacuityloss5}
	\end{figure}
	
	\begin{figure}
		\centering
		\includegraphics[width=0.8\textwidth]{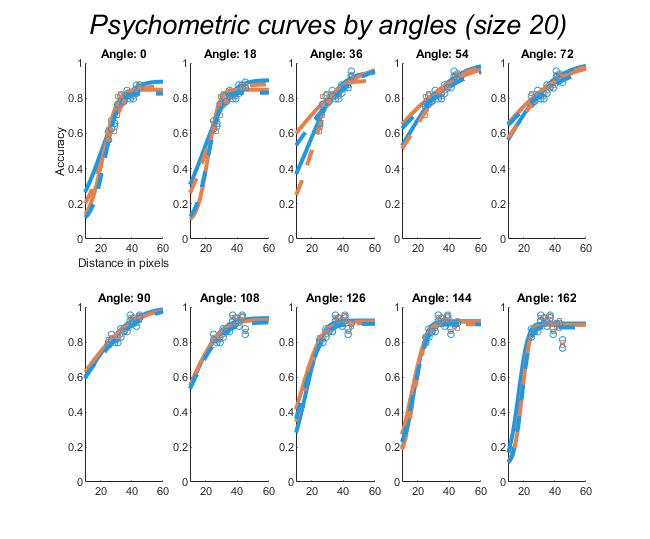}
		\caption{Psychometric curves for the randomly initialised DenseNet-121 with pair flankers without the flanker S and H, trained and tested with acuity loss. Even when the unseen flankers are omitted, the curve fits are not consistent. Markers show data points, while lines show Gauss error function fits.}
		\label{psychometric}
	\end{figure}
	
\end{document}